\documentclass{article} 
\usepackage{arxiv,times}

\usepackage{amsmath,amsfonts,bm}

\def\eqref#1{equation~\ref{#1}}

\def\1{\bm{1}}

\DeclareMathAlphabet{\mathsfit}{\encodingdefault}{\sfdefault}{m}{sl}
\SetMathAlphabet{\mathsfit}{bold}{\encodingdefault}{\sfdefault}{bx}{n}

\PassOptionsToPackage{table,dvipsnames}{xcolor}

\usepackage[pagebackref=true]{hyperref}
\usepackage{url}
\usepackage{enumerate}
\usepackage{enumitem}
\usepackage{graphicx}
\usepackage{wrapfig}
\usepackage{booktabs,tabularx}
\usepackage{needspace}
\usepackage{tikz}
\usetikzlibrary{arrows.meta,positioning,calc,decorations.pathreplacing}
\usepackage{mathtools}
\usepackage{backref}
\usepackage{amsmath,amssymb,amsthm}
\usepackage{thm-restate}
\usepackage[noabbrev]{cleveref}

\usepackage[most]{tcolorbox}
\definecolor{keylavender}{RGB}{245,245,250}
\tcbset{common/.style={
  enhanced, breakable,
  boxsep=6pt, left=8pt, right=8pt, top=6pt, bottom=6pt,
  before skip=8pt, after skip=8pt,
  arc=10pt
}}
\newtcolorbox{lavenderbox}{
  common,
  colback=keylavender,
  colframe=keylavender,
  boxrule=0pt,
  arc=5pt,
  left=3pt, right=3pt, top=0pt, bottom=0pt
}

\newtheorem{proposition}{Proposition}

\newtheorem{corollary}{Corollary}
\newtheorem{lemma}{Lemma}
\theoremstyle{definition}

\theoremstyle{plain}
\usepackage{multirow}
\usepackage{comment}
\newcommand{\ours}{RLB}
\newcommand{\fitwidth}[1]{\resizebox{\ifdim\width>\linewidth\linewidth\else\width\fi}{!}{#1}}

\renewcommand*{\backref}[1]{}
\renewcommand*{\backrefalt}[4]{\ifcase#1\relax\or(page #2)\else(pages #2)\fi}

\AddToHook{cmd/appendix/before}{%
    \crefalias{section}{appendix}%
    \crefalias{subsection}{appendix}
}
\renewcommand{\eqref}[1]{(\ref{#1})}
\crefformat{equation}{(#2#1#3)}
\Crefformat{equation}{(#2#1#3)}
\crefrangeformat{equation}{(#3#1#4)--(#5#2#6)}
\Crefrangeformat{equation}{(#3#1#4)--(#5#2#6)}
\crefmultiformat{equation}{(#2#1#3)}{ and~(#2#1#3)}{, (#2#1#3)}{, and~(#2#1#3)}
\Crefmultiformat{equation}{(#2#1#3)}{ and~(#2#1#3)}{, (#2#1#3)}{, and~(#2#1#3)}

\title{Look Back: Reinforcement Learning with\\Counterfactual World Models}

\author{%
\parbox{\linewidth}{
Jinwoo Kim$^1$\thanks{Equal contribution $^\ddag$Equal advising $^\dag$Work partially done while visiting the Global AI Frontier Lab at NYU.}\hspace{0.5em}$^\dag$\hspace{1.4em}
Toghrul Abbasov$^1$\footnotemark[1]\hspace{1.4em}
Lucius E.J. Bynum$^2$\hspace{1em}
Aahlad Puli$^2$\hspace{1em}\\
Stephanie Milani$^2$\hspace{1em}
Hyunwoo Ryu$^3$\hspace{1em}
Seunghoon Hong$^1$\\
Kyunghyun Cho$^2$$^\ddag$\hspace{1em}
Rajesh Ranganath$^2$$^\ddag$
}\vspace{0.1cm}\\
$^1$KAIST\hspace{1em}
$^2$NYU\hspace{1em}
$^3$MIT
}

\definecolor{revcolor}{HTML}{000000}
\newcommand{\rev}[1]{{\color{revcolor}#1}}
\newcommand{\revon}{\color{revcolor}}

\iclrarxivcopy
\begin{document}

\maketitle

\begin{abstract}
Agents in complex, stochastic environments must contend with dynamics that can exceed their world models’ predictive capacity and randomness that obscure the consequences of their actions.
We hypothesize that, compared to imagining entirely new trajectories in such complex environments, it can be easier to transform observed trajectories with alternative actions.
\textbf{Counterfactual world models} condition on observations to predict alternative outcomes under the same underlying environmental randomness, using experience as a reference for predicting action-dependent changes. We present a policy gradient estimator, \textbf{REINFORCE Look Back (\ours{})}, that uses counterfactual returns as baselines to distinguish action effects from environmental luck, and prove conditions for unbiasedness and variance reduction. In settings where environmental evolution is harder to predict than action-induced changes, we show that counterfactual world models are easier to learn and support better policy learning than ordinary world models. Our findings suggest that counterfactual thinking can help agents learn effective behavior without independently reproducing the full complexity of their worlds.\footnote{Code is available at \url{https://github.com/ToghrulAbbasov19/look-back}.}
\end{abstract}

\section{Introduction}

Picture life on a planet with three suns, whose chaotic motion brings lethal heat and cold at unexpected times~\citep{liu2014threebody}. Having barely survived after taking shelter a day too late, one could evaluate alternative times to shelter by trying to imagine the skies and their outcomes~\citep{sutton1991dyna, ha2018world}. Yet realistic imagination needs predicting the motion of the three suns, a chaotic three-body system beyond anyone's control. Moreover, random variations across independently simulated skies can obscure whether sheltering earlier would actually help.

This example illustrates two challenges for learning with world models in a big world~\citep{javed2024big}. World models allow agents to evaluate alternative actions without additional environment interaction, but their usefulness depends on both predictive accuracy and the informativeness of simulated outcomes. When environmental complexity exceeds an agent's modeling capacity, errors can compound across imagined trajectories~\citep{janner2019trust,assran2025v}. Even with an accurate model, environmental randomness can dominate simulated returns, obscuring action effects and increasing the variance of policy gradient estimates~\citep{mao2018variance}.

In this paper, we study \textbf{counterfactual world models}, which take observed trajectories to predict alternative outcomes under the same underlying environmental randomness. \emph{Looking back}~\citep{fujimoto2021lookback}, one might ask: given how the suns actually moved, would taking shelter a day earlier have helped? Answering this question may be easier since the actual experience provides a reference for predicting what would change under a different decision~\citep{roese1997counterfactual}. A counterfactual model can use information about how the world unfolded and focus its predictive capacity on action-dependent changes~\citep{buesing2018woulda}. Holding environmental randomness fixed can also help distinguish action effects from luck, reducing the variance of policy gradient estimates~\citep{mesnard2020counterfactual}.

We explore two approaches to constructing counterfactual world models: (i) directly training models on paired trajectories generated under shared environmental randomness, and (ii) applying diffusion counterfactual inference~\citep{sanchez2022diffusion} to trained diffusion world models~\citep{alonso2024diffusion,valevski2025diffusion}. To leverage these models for reinforcement learning, we build on RLOO~\citep{kool2019buy} and introduce \textbf{REINFORCE Look Back (\ours{})}, a policy gradient algorithm that generates multiple counterfactual trajectories and uses their predicted returns as baselines.

Our main contributions are:
\begin{itemize}[leftmargin=*]
\item We formalize counterfactual world models and two approaches to building them, covering settings with and without access to trajectory pairs generated under shared environmental randomness.
\item We introduce \ours{}, a policy gradient algorithm based on counterfactual return baselines, and prove \rev{conditions for unbiasedness and variance reduction, and error bounds under learned models}.
\item We show that counterfactual world models are easier to learn and yield better policies than ordinary world models when action effects are more predictable than environmental evolution.
\end{itemize}
Our results suggest that looking back at past experience can reduce the predictive burden of learning in complex environments and help agents more precisely identify the consequences of their actions.

\begin{figure}[t]
    \vspace{-1em}
    \centering
    \includegraphics[width=\linewidth]{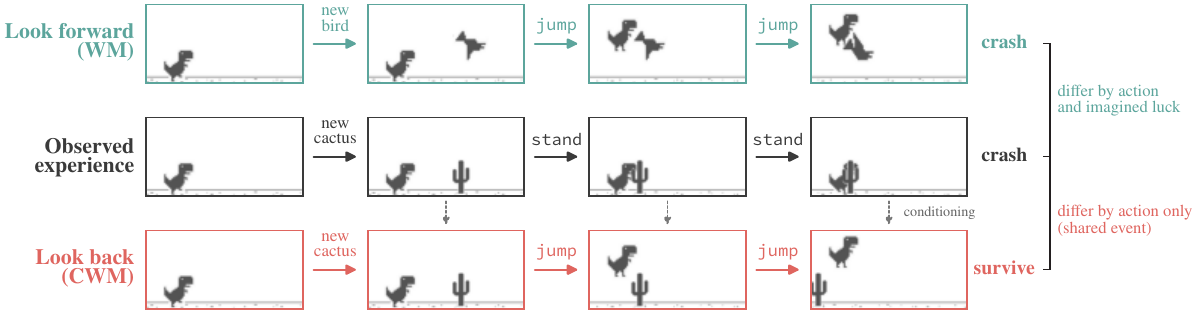}
    \caption{Overview. To evaluate an action \texttt{jump} against the observed action \texttt{stand}, a world model imagines the future anew (top), mixing the action's effect with luck, whereas a counterfactual world model edits the observed experience, isolating it (bottom).}
    \label{fig:overview}
\end{figure}

\section{Background}
\label{sec:background}

We provide an overview of the background on policy gradient methods for reinforcement learning and refer the reader to \citet{Sutton1998} for further details.

\paragraph{Setup.}
We consider Markov decision processes (MDPs) with state ${\bf x}\in\mathbb{R}^d$, action $a\in A$ in finite $A$, initial state distribution $\mu$, transition probability $p$, reward function $r$, finite horizon $n$, and discount factor $\gamma\in(0,1]$.
Given a policy $\pi_\theta$ parameterized with $\theta$, where the environment is independent of $\theta$, a trajectory $\tau\coloneqq ({\bf x}_0, a_0, ..., {\bf x}_n)$ with state ${\bf x}_i$ and action $a_i$ at step $i$ is sampled~as:
\begin{equation}\label{eq:mdp}
    {\bf x}_0\sim \mu, \qquad a_i\sim \pi_\theta(\cdot|{\bf x}_i), \qquad {\bf x}_{i+1}\sim p(\cdot|{\bf x}_i, a_i),\qquad i = 0, ..., n-1.
\end{equation}
\rev{We write $\tau\sim p_\theta$ for a trajectory sampled with \Cref{eq:mdp}, and write $\tau_{\geq i}\sim p_\theta(\cdot|{\bf x}_i)$ for suffix $({\bf x}_i,a_i,...,{\bf x}_n)$ given ${\bf x}_i$.}
We denote the reward by $r_i\coloneqq r({\bf x}_i,a_i)$, and assume the usual regularity conditions for exchanging differentiation with expectation.
Our goal is to optimize the policy to maximize the expected total return $J(\theta)\coloneqq \mathbb{E}_{\tau\sim p_\theta}[G_0]$ where $G_i\coloneqq \sum_{j=i}^{n-1}\gamma^{j-i}r_j$ is the return-to-go at step~$i$.

\paragraph{Policy gradient methods.}
The policy can be optimized by updating $\theta$ with the gradient $\nabla_\theta J(\theta)$.
Denoting the score of the policy at step $i$ as ${\bf z}_i\coloneqq \nabla_\theta \log \pi_\theta(a_i|{\bf x}_i)$, REINFORCE~\citep{williams1992simple} estimates the policy gradient using $m\geq 1$ independently sampled trajectories:
\begin{equation}\label{eq:reinforce}
    \nabla_\theta J(\theta) = \mathbb{E}\left[ \sum_{i=0}^{n-1} \gamma^i G_i {\bf z}_i \right] \approx
    \frac{1}{m}\sum_{j=1}^m\sum_{i=0}^{n-1} \gamma^i G_i^{(j)} {\bf z}_i^{(j)},\qquad \tau^{(1)},...,\tau^{(m)}\overset{\text{iid}}{\sim} p_\theta,
\end{equation}
where $G_i^{(j)}$ and ${\bf z}_i^{(j)}$ denote the return-to-go and score of each trajectory $\tau^{(j)}$ at step $i$, respectively.
While the estimation is unbiased, sampled returns can vary substantially, leading to a high variance.
A common approach to reduce the variance without biasing the estimate is to use a possibly random baseline $b_i$ that is independent of the action $a_i$ given the state ${\bf x}_i$~\citep{greensmith2004variance}:
\begin{equation}\label{eq:reinforce_with_baseline}
    \nabla_\theta J(\theta) = \mathbb{E}\left[ \sum_{i=0}^{n-1} \gamma^i (G_i - b_i) {\bf z}_i \right],\qquad b_i\perp a_i| {\bf x}_i.
\end{equation}
A standard choice is to fit a value function $b_i = \hat{V}_i({\bf x}_i)$ to approximate the expected return $\mathbb{E}[G_i|{\bf x}_i]$.
REINFORCE Leave-One-Out (RLOO)~\citep{kool2019buy,ahmadian2024back} instead samples $m\geq 2$ trajectories at each step and uses leave-one-out returns to construct an empirical baseline:
\begin{equation}\label{eq:rloo}
    \nabla_\theta J(\theta)
    \!\approx\! \frac{1}{m} \!\sum_{j=1}^m \!\sum_{i=0}^{n-1} \gamma^i (G_i^{(j)} - b_i^{(j)}){\bf z}_i^{(j)},
    \quad b_i^{(j)} \!\coloneqq\! \frac{1}{m\!-\!1} \! \sum_{\ell\neq j} G_i^{(\ell)},
    \quad \tau_{\geq i}^{(1)} ,..., \tau_{\geq i}^{(m)} \overset{\text{iid}}{\sim} p_\theta(\cdot|{\bf x}_i),
\end{equation}
where $G_i^{(j)}$ and ${\bf z}_i^{(j)}$ are return-to-go and score of each trajectory $\tau_{\geq i}^{(j)}$ at step $i$, and the trajectories at step $i$ are sampled independently from the same state ${\bf x}_i \sim p_\theta$.
To implement \Cref{eq:rloo}, one can sample a trajectory as root and then sample $m-1$ branching trajectories from every state, or a subset $B\subseteq\mathbb{R}^d$ to reduce cost~\citep{schulman2015trust}.
Either way, this requires sampling new trajectories from an observed state.
This can be done by resetting the environment, or by using a world model (WM) $p_\omega({\bf x}_{i+1}|{\bf x}_i,a_i)$~\citep{ha2018world,xiao2025world} with parameters $\omega$ learned via:
\begin{equation}\label{eq:wm_loss}
    \mathcal{L}_\mathsf{WM}(\omega) \coloneqq \sum_{i=0}^{n-1}\mathbb{E}_{({\bf x}_i,a_i,{\bf x}_{i+1})\sim p_\phi}\left[-\log p_\omega({\bf x}_{i+1}|{\bf x}_i,a_i) \right],
\end{equation}
where a reference policy $\pi_\phi$ collects data $({\bf x}_i,a_i,{\bf x}_{i+1})\sim p_\phi$.
In this way, policy gradient estimates can be improved by evaluating alternative actions using a world model.
Yet, complex and stochastic environments, or ``big worlds''~\citep{javed2024big}, remain challenging.
When environmental complexity exceeds the capacity of the world model, its predictions would suffer from errors~\citep{janner2019trust,assran2025v}.
Even with an accurate model, environmental randomness can dominate simulated returns, obscuring action effects and increasing the variance~\citep{mao2018variance}.

\section{Methods}

We define counterfactual world models (CWMs), show that they can be easier to learn than ordinary WMs when environmental complexity exceeds actions, and outline two methods for their learning.
Then, we present a policy gradient estimator REINFORCE Look Back~(\ours{}) via CWMs, separating action effects from environmental luck to reduce variance.
\rev{All proofs are in \Cref{sec:proofs}.}

\subsection{Counterfactual world models}\label{sec:counterfactual-models}

RLOO with a world model \Cref{eq:rloo} predicts branching trajectories independently from the root.
Our idea behind counterfactual world models is that conditioning on the whole root trajectory and predicting how it would become different under an alternative action can be easier, as it allows reusing observed environmental information.
We formalize this intuition using structural causal models~\citep{Pearl09a}.

\paragraph{Structural causal models (SCMs).}
Any MDP in the form of \Cref{eq:mdp} can be expressed as a collection of deterministic functions that take noise variables~\citep{buesing2018woulda}, defining an SCM:
\begin{equation}\label{eq:scm}
{\bf x}_0=\mu({\bf u}_{-1}),\qquad {\bf x}_{i+1} = f({\bf x}_i,a_i,{\bf u}_i),\qquad a_i=\pi_\theta({\bf x}_i,\boldsymbol{\epsilon}_i),\qquad i = 0, ..., n-1,
\end{equation}
where \rev{all noise variables ${\bf u}_{-1},{\bf u}_0,\ldots,{\bf u}_{n-1}$ and $\boldsymbol{\epsilon}_0,\ldots,\boldsymbol{\epsilon}_{n-1}$} are mutually independent, and we abuse notations to write the functions underlying the initial distribution and the policy by $\mu$ and $\pi_\theta$, respectively.
Through the SCM formulation, all randomness in the MDP is captured by $({\bf u}, \boldsymbol{\epsilon})$.
There could be multiple SCMs consistent with a given MDP~\citep{oberst2019counterfactual}.
\rev{This is not an issue, as arbitrarily choosing one of them} suffices for unbiasedness of policy gradient estimation (\Cref{sec:rlb}).
\rev{Without loss of generality, we can write the environmental noise ${\bf u}_i=({\bf v}_i,{\bf h}_i)$ with a \emph{visible} part ${\bf v}_i$, deducible from a transition $({\bf x}_i,a_i,{\bf x}_{i+1})$ deterministically, and a \emph{hidden} part ${\bf h}_i$, such as unobserved events (\Cref{sec:hidden}).
We assume that this decomposition is the same across $i$.}

\paragraph{Counterfactuals.}
Having expressed our MDP as an SCM, we can formalize counterfactual thinking after observing a transition $o_i \coloneqq (\hat{\bf x}_i, \hat{a}_i, \hat{\bf x}_{i+1})$.
If we took an alternative action $a_i\sim \pi_\theta(\cdot|{\bf x}_i)$, and if all other factors in the environment unfolded identically, will the outcome change?
Under the SCM, this amounts to evaluating an alternative outcome ${\bf x}_{i+1}\!=\!f({\bf x}_i,a_i,{\bf u}_i)$ where ${\bf u}_i$ is \emph{the} environmental noise that has produced the factual observation $\hat{\bf x}_{i+1}\!=\!f(\hat{\bf x}_i,\hat{a}_i,{\bf u}_i)$.
In general, ${\bf u}_i$ can be uncertain given $o_i$ \rev{(due to its hidden part ${\bf h}_i$)}, and the counterfactual transition probability is
\begin{equation}\label{eq:counterfactual}
    p({\bf x}_{i+1} | {\bf x}_i, a_i, o_i) = \int p({\bf u}_i|o_i)\cdot{\bf 1}({\bf x}_{i+1}=f({\bf x}_i,a_i,{\bf u}_i))\,{\rm d}{\bf u}_i,
\end{equation}
where ${\bf 1}(\cdot)$ is delta function.
\rev{Here, $\hat{\bf x}_i$ and ${\bf x}_i$ need not be the same, which allows us to reuse environmental noise underlying the transition $(\hat{\bf x}_i, \hat{a}_i, \hat{\bf x}_{i+1})$ to make predictions at a different state~${\bf x}_i$.}

\paragraph{Counterfactual world models (CWMs).}
Based on the SCM framework, we define a counterfactual world model as a model $p_\omega({\bf x}_{i+1} | {\bf x}_i, a_i, o_i)$ of the counterfactual probability \Cref{eq:counterfactual} with parameters $\omega$.
It is distinguished from an ordinary WM $p_\omega({\bf x}_{i+1}|{\bf x}_i,a_i)$, as it additionally conditions on factual observation $o_i =(\hat{\bf x}_i,\hat{a}_i,\hat{\bf x}_{i+1})$ and predicts alternative outcome of taking an action $a_i$ instead.
The relation between WM and CWM is roughly that between a generative model and an editing model.

\begin{figure}[t!]
    \vspace{-3.5em}
    \centering
    \providecolor{rlbcol}{HTML}{DF655F}
\providecolor{rloocol}{HTML}{5CA59C}
\providecolor{factcol}{HTML}{3A3A3A}
\begin{tikzpicture}[
  every node/.style={font=\scriptsize},
  lbl/.style={font=\scriptsize,inner sep=1pt},
  st/.style={circle,draw=#1,fill=#1!12,line width=0.7pt,minimum size=18pt,inner sep=0pt,font=\scriptsize},
  nz/.style={circle,draw=black!45,fill=black!6,line width=0.7pt,minimum size=16pt,inner sep=0pt,font=\scriptsize},
  sim/.style={-{Stealth[length=4pt]},black!50,line width=0.7pt},
  wm/.style={-{Stealth[length=4pt]},rloocol,line width=0.8pt},
  cf/.style={-{Stealth[length=4.5pt]},rlbcol,line width=1pt},
  step/.style={circle,fill=#1,inner sep=0pt,minimum size=2.4pt},
]
\def\dens#1{0.50*exp(-((#1-1.20)/0.225)^2)+0.78*exp(-((#1-1.80)/0.165)^2)+0.30*exp(-((#1-2.40)/0.21)^2)+0.12*exp(-((#1-1.50)/0.375)^2)}
\def\cfmap#1{(1.05+0.583*(#1)+0.278*(#1)^2)}
\def\cfslope#1{(0.583+0.556*(#1))}
\def\xf{1.20}
\begin{scope}
  \node[anchor=base west,font=\small,inner sep=0] at (-0.05,2.72) {};
  \fill[rloocol!18] plot[domain=0.7:2.9,samples=90,smooth] (\x,{1.55+\dens{\x}}) -- (2.9,1.55) -- (0.7,1.55) -- cycle;
  \draw[rloocol,line width=0.8pt] plot[domain=0.7:2.9,samples=90,smooth] (\x,{1.55+\dens{\x}});
  \draw[black!60,line width=0.5pt] (0,1.55) -- (4.9,1.55);
  \node[lbl,anchor=west,text=rloocol!80!black] at (2.85,2.05) {$p({\bf x}_{i+1}|\hat{\bf x}_i,\hat a_i)$};
  \fill[rloocol!18] plot[domain=0.7:2.8,samples=90,smooth] ({\cfmap{\x}},{\dens{\x}/\cfslope{\x}}) -- ({\cfmap{2.8}},0) -- ({\cfmap{0.7}},0) -- cycle;
  \draw[rloocol,line width=0.8pt] plot[domain=0.7:2.8,samples=90,smooth] ({\cfmap{\x}},{\dens{\x}/\cfslope{\x}});
  \draw[black!60,line width=0.5pt] (0,0) -- (4.9,0);
  \node[lbl,anchor=west,text=rloocol!80!black] at (0.0,0.36) {$p({\bf x}_{i+1}|{\bf x}_i,a_i)$};
  \foreach \xg in {1.80,2.40} {
    \draw[-{Stealth[length=3pt]},rlbcol!55,line width=0.5pt,shorten >=1pt] (\xg,1.55) -- ({\cfmap{\xg}},0);
    \fill[factcol!45] (\xg,1.55) circle (1.1pt);
  }
  \node[lbl,anchor=west,text=rlbcol!70] at (3.15,1.12) {other ${\bf u}$};
  \draw[cf,shorten >=1.6pt] (\xf,1.55) -- ({\cfmap{\xf}},0);
  \fill[factcol] (\xf,1.55) circle (1.8pt);
  \node[lbl,anchor=north east,text=factcol,inner sep=1.5pt] at (\xf,1.55) {$\hat{\bf x}_{i+1}$};
  \fill[rlbcol] ({\cfmap{\xf}},0) circle (1.8pt);
  \node[lbl,anchor=north,text=rlbcol!85!black] at ({\cfmap{\xf}},-0.02) {${\bf x}_{i+1}$};
  \node[lbl,anchor=east,text=rlbcol!85!black] at (1.42,0.90) {same ${\bf u}_i$};
  \matrix[anchor=west,inner sep=0pt,column sep=2pt,ampersand replacement=\&] at (0.02,-0.45) {
    \fill[rloocol!18] (0,-0.08) rectangle (0.28,0.08);
    \draw[rloocol,line width=0.8pt] (0,0.08) -- (0.28,0.08); \&
    \node[lbl] {world model}; \&[3pt]
    \draw[cf] (0,0) -- (0.28,0); \&
    \node[lbl] {counterfactual world model}; \\
  };
\end{scope}
\begin{scope}[xshift=6.25cm]
  \node[anchor=base west,font=\small,inner sep=0] at (-0.05,2.72) {};
  \node[anchor=base,font=\scriptsize\bfseries] at (1.55,2.15) {Paired};
  \node[st=factcol] (pf) at (0.35,1.55) {$\hat{\bf x}_{i+1}$};
  \node[st=rlbcol]  (pc) at (2.75,1.55) {${\bf x}_{i+1}$};
  \node[nz]         (pu) at (1.55,0.0) {${\bf u}_i$};
  \draw[sim] (pu) -- node[lbl,left=1pt,pos=0.5,align=right] {$f(\hat{\bf x}_i,\hat a_i,\cdot)$} (pf);
  \draw[sim] (pu) -- node[lbl,right=1pt,pos=0.5,align=left] {$f({\bf x}_i,a_i,\cdot)$} (pc);
  \draw[cf,dashed] (pf) -- node[lbl,above,text=rlbcol!85!black] {learn $p_\omega$} (pc);
  \begin{scope}[xshift=4.0cm]
  \node[anchor=base,font=\scriptsize\bfseries] at (1.55,2.15) {Unpaired (WM-induced)};
  \node[st=factcol] (uf) at (0.35,1.55) {$\hat{\bf x}_{i+1}$};
  \node[st=rlbcol]  (uc) at (2.75,1.55) {${\bf x}_{i+1}$};
  \node[nz]         (uu) at (1.55,0.0) {$\tilde{\bf u}_i$};
  \draw[wm] (uf) -- node[lbl,left=-1pt,pos=0.5,align=left] {invert\\$|(\hat{\bf x}_i,\hat a_i)$} (uu);
  \draw[wm] (uu) -- node[lbl,right=6pt,pos=0.5,align=left] {denoise\\$|({\bf x}_i,a_i)$} (uc);
  \foreach \t in {0.3,0.5,0.7} {
    \node[step=rloocol] at ($(uf)!\t!(uu)$) {};
    \node[step=rloocol] at ($(uu)!\t!(uc)$) {};
  }
  \draw[cf,dashed] (uf) -- node[lbl,above,text=rlbcol!85!black] {counterfactual} (uc);
  \end{scope}
\end{scope}
\end{tikzpicture}
    \caption{Counterfactual world models (CWMs). (Left) A CWM transforms the factual outcome under the same environmental noise ${\bf u}_i$, requiring no model of $p({\bf u})$. (Right) CWMs can be learned from paired outcomes with shared noise, or from unpaired data by inverting a (diffusion) WM.}
    \vspace{-0.5em}
    \label{fig:cwm}
\end{figure}
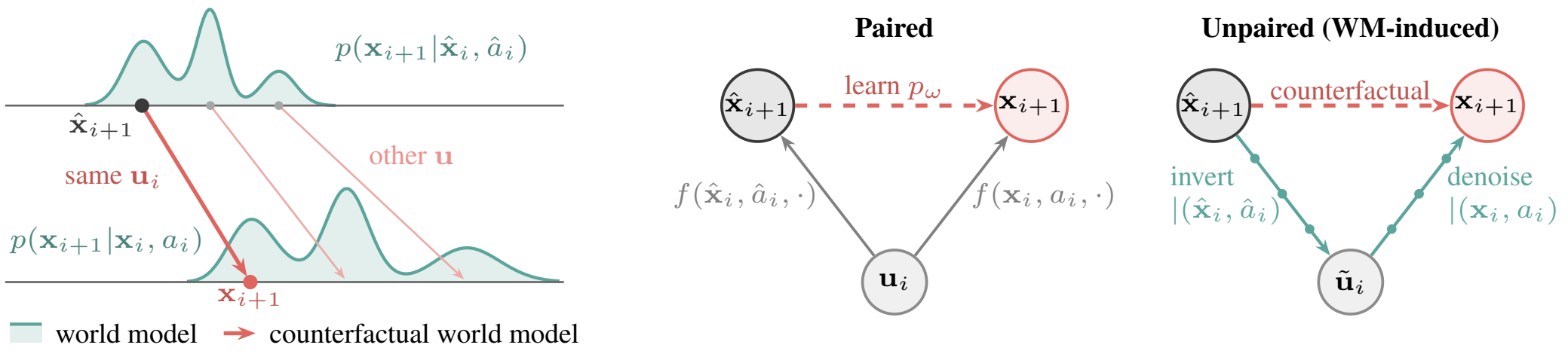

In cases where paired observations with shared environmental noise can be obtained or simulated, a CWM can be learned with ordinary maximum likelihoods, using the following loss:
\begin{equation}\label{eq:cwm_loss}
    \mathcal{L}_\mathsf{CWM}(\omega) \coloneqq \sum_{i=0}^{n-1}\mathbb{E}_{{\bf x}_i, \hat{\bf x}_i, a_i,\hat{a}_i, {\bf u}_i}\left[
    -\log p_\omega ({\bf x}_{i+1}|{\bf x}_i,a_i, o_i)
    \right],\qquad o_i = (\hat{\bf x}_i, \hat{a}_i, \hat{\bf x}_{i+1}),
\end{equation}
where ${\bf x}_i,\hat{\bf x}_i\sim p_\phi$, $a_i\sim \pi_\phi(\cdot|{\bf x}_i)$ $\hat{a}_i\sim \pi_\phi(\cdot|\hat{\bf x}_i)$ are drawn with a reference policy $\pi_\phi$, the environmental noise ${\bf u}_i$ is independently drawn from its SCM distribution, and the factual and counterfactual outcomes for training are obtained as ${\bf x}_{i+1} = f({\bf x}_i,a_i,{\bf u}_i)$ and $\hat{\bf x}_{i+1} = f(\hat{\bf x}_i, \hat{a}_i, {\bf u}_i)$, respectively.
We prove that, under mild assumptions, minimizing $\mathcal{L}_\mathsf{CWM}$ \Cref{eq:cwm_loss} recovers true counterfactuals \Cref{eq:counterfactual}:
\begin{lavenderbox}
\begin{restatable}{proposition}{propcwmloss}\label{prop:cwm_loss}
Assume that, for every step $i$,
\begin{enumerate}[label=(\roman*),leftmargin=*]
    \item The true and learned counterfactuals admit densities w.r.t.\ a reference measure at each input, whose log densities at ${\bf x}_{i+1}$ are integrable under the training distribution, and
    \item There exists $\omega_*$ s.t. $p_{\omega_*}(\cdot|{\bf x}_i,a_i,o_i)=p(\cdot|{\bf x}_i,a_i,o_i)$ a.s. under the training distribution.
\end{enumerate}
Then $\omega_*$ is a population minimizer of $\mathcal L_{\mathsf{CWM}}$, and every population minimizer recovers the true counterfactual distribution a.s. under the training distribution.
\end{restatable}
\end{lavenderbox}
A proof is in \Cref{sec:proof-cwm-loss}.
The result holds for any reference policy $\pi_\phi$.
\rev{We now prove that, for maximum likelihood training of CWMs, paired samples logarithmic in the model class suffice:}
\begin{lavenderbox}
\begin{restatable}{proposition}{propcwmupper}\label{prop:cwm-upper}
\revon
Let $\Omega$ be a finite set of parameters containing ${\omega_\ast}$ where $p_{\omega_\ast}$ is the true counterfactual, and let $\mathcal{D}$ be a set of $N$ i.i.d.\ paired transitions with shared noise.
Then every maximum likelihood CWM $p_{\hat{\omega}}$ on $\mathcal{D}$ has the following expected error in squared Hellinger distance $\mathsf{H}^2$:
\begin{equation}\label{eq:error_bound}
    \mathbb{E}_\mathcal{D}\mathbb{E}_{{\bf x}_i,a_i,o_i}\left[\mathsf{H}^2(p_{\hat{\omega}}(\cdot|{\bf x}_i,a_i,o_i), p(\cdot|{\bf x}_i,a_i,o_i))\right] \leq \frac{1+\log|\Omega|}{N}.
\end{equation}
\end{restatable}
\end{lavenderbox}
\rev{A proof is in \Cref{sec:proof-cwm-upper}.
In contrast, for ordinary WMs, the learning error of is at least that of the environmental noise distribution, leading to an irreducible worst-case lower bound on their error:
\begin{lavenderbox}
\begin{restatable}{proposition}{propwmlower}\label{prop:wm-lower}
\revon
Suppose the decomposition ${\bf u}_i=({\bf v}_i,{\bf h}_i)$ is the same for every step $i$, and the true distribution $\nu$ of the visible noise ${\bf v}_i$ is only known to belong to a family~$P$.
For a set $\mathcal{D}'$ of $2N$ i.i.d.\ transitions, every WM $p_{\hat{\omega}}$ estimated from $\mathcal{D}'$ has worst-case expected error
\begin{equation}\label{eq:wm-lower}
    \sup_{\nu\in P}\,\mathbb{E}_{\mathcal{D}'}\mathbb{E}_{{\bf x}_i,a_i}\left[\mathsf{H}^2(p_{\hat{\omega}}(\cdot|{\bf x}_i,a_i),\,p(\cdot|{\bf x}_i,a_i))\right]
    \;\geq\;\inf_{\hat\nu}\,\sup_{\nu\in P}\,\mathbb{E}_{{\bf v}^{(1)},\ldots,{\bf v}^{(M)}\sim\nu}\left[\mathsf{H}^2(\hat\nu,\nu)\right],
\end{equation}
where $M\coloneqq2N(i\!+\!1)\!+\!i$ and the infimum is over all estimates $\hat\nu$ of $\nu$ from $M$ i.i.d.\ visible noise.
\end{restatable}
\end{lavenderbox}
A proof is in \Cref{sec:proof-wm-lower}.
Whenever the RHS of \Cref{eq:wm-lower} exceeds that of \Cref{eq:error_bound}, $N$ paired data yield a more accurate CWM than $2N$ data yield any WM in the worst case.
A sufficient condition is when action effect is more predictable than the environment, in the sense of the following two cases:}
\begin{lavenderbox}
\begin{restatable}{proposition}{propcomplex}\label{prop:complex}
\revon
If the visible noise ranges over at least $34(i+1)(1+\log|\Omega|)$ values and its distribution is unrestricted, the upper bound of the CWM error is smaller than the worst-case lower bound of the WM error whenever $N\geq 34(1+\log|\Omega|)$.
\end{restatable}
\end{lavenderbox}
A proof is in \Cref{sec:proof-complex}. This intuitively amounts to complex environmental noise.
Another is:
\begin{lavenderbox}
\begin{restatable}{proposition}{propoutweighing}\label{prop:outweighing}
\revon
Suppose that, at a given step $i$,
\begin{enumerate}[label=(\roman*),leftmargin=*]
    \item It holds that $\mathbb{E}_{{\bf x}_{i+1}\sim p_\omega(\cdot|{\bf x}_i,a_i,o_i)}\|{\bf x}_{i+1}-\hat{\bf x}_{i+1}\|^2\leq A^2$ for all $\omega\in\Omega$ and all pairs of factual and counterfactual transitions, and
    \item The distribution of visible noise is unrestricted, and there are two values of visible noise whose next states, averaged over hidden noise, are at distance $\geq E$ at every state-action.
\end{enumerate}
Then, in squared $1$-Wasserstein distance $\mathsf{W}^2$, the upper bound of CWM error is smaller than the worst-case lower bound of WM error whenever $E^2>48A^2(3i+4)(1+\log|\Omega|)$.
\end{restatable}
\end{lavenderbox}
\rev{A proof is in \Cref{sec:proof-outweighing}.
This amounts to noise outweighing action effect, as $A$ upper-bounds the effect of action with fixed noise, and $E$ lower-bounds the effect of noise with fixed action.}

\vspace{-0.1em}
\paragraph{WM-induced CWMs.}
Training CWMs with \Cref{eq:cwm_loss} needs access to paired factual and counterfactual observations, which can be unavailable for realistic environments.
In these cases, we propose to train an ordinary WM $p_\omega({\bf x}_{i+1}|{\bf x}_i,a_i)$ based on state-action-conditioned diffusion models~\citep{alonso2024diffusion,valevski2025diffusion} to maximize the observation likelihood \Cref{eq:wm_loss}, then apply diffusion counterfactual inference~\citep{sanchez2022diffusion} to turn it into a CWM $p_\omega({\bf x}_{i+1} | {\bf x}_i, a_i, o_i)$.
This can be done, for each observation $o_i=(\hat{\bf x}_i,\hat{a}_i,\hat{\bf x}_{i+1})$, by inferring the latent noise $\tilde{\bf u}_i$ underlying the transition $(\hat{\bf x}_i, \hat{a}_i)\mapsto \hat{\bf x}_{i+1}$ via noise inversion~\citep{song2020denoising}, and then starting deterministic generation \rev{${\bf x}_{i+1}=f_\omega({\bf x}_i,a_i,\tilde{\bf u}_i)$} from $\tilde{\bf u}_i$ with altered conditioning $({\bf x}_i,a_i)$ to get ${\bf x}_{i+1}$ (\Cref{fig:cwm}).

While this approach requires learning a WM, its prediction during inference is now counterfactual.
\rev{We now show that those diffusion counterfactuals can be exact even when the WM itself has errors:}
\begin{lavenderbox}
\begin{restatable}{proposition}{propdiffusion}\label{prop:diffusion}
\revon
Suppose a WM $f_\omega$ uses noise $\tilde{\bf u}\!\sim\!\mathsf{N}(0,I)$, and $f_\omega({\bf x}_i,a_i,\cdot)$ is invertible.
If there is a distribution $\tilde\nu$ s.t.\ setting $\tilde{\bf u}\!\sim\!\tilde\nu$ removes the WM's error, then the induced CWM gives exact counterfactuals of an SCM consistent with the MDP, whereas the WM has error $\mathsf{H}^2(\mathsf{N}(0,I),\tilde\nu)$.
\end{restatable}
\end{lavenderbox}
\rev{
A proof is in \Cref{sec:proof-diffusion}.
Whenever $\tilde\nu \neq \mathsf{N}(0,I)$, the WM is wrong about the noise distribution, an error no learned WM can avoid (\Cref{prop:wm-lower}).
Yet this error cancels between inversion and generation for the counterfactuals, making them exact, although obtained without paired data.
We note that the above result holds for any WM invertible on its noise.
While in practice there can be errors in $f_\omega$ and its inversion, their impact is scaled at most by a constant under proper conditions:}
\begin{lavenderbox}
\begin{restatable}{proposition}{propdiffusionstability}\label{prop:diffusion-stability}
\rev{Let $f$ be an SCM consistent with the MDP such that, for some constant $\Lambda$,
\begin{equation}\label{eq:noise-lipschitz}
    \|f({\bf x},a,{\bf u})-f({\bf x},a,{\bf u}')\|\leq\Lambda\,\|f(\hat{\bf x},\hat a,{\bf u})-f(\hat{\bf x},\hat a,{\bf u}')\|,\qquad \forall ({\bf x},a),(\hat{\bf x},\hat a),{\bf u},{\bf u}'.
\end{equation}
Suppose that $\|f_\omega({\bf x},a,\tilde{\bf u})-f({\bf x},a,T(\tilde{\bf u}))\|\leq\varepsilon$ for all $({\bf x},a)$ and $\tilde{\bf u}$, for some map $T$ that does not depend on $({\bf x},a)$, and that inversion returns a latent $\tilde{\bf u}$ that reproduces the factual next state up to $\varepsilon_{\mathsf{inv}}$, i.e., $\|f_\omega(\hat{\bf x}_i,\hat a_i,\tilde{\bf u})-\hat{\bf x}_{i+1}\|\leq\varepsilon_{\mathsf{inv}}$.
Then the CWM is within $(1+\Lambda)\,\varepsilon+\Lambda\,\varepsilon_{\mathsf{inv}}$ of the true counterfactual of $f$, for every factual transition of $f$ and every $({\bf x}_i,a_i)$.}
\end{restatable}
\end{lavenderbox}
A proof is in \Cref{sec:proof-diffusion-stability}.
In summary, WM-induced CWMs allow learning an approximate SCM and leveraging its counterfactuals from only observational data. This is useful for training agents with CWMs as we describe next.

\subsection{REINFORCE Look Back}\label{sec:rlb}

\begin{figure}[t]
    \centering
    \providecolor{rlbcol}{HTML}{DF655F}
\providecolor{rloocol}{HTML}{5CA59C}
\providecolor{factcol}{HTML}{3A3A3A}
\begin{tikzpicture}[
  factual/.style={factcol,line width=1.1pt},
  branch/.style={line width=0.8pt},
  cond/.style={black!45,densely dotted,line width=0.45pt},
]
\begin{scope}[xshift=0cm]
  \node[anchor=base west,font=\small,inner sep=0] at (-0.05,1.70) {\textbf{RLOO}: independent branches};
  \node[anchor=base west,font=\scriptsize,inner sep=0,text=rloocol!80!black] at (-0.05,1.36) {$\tau^{(j)}_{\geq i}\sim p_\theta(\cdot\,|\,\hat{\bf x}_i)$};
  \draw[branch,rloocol] plot[smooth,tension=0.4] coordinates {(1.24,0.00) (1.86,0.32) (2.48,0.09) (3.10,-0.43) (3.72,-1.30) (4.34,-1.32) (4.96,-0.82)};
  \fill[rloocol] (1.86,0.32) circle (1pt);
  \fill[rloocol] (2.48,0.09) circle (1pt);
  \fill[rloocol] (3.10,-0.43) circle (1pt);
  \fill[rloocol] (3.72,-1.30) circle (1pt);
  \fill[rloocol] (4.34,-1.32) circle (1pt);
  \fill[rloocol] (4.96,-0.82) circle (1pt);
  \draw[branch,rloocol] plot[smooth,tension=0.4] coordinates {(1.24,0.00) (1.86,-0.08) (2.48,0.09) (3.10,0.33) (3.72,-0.05) (4.34,0.73) (4.96,0.44)};
  \fill[rloocol] (1.86,-0.08) circle (1pt);
  \fill[rloocol] (2.48,0.09) circle (1pt);
  \fill[rloocol] (3.10,0.33) circle (1pt);
  \fill[rloocol] (3.72,-0.05) circle (1pt);
  \fill[rloocol] (4.34,0.73) circle (1pt);
  \fill[rloocol] (4.96,0.44) circle (1pt);
  \draw[branch,rloocol] plot[smooth,tension=0.4] coordinates {(1.24,0.00) (1.86,0.12) (2.48,1.12) (3.10,1.08) (3.72,1.12) (4.34,1.05) (4.96,1.16)};
  \fill[rloocol] (1.86,0.12) circle (1pt);
  \fill[rloocol] (2.48,1.12) circle (1pt);
  \fill[rloocol] (3.10,1.08) circle (1pt);
  \fill[rloocol] (3.72,1.12) circle (1pt);
  \fill[rloocol] (4.34,1.05) circle (1pt);
  \fill[rloocol] (4.96,1.16) circle (1pt);
  \draw[factual] plot[smooth,tension=0.4] coordinates {(0.00,0.11) (0.62,0.50) (1.24,0.00) (1.86,0.30) (2.48,-0.05) (3.10,0.14) (3.72,0.39) (4.34,-0.06) (4.96,0.08)};
  \fill[factcol] (0.00,0.11) circle (1.2pt);
  \fill[factcol] (0.62,0.50) circle (1.2pt);
  \fill[factcol] (1.24,0.00) circle (1.2pt);
  \fill[factcol] (1.86,0.30) circle (1.2pt);
  \fill[factcol] (2.48,-0.05) circle (1.2pt);
  \fill[factcol] (3.10,0.14) circle (1.2pt);
  \fill[factcol] (3.72,0.39) circle (1.2pt);
  \fill[factcol] (4.34,-0.06) circle (1.2pt);
  \fill[factcol] (4.96,0.08) circle (1.2pt);
  \draw[factcol,fill=white,line width=0.8pt] (1.24,0.00) circle (2pt);
  \node[font=\scriptsize,anchor=south,inner sep=2.5pt] at (1.24,0.00) {$\hat{\bf x}_i$};
  \node[font=\scriptsize,anchor=east,inner sep=2pt,text=factcol] at (0.00,0.11) {$\hat\tau$};
  \draw[decorate,decoration={brace,amplitude=2.5pt},black,line width=0.6pt] (5.11,1.16) -- (5.11,-0.82);
  \node[font=\scriptsize,anchor=west,align=left,inner sep=0,text=black] at (5.29,0.17) {returns differ\\by action\\\emph{and} luck};
\end{scope}
\begin{scope}[xshift=7.0cm]
  \node[anchor=base west,font=\small,inner sep=0] at (-0.05,1.70) {\textbf{\ours{}}: counterfactual branches};
  \node[anchor=base west,font=\scriptsize,inner sep=0,text=rlbcol!80!black] at (-0.05,1.36) {$\tau^{(j)}_{\geq i}\sim p_\theta(\cdot\,|\,\hat{\bf x}_i,\hat\tau)$};
  \draw[cond] (1.86,0.30) -- (1.86,0.40);
  \draw[cond] (2.48,-0.05) -- (2.48,0.13);
  \draw[cond] (3.10,0.14) -- (3.10,0.32);
  \draw[cond] (3.72,0.39) -- (3.72,0.67);
  \draw[cond] (4.34,-0.06) -- (4.34,0.32);
  \draw[cond] (4.96,0.08) -- (4.96,0.55);
  \draw[cond] (1.86,0.30) -- (1.86,0.21);
  \draw[cond] (2.48,-0.05) -- (2.48,-0.15);
  \draw[cond] (3.10,0.14) -- (3.10,-0.05);
  \draw[cond] (3.72,0.39) -- (3.72,0.10);
  \draw[cond] (4.34,-0.06) -- (4.34,-0.43);
  \draw[cond] (4.96,0.08) -- (4.96,-0.29);
  \draw[cond] (1.86,0.30) -- (1.86,0.30);
  \draw[cond] (2.48,-0.05) -- (2.48,0.04);
  \draw[cond] (3.10,0.14) -- (3.10,0.23);
  \draw[cond] (3.72,0.39) -- (3.72,0.48);
  \draw[cond] (4.34,-0.06) -- (4.34,0.13);
  \draw[cond] (4.96,0.08) -- (4.96,0.27);
  \draw[branch,rlbcol] plot[smooth,tension=0.4] coordinates {(1.24,0.00) (1.86,0.40) (2.48,0.13) (3.10,0.32) (3.72,0.67) (4.34,0.32) (4.96,0.55)};
  \fill[rlbcol] (1.86,0.40) circle (1pt);
  \fill[rlbcol] (2.48,0.13) circle (1pt);
  \fill[rlbcol] (3.10,0.32) circle (1pt);
  \fill[rlbcol] (3.72,0.67) circle (1pt);
  \fill[rlbcol] (4.34,0.32) circle (1pt);
  \fill[rlbcol] (4.96,0.55) circle (1pt);
  \draw[branch,rlbcol] plot[smooth,tension=0.4] coordinates {(1.24,0.00) (1.86,0.21) (2.48,-0.15) (3.10,-0.05) (3.72,0.10) (4.34,-0.43) (4.96,-0.29)};
  \fill[rlbcol] (1.86,0.21) circle (1pt);
  \fill[rlbcol] (2.48,-0.15) circle (1pt);
  \fill[rlbcol] (3.10,-0.05) circle (1pt);
  \fill[rlbcol] (3.72,0.10) circle (1pt);
  \fill[rlbcol] (4.34,-0.43) circle (1pt);
  \fill[rlbcol] (4.96,-0.29) circle (1pt);
  \draw[branch,rlbcol] plot[smooth,tension=0.4] coordinates {(1.24,0.00) (1.86,0.30) (2.48,0.04) (3.10,0.23) (3.72,0.48) (4.34,0.13) (4.96,0.27)};
  \fill[rlbcol] (1.86,0.30) circle (1pt);
  \fill[rlbcol] (2.48,0.04) circle (1pt);
  \fill[rlbcol] (3.10,0.23) circle (1pt);
  \fill[rlbcol] (3.72,0.48) circle (1pt);
  \fill[rlbcol] (4.34,0.13) circle (1pt);
  \fill[rlbcol] (4.96,0.27) circle (1pt);
  \draw[factual] plot[smooth,tension=0.4] coordinates {(0.00,0.11) (0.62,0.50) (1.24,0.00) (1.86,0.30) (2.48,-0.05) (3.10,0.14) (3.72,0.39) (4.34,-0.06) (4.96,0.08)};
  \fill[factcol] (0.00,0.11) circle (1.2pt);
  \fill[factcol] (0.62,0.50) circle (1.2pt);
  \fill[factcol] (1.24,0.00) circle (1.2pt);
  \fill[factcol] (1.86,0.30) circle (1.2pt);
  \fill[factcol] (2.48,-0.05) circle (1.2pt);
  \fill[factcol] (3.10,0.14) circle (1.2pt);
  \fill[factcol] (3.72,0.39) circle (1.2pt);
  \fill[factcol] (4.34,-0.06) circle (1.2pt);
  \fill[factcol] (4.96,0.08) circle (1.2pt);
  \draw[factcol,fill=white,line width=0.8pt] (1.24,0.00) circle (2pt);
  \node[font=\scriptsize,anchor=south,inner sep=2.5pt] at (1.24,0.00) {$\hat{\bf x}_i$};
  \node[font=\scriptsize,anchor=east,inner sep=2pt,text=factcol] at (0.00,0.11) {$\hat\tau$};
  \draw[decorate,decoration={brace,amplitude=2.5pt},black,line width=0.6pt] (5.11,0.55) -- (5.11,-0.29);
  \node[font=\scriptsize,anchor=west,align=left,inner sep=0,text=black] at (5.29,0.13) {returns differ\\by action\\only};
\end{scope}
\end{tikzpicture}
    \caption{RLOO vs.\ \ours{}. Independent branches differ by both action and luck, whereas counterfactual branches share the factual noise and differ by action only.}
    \label{fig:rlb}
\end{figure}
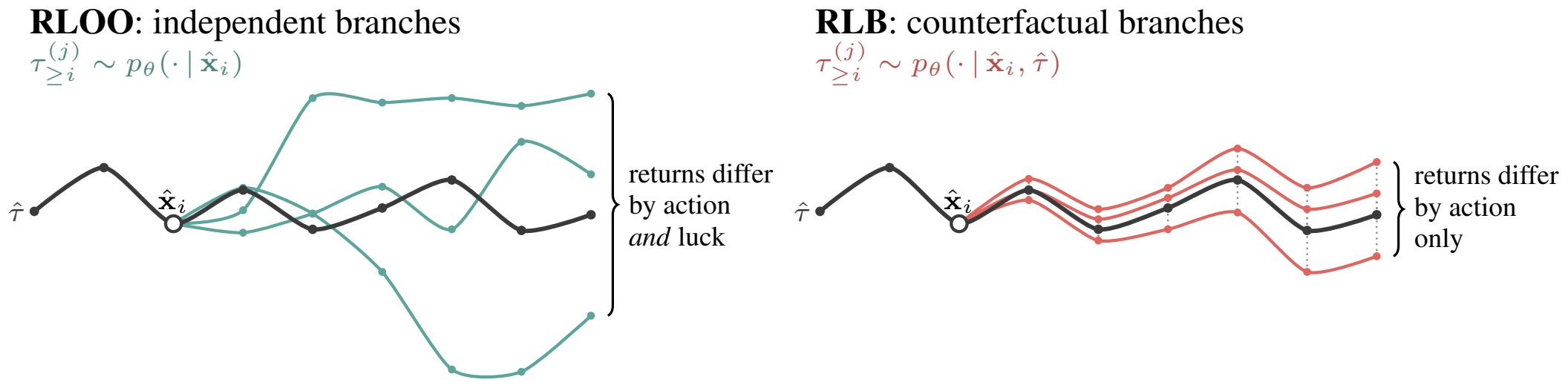

We now present our policy gradient estimator.
To motivate the method, we remark that the RLOO estimator \Cref{eq:rloo} samples the root trajectory, here denoted $\hat{\tau}$, and branching trajectories, independently:
\begin{equation}
    \tau^{(1)} = \hat{\tau} = (\hat{\bf x}_0, \hat{a}_0 ..., \hat{\bf x}_n) \sim p_\theta,
    \qquad \tau_{\geq i}^{(2)}, ..., \tau_{\geq i}^{(m)}
    \overset{\text{iid}}{\sim} p_\theta(\cdot|\hat{\bf x}_i).
\end{equation}
While modeling the branches with a world model allows evaluating alternative actions, it can obscure action effects with environmental randomness.
Our key idea is to replace the branching trajectories with counterfactuals relative to the root as the factual, thereby coupling environmental randomness.
This leads to our policy gradient estimator, \textbf{REINFORCE Look Back (\ours{})} (\Cref{fig:rlb}):
\begin{equation}\label{eq:rlb}
    \!\!\!\!\nabla_\theta J(\theta) \!\approx\! \frac{1}{m} \!\sum_{j=1}^m \!\sum_{i=0}^{n-1}\! \gamma^i (G_i^{(j)} - b_i^{(j)}){\bf z}_i^{(j)},
    \quad b_i^{(j)} \!\coloneqq\!\! \frac{1}{m\!-\!1} \! \sum_{\ell\neq j} \!G_i^{(\ell)},
    \quad \tau_{\geq i}^{(2)}, ..., \tau_{\geq i}^{(m)} \overset{\text{iid}}{\sim} p_\theta(\cdot|\hat{\bf x}_i, \hat\tau)
\end{equation}
where we define the trajectory-level counterfactual probability $p_\theta(\tau_{\geq i}|\hat{\bf x}_i, \hat\tau)$ as follows:
\begin{equation}
    p_\theta(\tau_{\geq i}|{\bf x}_i, \hat\tau) \coloneqq \prod_{k=i}^{n-1} \pi_\theta(a_k|{\bf x}_k) \cdot p({\bf x}_{k+1} | {\bf x}_k, a_k, o_k),
    \qquad o_k = (\hat{\bf x}_k,\hat{a}_k,\hat{\bf x}_{k+1}),
\end{equation}
and model $p({\bf x}_{k+1} | {\bf x}_k, a_k, o_k)$ using a CWM $p_\omega$ (\Cref{sec:counterfactual-models}).
By coupling environmental randomness across trajectories, much of the sampled return can be attributed to the specific action taken, potentially improving the policy gradient estimate.
\ours{} is unbiased under proper conditions:
\begin{lavenderbox}
\begin{restatable}{proposition}{proprlbunbiased}
\label{prop:rlb-unbiased}
Assume that all noise variables in the SCM \Cref{eq:scm} are mutually independent and have distributions independent of $\theta$.
Let $m\geq 2$ and $\hat\tau\sim p_\theta$.
At every step $i$, sample the counterfactual trajectories as in \Cref{eq:rlb}, using exact counterfactual distributions and fresh independent policy randomness.
Under the usual regularity and integrability conditions, \ours{} is unbiased.
\end{restatable}
\end{lavenderbox}
\rev{A proof is in \Cref{sec:proof-rlb-unbiased}.

Notably, \ours{} does not require identifiability of the SCM, as the choice of the SCM does not affect unbiasedness as we show next.
It affects only the variance of estimation, which we discuss shortly.}
\begin{lavenderbox}
\begin{restatable}{corollary}{coridentifiability}\label{cor:identifiability}
\rev{Under the assumptions of \Cref{prop:rlb-unbiased}, \ours{} is unbiased when using the exact counterfactuals of any other SCM of the form \Cref{eq:scm} that is consistent with the MDP.}
\end{restatable}
\end{lavenderbox}
\rev{A proof is in \Cref{sec:proof-identifiability}.
This allows (i) using a simulator of the environment for paired training of CWMs as long as it matches transition probabilities of the MDP, and (ii) using WM-induced CWMs built from unpaired data without the need to know the true SCM of the environment.

We now characterize when \ours{} reduces variance.
For simplicity, we consider one \emph{group}, the factual and its $m\!-\!1$ branches from a state $\hat{\bf x}_i$.
Let $\widehat{\bf g}_{\mathsf{\ours{}}}$ be its term in \Cref{eq:rlb} without the factor $\gamma^i$, and $\widehat{\bf g}_{\mathsf{RLOO}}$ that of \Cref{eq:rloo}.
The return of each trajectory is a function $G(\boldsymbol{\epsilon},{\bf u})$ of action $\boldsymbol{\epsilon}$ and environmental noise~${\bf u}$.
We define some quantities that measure environmental luck and its interactions with action effect:
\begin{equation}
    V({\bf u})\coloneqq\mathbb{E}_{\boldsymbol{\epsilon}}[G(\boldsymbol{\epsilon},{\bf u})],
    \qquad \sigma^2_{\mathsf{luck}}\coloneqq \operatorname{Var}_{\bf u}V({\bf u}),
    \qquad \sigma^2_{\mathsf{inter}}\coloneqq\sup_{\boldsymbol{\epsilon}}\operatorname{Var}_{\bf u}(G(\boldsymbol{\epsilon},{\bf u})-V({\bf u})).
\end{equation}
Here, $V({\bf u})$ is the return that depends on environmental noise but not on the actions.
Its variance $\sigma^2_{\mathsf{luck}}$ measures luck that no action is responsible for.
The difference $G(\boldsymbol{\epsilon},{\bf u})-V({\bf u})$ is the effect of actions relative to luck, and $\sigma^2_{\mathsf{inter}}$ measures how much this effect changes with the noise.
We show:}
\begin{lavenderbox}
\begin{restatable}{proposition}{propvariance}\label{prop:variance}
\revon
Suppose that counterfactuals are exact. Then, for every group and given $\hat{\bf x}_i$,
\begin{equation}
    \operatorname{Cov}(\widehat{\bf g}_{\mathsf{RLOO}})-\operatorname{Cov}(\widehat{\bf g}_{\mathsf{\ours{}}})\succeq\frac{(\sigma_{\mathsf{luck}}-\sigma_{\mathsf{inter}})^2-(m+4)\,\sigma^2_{\mathsf{inter}}}{m-1}\,\mathbb{E}[{\bf z}_i{\bf z}_i^{\top}],
\end{equation}
with equality if $\sigma_{\mathsf{inter}}=0$, where $A\succeq B$ means $A-B$ is PSD.
Against REINFORCE with $m$ independent trajectories and exact value baseline, the bound holds with $\frac{2\operatorname{Var}(G)-\sigma^2_{\mathsf{luck}}}{m(m-1)}\,\mathbb{E}[{\bf z}_i{\bf z}_i^{\top}]$ subtracted.
Hence \ours{} has the smaller covariance whenever $\sigma_{\mathsf{luck}}\!\geq\!(1\!+\!\sqrt{m\!+\!4})\,\sigma_{\mathsf{inter}}$ against RLOO, and whenever $\sqrt{2\!\operatorname{Var}(G)/(m\!+\!1)}$ is added to the threshold against REINFORCE.
\end{restatable}
\end{lavenderbox}
\rev{A proof is in \Cref{sec:proof-variance}.
For variance reduction, the condition on $\sigma_{\mathsf{luck}}$ asks that luck outweighs the interaction of actions and noise.
In \Cref{sec:hurt}, we prove that this condition is tight up to a constant.
For $\sigma_{\mathsf{inter}}=0$, the bound is the classical effect of common random numbers~\citep{glasserman1992some}, which we extend to noise that interacts with actions and is partly hidden.
We note that only $\sigma_{\mathsf{luck}}$ and $\sigma_{\mathsf{inter}}$ depend on the choice of SCM.
Every consistent SCM gives an unbiased estimate (\Cref{cor:identifiability}), but SCMs whose noise consistently acts as luck reduce variance more.

We now bound errors under learned models.}
\begin{lavenderbox}
\begin{restatable}{proposition}{proplearned}\label{prop:learned}
\revon
Let the branches of \ours{} be generated by a CWM, let $\eta$ be the mean absolute error of the model's expected return of a branch, and let scores be bounded by $z_{\max}$.
Then, for every group and $\hat{\bf x}_i$, the bias of \ours{} is at most $z_{\max}\eta$ in norm.
If returns of a branch and its factual differ by at most $A$, then
\begin{equation}
    \operatorname{Cov}(\widehat{\bf g})-\operatorname{Cov}(\widehat{\bf g}_{\mathsf{\ours{}}})\succeq\left(\frac{(\sigma_{\mathsf{luck}}-\sigma_{\mathsf{inter}})_+^2}{m(m-1)}-A^2\right)\,\mathbb{E}[{\bf z}_i{\bf z}_i^{\top}]
\end{equation}
for $\widehat{\bf g}$ either RLOO with any WM or REINFORCE with any state-dependent baseline, where $(\cdot)_+$ is the positive part.
Hence \ours{} has the smaller covariance whenever $\sigma_{\mathsf{luck}}\geq\sigma_{\mathsf{inter}}+mA$, and the smaller mean squared error if also $z_{\max}\eta\leq A\,(\mathbb{E}\|{\bf z}_i\|^2/m)^{1/2}$.
\end{restatable}
\end{lavenderbox}
\rev{A proof is in \Cref{sec:proof-learned}.
Here $\eta$ may refer to the error with respect to the exact counterfactuals of \emph{any} SCM consistent with the MDP, so the model need not identify a particular one.
The result shows that the bias of \ours{} is bounded by the model error $\eta$, whereas variance reduction requires a condition on environmental luck similar as in \Cref{prop:variance}.

We finally show that the bias of \ours{} for the whole estimator in \Cref{eq:rlb} decays with the number of paired transitions for CWM training, at a rate independent of the noise distribution:}
\begin{lavenderbox}
\begin{restatable}{proposition}{proprate}\label{prop:rate}
\revon
In \Cref{prop:cwm-upper,prop:learned}, suppose that every input of the CWM along the branches is at most $\kappa$ times as likely as in the training distribution.
Then, on average over the training set, every maximum likelihood CWM has $\mathbb{E}_{\hat{\bf x}_i}[\eta]\leq2A\,{(n-i)}\sqrt{2\kappa(1+\log|\Omega|)/N}$ for step $i$.
Hence \ours{} \Cref{eq:rlb} has a bias at most $A\,z_{\max}\,{n(n+1)}\sqrt{2\kappa(1+\log|\Omega|)/N}$ in norm.
\end{restatable}
\end{lavenderbox}
\rev{A proof is in \Cref{sec:proof-rate}.
Here $\kappa$ bounds the distribution shift from training data, collected with the reference policy $\pi_\phi$, to the branches of the policy $\pi_\theta$.

In \Cref{sec:branching}, we further provide an extension of the main results to sparse branching.}

\section{Related Work}

We provide an overview of related work.
\rev{An extended discussion can be found in \Cref{sec:related}.
}

\paragraph{World models.}
In model-based reinforcement learning, world models support policy learning through imagined experience~\citep{sutton1991dyna,ha2018world}. Yet, model errors arising from limited capacity can limit their utility~\citep{janner2019trust}, especially when modeling a big world~\citep{javed2024big}. Prior work has addressed this challenge by directing model capacity toward predictions needed for value estimation and planning~\citep{grimm2020value,lecun2022path}. We study a complementary approach of using observed trajectories as conditioning information when predicting the consequences of alternative actions. This can simplify prediction when environmental evolution is difficult to model but action effects are comparatively simple~\citep{buesing2018woulda}.

\paragraph{Counterfactual reinforcement learning.}
Prior work has used counterfactual reasoning~\citep{roese1997counterfactual} to improve reinforcement learning methods. \citet{buesing2018woulda} show that conditioning simulations on logged experience can improve policy evaluation and search. CCA~\citep{mesnard2020counterfactual} learns value baselines that estimate counterfactual returns in order to better distinguish action effects from environmental luck, adding an action-independence loss to encourage unbiasedness. Our approach learns imaginations using a counterfactual world model to construct policy gradient baselines, harnessing benefits of counterfactuals in both model learning and credit assignment.

\paragraph{Policy gradient baselines.}
Our policy gradient estimator builds on prior work on baselines, which can reduce variance without biasing the gradient under proper conditions~\citep{greensmith2004variance}. RLOO~\citep{kool2019buy} constructs baselines from independent trajectories from the same initial state, and world models can provide these trajectories when the environment cannot be reset~\citep{xiao2025world}. For environments with simulators, prior work has shown that constructing baselines from trajectories with shared randomness can further reduce variance~\citep{schulman2015trust,mao2018variance,bazerghi2025relative} through a common random number effect.

\paragraph{Counterfactual modeling.}
Counterfactual modeling builds upon the causal inference framework~\citep{Pearl09a}, with approaches including learning from paired observations with shared underlying randomness~\citep{baradel2019cophy} and learning a diffusion model to generate counterfactuals from noise inversion of observed data~\citep{sanchez2022diffusion}. We adapt both to world modeling, using trajectory pairs generated under different actions and shared environmental randomness, or extending diffusion counterfactuals to action-conditioned prediction. The latter fits naturally with model-based reinforcement learning, as it builds upon diffusion world models trained on interaction data~\citep{alonso2024diffusion,valevski2025diffusion} without requiring paired counterfactual supervision.

\section{Experiments}\label{sec:experiments}

Through our experiments, we aim to answer the following key questions:
\begin{enumerate}[label={\bf Q\arabic*}.,leftmargin=*]
    \item When using ground-truth counterfactuals, does \ours{} yield better policies than RLOO that uses ground-truth environmental simulations, as well as other policy gradient methods?
    \item Are counterfactual world models easier to learn, and do they yield better policies via \ours{} than ordinary world models when action effects are more predictable than environmental evolution?
    \item Can diffusion world models trained with observational data be turned into counterfactual world models and support effective policy learning through \ours{}?
\end{enumerate}

\rev{Details can be found in \Cref{sec:implementation}, and supplementary results can be found in \Cref{sec:additional}.}

\subsection{Contextual bandit}
\label{sec:bandit}

\begin{figure}[t]
    \vspace{-1em}
    \centering
    \includegraphics[width=0.95\textwidth]{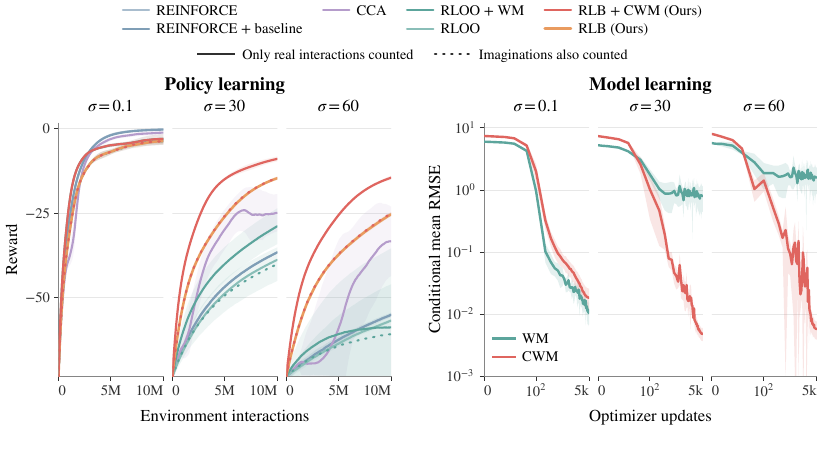}
    \vspace{-2em}
    \caption{Results for the contextual bandit problem across six independent randomized trials.
    (Left) Policy learning across various levels of environmental noise $\sigma$.
    (Right) Prediction error of ordinary and counterfactual world models over training, where \emph{for both, the minimum attainable error is zero}.}
    \label{fig:bandit}
\end{figure}

We first study the benefits of counterfactuals in a setting where the optimal policy is known.

\paragraph{Setup.}
We consider a contextual bandit problem, adapted from \citet{mesnard2020counterfactual}.
The agent receives a context ${\bf c}\in \{-10, ..., 10\}$, and based on the context, selects an action $a\in \{-10, ..., 10\}$.
Then, based on an environmental noise ${\bf u} \sim \mathsf{N}(0,\sigma^2)$ with noise level $\sigma> 0$, the agent receives a reward $r =\sigma^2 -({\bf c}-(a+{\bf u}))^2$.
This decision process can be expressed as an SCM of the form \Cref{eq:scm}:
\begin{equation}
    {\bf c}\sim \mathsf{U}(\{-10, ..., 10\}),
    \qquad {\bf x} = ({\bf c}, a+{\bf u}) = f({\bf c}, a, {\bf u}),
    \qquad a = \pi_\theta({\bf c},\boldsymbol{\epsilon}).
\end{equation}
For this problem, the optimal policy is choosing $a = {\bf c}$, which yields an average reward of zero.
Yet, the reward signal is contaminated by the environmental noise ${\bf u}$ which is uncorrelated to the action.
The higher the noise level $\sigma$ is, the more difficult it becomes to clarify the effect of an action, since high rewards could be due to a fortunate draw ${\bf u}\approx {\bf c}-a$ instead of a proper choice of action.

In this setting, we evaluate REINFORCE, REINFORCE with a learned value baseline, RLOO, and the proposed \ours{}.
We additionally test CCA~\citep{mesnard2020counterfactual}, a method that learns a hindsight value baseline and is specialized for settings with high environmental uncertainty.
For RLOO and \ours{}, we use \rev{groups of $m=4$ trajectories}, and test both branching with ground-truth transitions and counterfactuals, and also with a world model $p({\bf x}|{\bf c},a)$ or with a counterfactual world model $p({\bf x}|{\bf c},a,o)$ conditioned on factual observations $o = (\hat{\bf c},\hat{a},\hat{\bf x})$, trained with maximum likelihoods.

\paragraph{Results.}
\Cref{fig:bandit} shows the results.
As stochasticity in the environment increases, REINFORCE and RLOO degrade rapidly, presumably due to the difficulty of isolating action effects from environmental luck.
Although CCA attains better performance, it exhibits substantial variation across runs, which we conjecture is caused by optimization instabilities from having to balance policy learning and auxiliary independence regularizer.
In contrast, \ours{} with ground-truth counterfactuals can separate action effects without using an additional loss, converging faster and in a more stable manner than the baselines as noise increases (\textbf{Q1}).
Furthermore, learning a counterfactual world model saves $m-1=3$ environment interactions out of $m=4$, attaining the best result.
In addition, we find that as environmental noise becomes larger, it becomes easier to learn counterfactual world models than ordinary ones.
This is reflected in policy learning, as when compared at matched environment interactions and model imaginations, \ours{} with counterfactual model matches using true counterfactuals, whereas RLOO with world model lags behind using true simulations (\textbf{Q2}).

\subsection{Multi-agent bandit}
\label{sec:multiagent}

\begin{figure}[t]
    \vspace{-1em}
    \centering
    \includegraphics[width=0.95\textwidth]{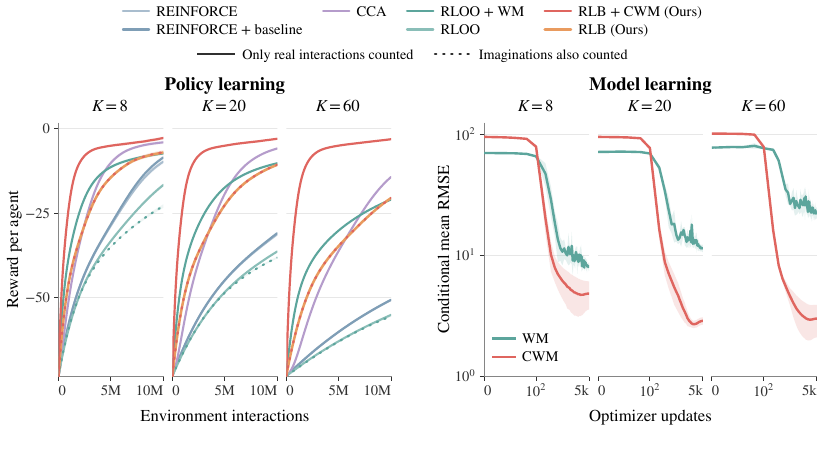}
    \vspace{-2em}
    \caption{Results for the multi-agent bandit problem across six randomized trials.
    (Left) Policy learning across various numbers $K$ of agents in the environment.
    (Right) Prediction error of ordinary and counterfactual world models over training, where \emph{for both, the minimum attainable error is zero}.}
    \label{fig:multiagent}
\end{figure}

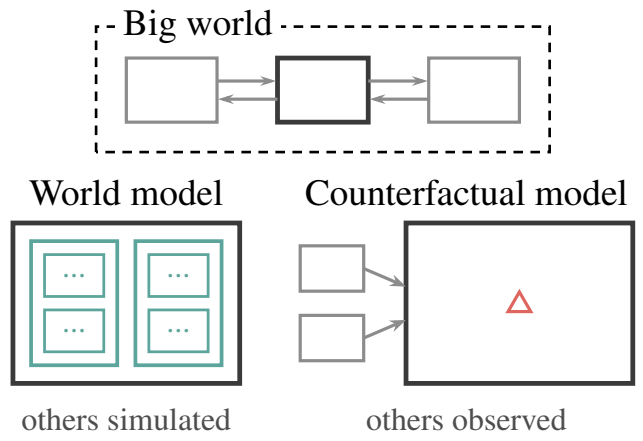
\begin{wrapfigure}{r}{0.34\textwidth}
    \vspace{-1em}
    \centering
\providecolor{rlbcol}{HTML}{DF655F}
\providecolor{rloocol}{HTML}{5CA59C}
\providecolor{factcol}{HTML}{3A3A3A}
\begin{tikzpicture}[
  agentbox/.style={draw=factcol,line width=0.9pt},
  simbox/.style={draw=rloocol},
  obsbox/.style={draw=black!45,line width=0.6pt},
  obs/.style={-{Stealth[length=3pt]},black!45,line width=0.6pt},
]
\begin{scope}
  \node[anchor=base,font=\scriptsize] at (0.75,1.18) {World model};
  \draw[agentbox] (0.000,0.000) rectangle (1.500,1.061);
  \draw[simbox,line width=0.70pt] (0.120,0.120) rectangle (0.690,0.941);
  \draw[simbox,line width=0.52pt] (0.209,0.575) rectangle (0.601,0.852);
  \node[font=\tiny,text=rloocol,inner sep=0pt] at (0.405,0.713) {...};
  \draw[simbox,line width=0.52pt] (0.209,0.209) rectangle (0.601,0.486);
  \node[font=\tiny,text=rloocol,inner sep=0pt] at (0.405,0.347) {...};
  \draw[simbox,line width=0.70pt] (0.810,0.120) rectangle (1.380,0.941);
  \draw[simbox,line width=0.52pt] (0.899,0.575) rectangle (1.291,0.852);
  \node[font=\tiny,text=rloocol,inner sep=0pt] at (1.095,0.713) {...};
  \draw[simbox,line width=0.52pt] (0.899,0.209) rectangle (1.291,0.486);
  \node[font=\tiny,text=rloocol,inner sep=0pt] at (1.095,0.347) {...};
  \node[anchor=north,font=\tiny,text=black!70] at (0.75,-0.04) {others simulated};
\end{scope}
\begin{scope}
  \node[anchor=base,font=\scriptsize] at (3.00,1.18) {Counterfactual model};
  \draw[agentbox] (2.600,0.000) rectangle (4.100,1.061);
  \draw[obsbox] (1.900,0.610) rectangle (2.320,0.910);
  \draw[obs] (2.320,0.760) -- (2.600,0.640);
  \draw[obsbox] (1.900,0.150) rectangle (2.320,0.450);
  \draw[obs] (2.320,0.300) -- (2.600,0.420);
  \draw[rlbcol,line width=0.6pt] (3.350,0.605) -- ++(-0.075,-0.13) -- ++(0.15,0) -- cycle;
  \node[anchor=north,font=\tiny,text=black!70] at (3.00,-0.04) {others observed};
\end{scope}
\begin{scope}[yshift=0.1cm]
  \draw[black,dash pattern=on 1.8pt off 1.4pt,line width=0.5pt] (0.550,1.430) rectangle (3.550,2.260);
  \node[anchor=west,font=\scriptsize,fill=white,inner sep=1.5pt] at (0.670,2.260) {Big world};
  \foreach \x in {0.750,2.750} \draw[obsbox] (\x,1.628) rectangle ++(0.600,0.424);
  \draw[agentbox] (1.750,1.628) rectangle ++(0.600,0.424);
  \begin{scope}[inter/.style={-{Stealth[length=3pt,width=1.8pt]},black!45,line width=0.6pt}]
    \draw[inter] (1.350,1.900) -- (1.742,1.900);
    \draw[inter] (1.750,1.780) -- (1.355,1.780);
    \draw[inter] (2.350,1.900) -- (2.745,1.900);
    \draw[inter] (2.750,1.780) -- (2.358,1.780);
  \end{scope}
\end{scope}
\end{tikzpicture}
    \vspace{-1em}
    \caption{An agent among others of equal capacity (top).
    A world model must simulate them (left).
    A counterfactual world model observes them (right).}
    \label{fig:bigworld}
    \vspace{-1.9em}
\end{wrapfigure}

In the prior experiment, difficulties in policy and model learning primarily arose from large environmental noise with a relatively simple distribution.
We now consider a setting where complexity of the environment exceeds that of learners.
We achieve this through a multi-agent setting where each agent faces a strictly bigger environment than itself in terms of computational capacity, because there are multiple other agents of equal capacity~\citep{javed2024big}, as illustrated in \Cref{fig:bigworld}.

\paragraph{Setup.}
For the environment, we employ $K>1$ copies of the previous contextual bandit~\citep{mesnard2020counterfactual}.
Each of the $K$ agents interacts with its own version of the environment, but rewards are now tied across agents.
The $i$-th agent receives a context ${\bf c}^i \sim \mathsf{U}(\{-10, ..., 10\})$ and selects an action $a^i\in\{-10, ..., 10\}$, where ${\bf c}$ and $a$ now denote the concatenation of all agents' contexts and actions.
Then, every agent receives reward $r = -\sum_i ({\bf c}^i - a^i)^2$.
We write this as an SCM \Cref{eq:scm}:
\begin{equation}
    {\bf x}^i = \sum_j ({\bf c}^j - a^j)^2 = f^i({\bf c}^i, a^i, {\bf u}^i),
    \qquad {\bf u}^i = ({\bf c}^j,\boldsymbol{\epsilon}^j)_{j\neq i},
    \qquad a^i = \pi_{\theta^i}({\bf c}^i, \boldsymbol{\epsilon}^i),
\end{equation}
where ${\bf x}^i$, $f^i$, $\pi_{\theta^i}$ denote the state, transition, and policy for the $i$-th agent, respectively.
The optimal policy for every agent is choosing $a^i = {\bf c}^i$, yielding an average reward of zero for all agents.
Yet, the reward signal is corrupted by other agents' contexts and actions, constituting the environmental noise ${\bf u}^i$.
As the number of interacting agents increases, it becomes harder for each agent to understand the effect of its action, because high or low rewards could be due to other agents' decisions.

We test \ours{} along with baselines from \Cref{sec:bandit}.
For RLOO and \ours{} with true simulations, there is one shared environmental interaction across the agents, and $m-1$ branching simulations for each of the $K$ agents, where we use $m = 4$.
This makes them costly, as each policy gradient estimation costs $1+K(m-1)$ environmental interactions.
If each agent has its own world model $p({\bf x}^i|{\bf c}^i,a^i)$ or a counterfactual world model $p({\bf x}^i|{\bf c}^i,a^i,o^i)$ with $o^i = (\hat{\bf c}^i,\hat{a}^i,\hat{\bf x}^i)$, this can be substantially reduced to a single interaction by replacing the rest with imaginations.
However, world model of each agent needs to predict the return contribution of all other agents, while the counterfactual one can condition on their observed contributions.
We thus expect counterfactual modeling to be easier (\Cref{fig:bigworld}).

\paragraph{Results.}
\Cref{fig:multiagent} shows the results.
As complexity of the environment increases, standard policy gradient methods degrade rapidly, and \ours{} with true counterfactuals overall outperform RLOO with true simulations (\textbf{Q1}).
While \ours{} with true counterfactuals is outperformed by CCA at matched interactions due to its high cost, we find that counterfactual world models can be learned well and easier than ordinary ones even with a large number of other agents, thereby substantially improving data efficiency of \ours{} and outperforming CCA at matched real interactions (\textbf{Q2}).

\subsection{The dinosaur game}\label{sec:exp-dino}

The preceding experiments show the benefits of counterfactual modeling, but the models therein are trained with paired data with coupled environmental noise, which might be unavailable in practice.
Therefore, we now test whether a counterfactual world model can be obtained by training a diffusion world model and employing diffusion counterfactual inference, as proposed in \Cref{sec:counterfactual-models}.

\paragraph{Setup.}
We use Chrome Dino~\citep{chromedino2014}, a platform game with a known method to train diffusion world models.
At each timestep, the agent receives 128$\times$64 image as an observation ${\bf x}_i$ and chooses an action $a_i\in \{\texttt{stand}, \texttt{jump}, \texttt{duck}\}$.
The goal is to survive by avoiding incoming obstacles including cacti on the ground and birds at varying heights, and the agent gets the reward $r_i=+1$ if it passes an obstacle and $-1$ if it collides.
Since the obstacles are generated randomly independent of the action, we can view their generation as the environmental noise ${\bf u}_i$ that obscures action effect.
We train a diffusion world model based on the protocol of \citet{valevski2025diffusion} and turn it into a counterfactual world model using diffusion counterfactual inference~\citep{sanchez2022diffusion} which was originally tested only on class-conditional image generation.
Further details are in \Cref{sec:implementation}.
We evaluate model imaginations following the criteria proposed in \citet{melistas2024benchmarking}, measuring composition, effectiveness, realism, and minimality (see \Cref{sec:implementation} for details).
We use $m=4$ for \ours{} and RLOO training, and evaluate each learned policy by running 100 test episodes and measuring collision rate per obstacle and mean survival time before the first collision.

\begin{figure}[t]
\vspace{-1em}
\centering
\includegraphics[width=\linewidth]{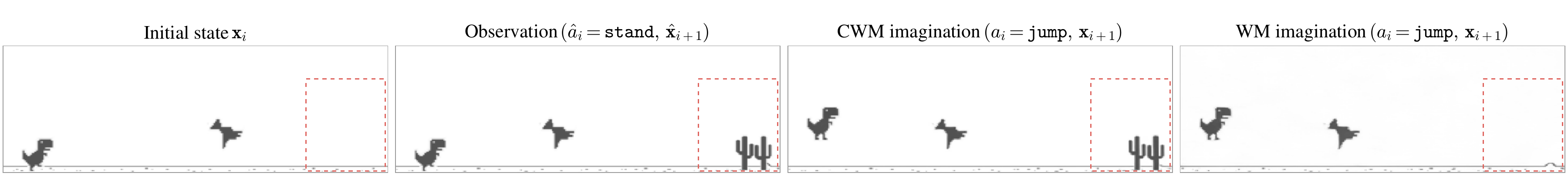}
\caption{Visualization of an environmental transition under action \texttt{stand}, a counterfactual world model imagination under action \texttt{jump}, and a world model imagination under action \texttt{jump}.}
\label{fig:dino:cf}
\end{figure}

\begin{table}[t]
\vspace{-1em}
\centering
\footnotesize
\begin{minipage}[t]{0.46\linewidth}
\centering
\caption{Model imaginations assessed following \citet{melistas2024benchmarking}.}
\label{tab:dino:cfquality}
\fitwidth{%
\begin{tabular}{lll}
\toprule
Criterion& CWM & WM \\
\midrule
Composition ($\ell_1$) $\downarrow$
    & \textbf{0.37} & 1.89 \\
Effectiveness (macro-F1) $\uparrow$
    & \textbf{0.999} & 0.993 \\
Realism (FID) $\downarrow$ & \textbf{15.94} & 37.72 \\
Minimality (CLD) $\downarrow$ & \textbf{0.296} & 0.327 \\
Minimality (obstacle $\ell_1$) $\downarrow$ & \textbf{2.34} & 9.57 \\
\bottomrule
\end{tabular}}%
\end{minipage}\hfill
\begin{minipage}[t]{0.5\linewidth}
\centering
\caption{Results for policy learning across \rev{fifteen} independent training runs.
}
\label{tab:dino:policy}
\fitwidth{%
\begin{tabular}{lll}
\toprule
Method & Collision (\%) $\downarrow$ & Survival $\uparrow$ \\
\midrule
REINFORCE & \rev{7.57 $\pm$ 3.47} & \rev{72.4 $\pm$ 18.6} \\
REINFORCE + baseline & \rev{9.53 $\pm$ 14.65} & \rev{89.7 $\pm$ 31.7} \\
\midrule
RLOO & \rev{7.31 $\pm$ 14.21} & \rev{103.6 $\pm$ 29.0} \\
\ours{} (Ours) & \textbf{\rev{3.29 $\pm$ 0.36}} & \textbf{\rev{113.9 $\pm$ 7.9}} \\
\midrule
RLOO + WM & \rev{7.89 $\pm$ 14.10} & \rev{97.5 $\pm$ 28.2} \\
\ours{} + CWM (Ours) & \textbf{\rev{3.77 $\pm$ 1.15}} & \textbf{\rev{109.0 $\pm$ 18.9}} \\
\bottomrule
\end{tabular}}%
\end{minipage}
\vspace{-1em}
\end{table}

\paragraph{Results.}
\Cref{fig:dino:cf} and \Cref{tab:dino:cfquality} demonstrate the quality of counterfactual world model imaginations. In addition to CLD, we measure minimality by the pixel $\ell_1$ error between
factual and imagined obstacles within the red regions in
\Cref{fig:dino:cf}.
We find that the counterfactual predictions preserve environmental randomness given in the factual observation while incorporating the effect of an alternative action, thereby attaining good quantitative metrics of being a proper counterfactual.
This supports that diffusion world models trained with observational data can be turned into counterfactual world models (\textbf{Q3}).
\Cref{tab:dino:policy} shows the results for downstream policy learning.
Using either true simulations or model imaginations, \ours{} overall outperforms other methods, supporting that diffusion counterfactual modeling supports effective policy learning (\textbf{Q1}--\textbf{Q3}).

\section{Conclusion}

We studied counterfactual world models, which predict how observations would change under alternative actions.
REINFORCE Look Back (\ours{}) uses their predicted returns as baselines to separate action effects from environmental luck.
It is unbiased under exact counterfactuals\rev{, has a bias bounded by the model error otherwise,} and can reduce variance when environmental noise dominates.
When action effects are more predictable than the environment, counterfactual world models are easier to learn and yield better policies than ordinary ones.
Future work includes larger diffusion models and connecting to GRPO-style methods.
Our results suggest that agents in big worlds may learn by looking back at what happened, without modeling the full complexity of their environments.

\section*{Acknowledgment}
This work is supported in part by the National Research Foundation of Korea (RS-2024-00351212) and the Institute of Information \& Communications Technology Planning \& Evaluation (IITP) (RS-202400509279, RS-2026-25518317, and RS-2019-II190075), funded by the Korean Government (MSIT).

\bibliography{iclr2027_conference}
\bibliographystyle{iclr2027_conference}

\clearpage
\appendix

\section{Theoretical results}\label{sec:proofs}

\subsection{Assumptions}\label{sec:assumptions}

\rev{We first summarize the assumptions of our results.
\Cref{prop:cwm_loss,prop:cwm-upper} assume a model class that contains the true counterfactual distribution, finite in \Cref{prop:cwm-upper} for simplicity, and do not restrict the form of the noise or the reference policy.
\Cref{prop:complex,prop:outweighing} take the worst case over the distribution of the visible noise, so they concern environments whose noise distribution is not known in advance.
\Cref{prop:diffusion} assumes exact inversion and a WM that is exact up to its noise distribution, which \Cref{prop:diffusion-stability} relaxes under \Cref{eq:noise-lipschitz}, as holds for additive noise with $\Lambda=1$.
\Cref{prop:rlb-unbiased,prop:variance} assume exact counterfactuals and hold for any return and horizon, also with hidden noise.
\Cref{prop:learned,prop:rate} instead assume bounded scores, a bound $A$ on the difference between the returns of a branch and its factual, and for the rate, coverage of the branches by the training pairs, which is comparably more restrictive (see \Cref{sec:proof-rate}).}

\subsection{Remarks on environmental noise}\label{sec:hidden}

\rev{The noise in the SCM \Cref{eq:scm} is indexed by time and independent across steps and of the action noise, which is a representation that every MDP admits~\citep{buesing2018woulda}.
A branch reuses the noise of the factual at the same step, even after its state has left that of the factual.
This does not alter the MDP, as every branch follows it by itself (\Cref{lem:posterior}) and only its joint distribution with the factual is coupled, as with common random numbers~\citep{glasserman1992some}.
Which consistent SCM is used affects the variance of \ours{} but not its unbiasedness (\Cref{cor:identifiability}).}

\rev{The split into visible and hidden parts loses no generality, as noise can be re-expressed freely~\citep{balke1994counterfactual}.
A function of the noise is visible if and only if it agrees on any two noise values that lead to the same next state for some state and action, so there is a largest visible part, possibly trivial.
All noise is visible when the next state records it, as for additive, multiplicative or other bijective mechanisms~\citep{nasr2023counterfactual} and for exogenous processes in the state, such as arrivals, obstacles, or the actions of other agents.
Noise is hidden when the factual transition leaves the outcome of other actions uncertain, as for randomness specific to the action taken, such as a gamble that was not played, or when the environment is only partly observed.
We treat partial observability only in this sense, as unobserved variables that persist over time violate the independence of the noise across steps in \Cref{eq:scm} and may confound actions and outcomes, which we leave for future work.}

\subsection{Proof of Proposition \ref{prop:cwm_loss}}\label{sec:proof-cwm-loss}

\begin{lavenderbox}
\propcwmloss*
\end{lavenderbox}
\begin{proof}
\rev{By construction of the training distribution in \Cref{eq:cwm_loss},} ${\bf u}_i$ is independent of $({\bf x}_i,\hat{\bf x}_i,a_i,\hat a_i)$ before the two next states are generated. Conditional on the factual state and action, the factual next state depends on ${\bf u}_i$ through $\hat{\bf x}_{i+1}=f(\hat{\bf x}_i,\hat a_i,{\bf u}_i)$.
Consequently, after conditioning on $o_i=(\hat{\bf x}_i,\hat a_i,\hat{\bf x}_{i+1})$, the queried state and action provide no additional information about ${\bf u}_i$\rev{, that is, $p({\bf u}_i|{\bf x}_i,a_i,o_i)=p({\bf u}_i|o_i)$.}
Since ${\bf x}_{i+1}=f({\bf x}_i,a_i,{\bf u}_i)$, its conditional distribution given $({\bf x}_i,a_i,o_i)$ is the distribution obtained by applying $f({\bf x}_i,a_i,\cdot)$ to ${\bf u}_i\sim p(\cdot| o_i)$, which is the counterfactual distribution $p(\cdot|{\bf x}_i,a_i,o_i)$ \Cref{eq:counterfactual}.

Integrability and \rev{the existence of $\omega_*$} allow us to subtract the expected loss at $\omega_*$ from \rev{that at $\omega$, and taking the expectation of ${\bf x}_{i+1}$ given $({\bf x}_i,a_i,o_i)$ first gives}
\begin{equation*}
    \mathcal L_{\mathsf{CWM}}(\omega) -\mathcal L_{\mathsf{CWM}}(\omega_*)
    =\sum_{i=0}^{n-1}\mathbb{E}\!\left[\log\frac{p({\bf x}_{i+1}|{\bf x}_i,a_i,o_i)}{p_\omega({\bf x}_{i+1}|{\bf x}_i,a_i,o_i)}\right]=\sum_{i=0}^{n-1}\mathbb{E}\!\left[\mathsf{KL}\!\left(p(\cdot|{\bf x}_i,a_i,o_i)\,\middle\|\,p_\omega(\cdot|{\bf x}_i,a_i,o_i)\right)\right].
\end{equation*}
Every term on the right-hand side is nonnegative, so $\omega_*$ is a population minimizer.
For any other population minimizer, the left-hand side is zero.
Since the sum has finitely many nonnegative terms, each conditional KL divergence must be zero almost surely under the corresponding training distribution\rev{, which gives $p_\omega(\cdot|{\bf x}_i,a_i,o_i)=p(\cdot|{\bf x}_i,a_i,o_i)$ almost surely, for every $i$.}
\end{proof}

\subsection{Proof of Proposition \ref{prop:cwm-upper}}\label{sec:proof-cwm-upper}

\rev{We show a useful lemma.}

\begin{lavenderbox}
\begin{restatable}{lemma}{lemcoupling}\label{lem:coupling}
\revon
A pair of transitions following the MDP in \Cref{eq:mdp} with shared environmental noise \Cref{eq:scm} has the same joint distribution as a factual transition and its counterfactual transition \Cref{eq:counterfactual}.
\end{restatable}
\end{lavenderbox}
\begin{proof}
\rev{Let the pair of transitions be $\hat{\bf x}_{i+1}=f(\hat{\bf x}_i,\hat a_i,{\bf u}_i)$ and ${\bf x}_{i+1}=f({\bf x}_i,a_i,{\bf u}_i)$ with shared noise ${\bf u}_i$, and write $o_i=(\hat{\bf x}_i,\hat a_i,\hat{\bf x}_{i+1})$.
By the definition of the SCM \Cref{eq:scm}, the noise ${\bf u}_i$ is independent of the two state-action pairs.
Thus, given the state-actions, the joint density of the two next states is
\begin{align*}
    p(\hat{\bf x}_{i+1},{\bf x}_{i+1}|\hat{\bf x}_i,\hat a_i,{\bf x}_i, a_i)
    &=\int p({\bf u}_i)\cdot{\bf 1}(\hat{\bf x}_{i+1}=f(\hat{\bf x}_i,\hat a_i,{\bf u}_i))\cdot{\bf 1}({\bf x}_{i+1}=f({\bf x}_i,a_i,{\bf u}_i))\,{\rm d}{\bf u}_i\\
    &=\int p({\bf u}_i)\cdot p(\hat{\bf x}_{i+1}|{\bf u}_i,\hat{\bf x}_i,\hat a_i) \cdot{\bf 1}({\bf x}_{i+1}=f({\bf x}_i,a_i,{\bf u}_i))\,{\rm d}{\bf u}_i\\
    &=p(\hat{\bf x}_{i+1}|\hat{\bf x}_i,\hat a_i)\int p({\bf u}_i|o_i)\cdot{\bf 1}({\bf x}_{i+1}=f({\bf x}_i,a_i,{\bf u}_i))\,{\rm d}{\bf u}_i\\
    &=p(\hat{\bf x}_{i+1}|\hat{\bf x}_i,\hat a_i) \cdot p({\bf x}_{i+1}|{\bf x}_i, a_i,o_i),
\end{align*}
where the third equality is Bayes' rule for $p({\bf u}_i|o_i)$.
The factor $p(\hat{\bf x}_{i+1}|\hat{\bf x}_i,\hat a_i)$ is the distribution of a factual transition, and $p({\bf x}_{i+1}|{\bf x}_i, a_i,o_i)$ is the counterfactual transition probability \Cref{eq:counterfactual}.}
\end{proof}

\rev{We now prove \Cref{prop:cwm-upper} by a classical argument for maximum likelihood over a finite class of models that contains the truth~\citep[see][also for infinite classes]{geer2000empirical}.}

\begin{lavenderbox}
\propcwmupper*
\end{lavenderbox}

\begin{proof}
\rev{
Denote the squared Hellinger distance between two distributions with densities $\nu$ and $\rho$ by
\begin{equation}
    \mathsf{H}^2(\nu, \rho) = 1-\int\sqrt{\nu({\bf x})\,\rho({\bf x})}\,{\rm d}{\bf x} =\frac12 \int\left(\sqrt{\nu({\bf x})} - \sqrt{\rho({\bf x})}\right)^2\,{\rm d}{\bf x} \qquad \in \qquad [0,1].
\end{equation}
By \Cref{lem:coupling}, a pair of transitions with shared environmental noise consists of a factual transition $o_i$ and a counterfactual transition $({\bf x}_i,a_i,{\bf x}_{i+1})$.
For each $\omega\in\Omega$, let
\begin{equation}\label{eq:error}
    \mathsf{error}(\omega) \coloneqq \mathbb{E}_{{\bf x}_i,a_i,o_i}
    \!\left[\mathsf{H}^2(p_\omega(\cdot|{\bf x}_i,a_i,o_i), p(\cdot|{\bf x}_i,a_i,o_i))\right]
\end{equation}
be its expected error in terms of squared Hellinger distance on average over $({\bf x}_i,a_i,o_i)$, where densities are taken with respect to a common reference measure for the models in $\{p_\omega:\omega\in\Omega\}$ at $({\bf x}_i,a_i,o_i)$, e.g., their sum.
We start with the fact that
\begin{equation}
    \mathbb{E}_{{\bf x}_i,a_i,o_i,{\bf x}_{i+1}}\left[\sqrt{\frac{p_\omega({\bf x}_{i+1}|{\bf x}_i,a_i,o_i)}{p({\bf x}_{i+1}|{\bf x}_i,a_i,o_i)}}\,\right]=1-\mathsf{error}(\omega)\leq e^{-\mathsf{error}(\omega)}.
\end{equation}
Let us denote the training data by $\mathcal{D}\coloneqq({\bf x}_i^{(l)},a_i^{(l)},o_i^{(l)},{\bf x}_{i+1}^{(l)})_{l=1}^N$.
By independence, they satisfy
\begin{equation*}
    \mathbb{E}_\mathcal{D}\left[\prod_{l=1}^N\sqrt{\frac{p_\omega({\bf x}_{i+1}^{(l)}|{\bf x}_i^{(l)},a_i^{(l)},o_i^{(l)})}{p({\bf x}_{i+1}^{(l)}|{\bf x}_i^{(l)},a_i^{(l)},o_i^{(l)})}}\,\right]\leq e^{-N\mathsf{error}(\omega)}.
\end{equation*}
By Markov's inequality, the probability that any given model $p_\omega$ assigns equal or higher likelihood on training data than the true counterfactual distribution $p$ is at most $e^{-N\mathsf{error}(\omega)}$.
This is smaller than $e^{-Nt}$ for any $t < \mathsf{error}(\omega)$.
Let us denote the set of parameters with model error larger than $t$ by $\Omega_t\coloneqq \{\omega\in\Omega: \mathsf{error}(\omega)>t\}$.
By union bound, the probability that $p_\omega$ for some $\omega\in\Omega_t$ attains at least the likelihood of $p$ is at most $|\Omega|e^{-tN}$.
If no model in $\Omega_t$ matches or beats $p$, every maximum likelihood model $p_{\hat{\omega}}$ satisfies $\mathsf{error}(\hat{\omega})\leq t$, because $p\in\{p_\omega:\omega\in\Omega\}$.
The largest value $\mathsf{error}_{\max}$ of $\mathsf{error}(\hat{\omega})$ among the maximum likelihood models therefore satisfies $\mathbb{P}(\mathsf{error}_{\max}>t)\leq\min\{1,|\Omega|e^{-tN}\}$ for all $t\geq0$.
Hence
\begin{equation}
\begin{aligned}
    \mathbb{E}_\mathcal{D}\big[\mathsf{error}_{\max}\big] &= \int_0^\infty \mathbb{P}\big(\mathsf{error}_{\max}>t\big){\rm d}t
    \leq\int_0^\infty\!\!\min\big\{1,|\Omega|e^{-tN}\big\}\,{\rm d}t \\
    &=\int_0^{\frac{\log|\Omega|}{N}}1\,{\rm d}t + \int_{\frac{\log|\Omega|}{N}}^\infty |\Omega|e^{-tN}\,{\rm d}t
    =\frac{1+\log|\Omega|}{N},
\end{aligned}
\end{equation}
which shows \Cref{eq:error_bound}.
}
\end{proof}

\subsection{Proof of Proposition~\ref{prop:wm-lower}}\label{sec:proof-wm-lower}

\rev{Let us write the environmental noise ${\bf u}_i=({\bf v}_i,{\bf h}_i)$ with a visible part ${\bf v}_i$ and a hidden part ${\bf h}_i$, where the visible noise is a function of the transition $({\bf x}_i,a_i,f({\bf x}_i,a_i,{\bf u}_i))$ for all ${\bf x}_i,a_i,{\bf u}_i$, which can be done without loss of generality (\Cref{sec:hidden}).
We prove that sample complexity of ordinary WMs depends on the distribution of the visible noise, leading to a worst-case lower bound on the $\mathsf{H}^2$ error.}
\propwmlower*
\begin{proof}
\rev{
The idea is that a world model contains an estimate of the noise distribution, because the visible noise can be read off its predictions.
If the model were more accurate than the right side of \Cref{eq:wm-lower}, this would give an estimate of $\nu$ that is better than the best possible one.

We write $p_\nu$ for $p$ \Cref{eq:mdp} when the visible noise has distribution $\nu$ at every step.
Throughout, $f$ and the distribution of the hidden noise given the visible noise are fixed, and only $\nu$ varies in $P$.

Let us construct an estimate $\tilde\nu$ of $\nu$ from $M$ i.i.d.\ samples of the visible noise such that
\begin{equation}\label{eq:reduction}
    \mathbb{E}\left[\mathsf{H}^2(\tilde\nu,\nu)\right]
    \leq\mathbb{E}_{\mathcal{D}'}\mathbb{E}_{{\bf x}_i,a_i}\left[\mathsf{H}^2\big(p_{\hat\omega}(\cdot|{\bf x}_i,a_i),\,p_\nu(\cdot|{\bf x}_i,a_i)\big)\right]
    \qquad\text{for every }\nu\in P.
\end{equation}
Taking the supremum over $\nu\in P$ on both sides then shows \Cref{eq:wm-lower}, because the infimum on its right side is at most the worst-case expected error of the particular estimate $\tilde\nu$.

For the construction, we start with the fact that since the visible noise is a function of every transition, there is a map $g$ with ${\bf v}_i=g({\bf x}_i,a_i,{\bf x}_{i+1})$.
For a state-action $({\bf x},a)$, let $\hat\nu_{{\bf x},a}$ be the distribution of $g({\bf x},a,{\bf y})$ for a next state ${\bf y}\sim p_{\hat\omega}(\cdot|{\bf x},a)$ predicted by the world model, where $g$ is extended arbitrarily to next states that cannot occur.
For a next state ${\bf y}\sim p_\nu(\cdot|{\bf x},a)$ of the environment, $g({\bf x},a,{\bf y})$ is the visible noise and has distribution $\nu$.
Applying the same map to two distributions does not increase their squared Hellinger distance, as a deterministic map can only merge outcomes, and $\int\!\sqrt{p\,q}$ does not decrease when outcomes are merged, by the Cauchy--Schwarz inequality.
Hence
\begin{equation}\label{eq:read-off}
    \mathsf{H}^2\big(p_{\hat\omega}(\cdot|{\bf x},a),\,p_\nu(\cdot|{\bf x},a)\big)\geq \mathsf{H}^2\big(\hat\nu_{{\bf x},a},\,\nu\big)\qquad\text{for all }({\bf x},a)\text{ and }\nu\in P.
\end{equation}

The estimate $\hat{\nu}_{{\bf x},a}$ is computed from a world model, and hence from a dataset of transitions, whereas the right side of \Cref{eq:wm-lower} concerns estimates computed from samples of the visible noise.
To compare the two, we generate the data from such samples.
Given $M = 2N(i+1)+i$ i.i.d.\ samples of the visible noise, the below procedure, which does not use $\nu$, generates a dataset $\tilde{\mathcal{D}}'$ and a state-action $(\tilde{\bf x}_i,\tilde a_i)$ with the same joint distribution as $\mathcal{D}'\coloneqq({\bf x}_i^{(l)},a_i^{(l)},{\bf x}_{i+1}^{(l)})_{l=1}^{2N}$ and a new state-action $({\bf x}_i,a_i)$:
\begin{enumerate}[leftmargin=*]
    \item For each sample, draw a hidden part from the distribution of the hidden noise given the visible noise. This gives $M$ values of the environmental noise.
    \item For each of $2N$ trajectories, draw the initial state and the actions as in \Cref{eq:scm}, compute $i+1$ transitions with $f$ from $i+1$ unused values of the noise, and add the last transition to $\tilde{\mathcal{D}}'$.
    \item For one more trajectory, compute $i$ transitions in the same way from the remaining $i$ values of the noise, and let $(\tilde{\bf x}_i,\tilde a_i)$ be its last state-action.
\end{enumerate}

Let $\tilde\nu\coloneqq\hat\nu_{\tilde{\bf x}_i,\tilde a_i}$ be the noise distribution read off at $(\tilde{\bf x}_i,\tilde a_i)$ from the world model $p_{\hat\omega}$ computed from $\tilde{\mathcal{D}}'$.
It is an estimate of $\nu$ computed from the $M$ samples, and
\begin{align*}
    \mathbb{E}\left[\mathsf{H}^2(\tilde\nu,\nu)\right]
    &\leq\mathbb{E}_{\tilde{\mathcal{D}}'}\mathbb{E}_{\tilde{\bf x}_i,\tilde a_i}\left[\mathsf{H}^2\big(p_{\hat\omega}(\cdot|\tilde{\bf x}_i,\tilde a_i),\,p_\nu(\cdot|\tilde{\bf x}_i,\tilde a_i)\big)\right]\\
    &=\mathbb{E}_{\mathcal{D}'}\mathbb{E}_{{\bf x}_i,a_i}\left[\mathsf{H}^2\big(p_{\hat\omega}(\cdot|{\bf x}_i,a_i),\,p_\nu(\cdot|{\bf x}_i,a_i)\big)\right],
\end{align*}
where the inequality is \Cref{eq:read-off}, and the equality holds because the generated data have the same distribution as the real data.
This shows \Cref{eq:reduction}.

Finally, $\tilde\nu$ depends on the randomness of the procedure in addition to the samples, whereas the infimum in \Cref{eq:wm-lower} is over estimates computed from the samples alone.
This makes no difference, as the average of $\tilde\nu$ over this randomness is computed from the samples alone, and its error is not larger, because the squared Hellinger distance is convex in the estimate.}
\end{proof}

\subsection{Proof of Proposition \ref{prop:complex}}\label{sec:proof-complex}

\rev{We show a useful lemma.}

\begin{lavenderbox}
\begin{lemma}\label{lem:worst}
\rev{Every estimate of a distribution $\nu$ on $S$ values from $M$ independent samples has, for the worst-case distribution, an expected $\mathsf{H}^2$ error of at least $(S-3)/(10(M+S))$.}
\end{lemma}
\end{lavenderbox}
\begin{proof}
\rev{To construct the worst case, let the distribution $\nu$ be drawn uniformly from the simplex, and let $M_j$ be the number of samples equal to the $j$-th value.
Given the samples, $\nu$ has the Dirichlet distribution with parameters $\beta_j=1+M_j$, which sum to $M+S$.
For any estimate $\hat\nu$, the expected $\mathsf{H}^2$ error given the samples is
\begin{equation*}
    1-\sum_{j=1}^S\sqrt{\hat\nu(j)}\;\mathbb{E}\big[\sqrt{\nu(j)}\big]\;\geq\;1-\Big(\sum_{j=1}^S\mathbb{E}\big[\sqrt{\nu(j)}\big]^2\Big)^{1/2}
\end{equation*}
by the Cauchy--Schwarz inequality, where the expectations are under this Dirichlet distribution.
It is the distribution of $(Y_1,\ldots,Y_S)/Y$ for independent Gamma variables $Y_j$ with shapes $\beta_j$ and their sum $Y$, which is a Gamma variable with shape $M+S$ and independent of the ratios, so that $\mathbb{E}[\sqrt{\nu(j)}]=\mathbb{E}[\sqrt{Y_j}]/\mathbb{E}[\sqrt{Y}]$.
A Gamma variable $Y$ with shape $\beta$ has mean and variance $\beta$ and satisfies $\mathbb{E}[Y^{3/2}]=(\beta+\frac12)\,\mathbb{E}[\sqrt{Y}]$.
Thus $\operatorname{Cov}(Y,\sqrt Y)=\frac12\mathbb{E}[\sqrt Y]$, and the Cauchy--Schwarz inequality gives $\frac14\mathbb{E}[\sqrt Y]^2\leq\beta\,(\beta-\mathbb{E}[\sqrt Y]^2)$, i.e., $\mathbb{E}[\sqrt Y]^2\leq\beta-\beta/(4\beta+1)$, which is at most $\beta-\frac15$ if $\beta\geq1$.
It also gives $\beta^2=\mathbb{E}[Y^{1/4}Y^{3/4}]^2\leq(\beta+\frac12)\,\mathbb{E}[\sqrt Y]^2$.
Applying the first bound to each $Y_j$ and the second to their sum,
\begin{equation*}
    \sum_{j=1}^S\mathbb{E}\big[\sqrt{\nu(j)}\big]^2\leq\Big(M+S-\frac S5\Big)\,\frac{M+S+\frac12}{(M+S)^2}\leq1-\frac{2S-5}{10(M+S)}.
\end{equation*}
Since $1-\sqrt{1-s}\geq s/2$, the expected error given the samples is at least $(2S-5)/(20(M+S))\geq(S-3)/(10(M+S))$, whatever the samples and the estimate.
The worst case over $\nu$ is at least the average.
}
\end{proof}

\rev{We now prove \Cref{prop:complex}.}

\begin{lavenderbox}
\propcomplex*
\end{lavenderbox}

\begin{proof}
\rev{By \Cref{lem:worst}, for the family of all distributions on the $S$ values, the RHS of \Cref{eq:wm-lower} is at least $(S-3)/(10(M+S))$ with $M=2N(i+1)+i$.
This exceeds $(1+\log|\Omega|)/N$ if and only if $N\,(S-3-20(i+1)(1+\log|\Omega|))>10\,(S+i)\,(1+\log|\Omega|)$.
If $N\geq34(1+\log|\Omega|)$ and $S\geq34(i+1)(1+\log|\Omega|)$, the left side is at least $34(1+\log|\Omega|)\,(S-3-20(i+1)(1+\log|\Omega|))$, which exceeds the RHS because $24S\geq816(i+1)(1+\log|\Omega|)>102+10i+680(i+1)(1+\log|\Omega|)$.}
\end{proof}

\subsection{Proof of Proposition \ref{prop:outweighing}}\label{sec:proof-outweighing}

\rev{We show two useful lemmas.
Since the $\mathsf{H}^2$ distance compares probabilities of next states, not their physical distances, we measure effects in the state space by the $1$-Wasserstein distance $\mathsf{W}$.}

\begin{lavenderbox}
\begin{restatable}{lemma}{lemcwmwupper}\label{lem:cwm-w-upper}
\revon
In \Cref{prop:cwm-upper}, suppose $\mathbb{E}_{{\bf x}_{i+1}\sim p_\omega(\cdot|{\bf x}_i,a_i,o_i)}\|{\bf x}_{i+1}-\hat{\bf x}_{i+1}\|^2\leq A^2$ for all $\omega\in\Omega$ and all pairs of factual and counterfactual transitions.
Then, the maximum likelihood CWM has an expected error, in squared $1$-Wasserstein distance $\mathsf{W}^2$, of at most $8A^2(1+\log|\Omega|)/N$.
\end{restatable}
\end{lavenderbox}
\begin{proof}
\rev{
We remark that the $1$-Wasserstein distance $\mathsf{W}(\nu,\rho)$ is defined by the infimum of $\mathbb{E}\|{\bf y}-{\bf y}'\|$ over all pairs of random variables ${\bf y}\sim\nu$ and ${\bf y}'\sim\rho$.

Fix a pair.
We write $p_\omega$ and $p$ for the densities of $p_\omega(\cdot|{\bf x}_i,a_i,o_i)$ and of the counterfactual distribution $p(\cdot|{\bf x}_i,a_i,o_i)$ at a next state ${\bf x}_{i+1}$, with respect to a common reference measure, and $\mathsf{W}$ and $\mathsf{H}^2$ for the $1$-Wasserstein and the squared Hellinger distance between these two distributions.
We first prove
\begin{equation}\label{eq:w-hellinger}
    \mathsf{W}^2\leq8A^2\,\mathsf{H}^2\qquad\text{for every }\omega\in\Omega.
\end{equation}
The two distributions share the mass $\min(p_\omega,p)$.
We obtain two random next states with these distributions by leaving the shared mass in place and moving the remaining mass $(p_\omega-p)_+$ of the model to the remaining mass $(p-p_\omega)_+$ of the counterfactual distribution in any way.
By the triangle inequality, a move from ${\bf x}_{i+1}$ to ${\bf x}_{i+1}'$ has a length of at most $\|{\bf x}_{i+1}-\hat{\bf x}_{i+1}\|+\|{\bf x}_{i+1}'-\hat{\bf x}_{i+1}\|$.
Hence, the expected distance between the two next states, and thus $\mathsf{W}$, satisfies
\begin{align*}
    \mathsf{W}
    &\leq\int\|{\bf x}_{i+1}-\hat{\bf x}_{i+1}\|\,(p_\omega-p)_+\,d{\bf x}_{i+1}+\int\|{\bf x}_{i+1}-\hat{\bf x}_{i+1}\|\,(p-p_\omega)_+\,d{\bf x}_{i+1}\\
    &=\int\|{\bf x}_{i+1}-\hat{\bf x}_{i+1}\|\,\big(\sqrt{p_\omega}+\sqrt{p}\big)\big|\sqrt{p_\omega}-\sqrt{p}\big|\,d{\bf x}_{i+1},
\end{align*}
where the equality holds as $(p_\omega-p)_++(p-p_\omega)_+=|p_\omega-p|$.
By the Cauchy--Schwarz inequality,
\begin{equation*}
    \mathsf{W}^2
    \leq\int\|{\bf x}_{i+1}-\hat{\bf x}_{i+1}\|^2\big(\sqrt{p_\omega}+\sqrt{p}\big)^2d{\bf x}_{i+1}\,\int\big(\sqrt{p_\omega}-\sqrt{p}\big)^2d{\bf x}_{i+1}.
\end{equation*}
The second integral is $2\,\mathsf{H}^2$.
The first is at most $4A^2$, because $(\sqrt{p_\omega}+\sqrt{p})^2\leq2\,(p_\omega+p)$ and both distributions have a mean squared distance of at most $A^2$ to $\hat{\bf x}_{i+1}$.
For the counterfactual distribution, this holds because it belongs to the class.
This shows \Cref{eq:w-hellinger}.
For $\omega=\hat{\omega}$, taking the expectation over the training pairs and the new pair and applying \Cref{eq:error_bound} gives the bound $8A^2(1+\log|\Omega|)/N$.
}
\end{proof}

\begin{lavenderbox}
\begin{restatable}{lemma}{lemwmwlower}\label{lem:wm-w-lower}
\revon
In \Cref{prop:wm-lower}, suppose that the visible noise distribution is unrestricted, and there are two values of visible noise whose next states, averaged over hidden noise, are at distance at least $E$ at every state-action.
Then, every WM has a worst-case expected $\mathsf{W}^2$ error of at least $E^2/(6(3i+4)N)$.
\end{restatable}
\end{lavenderbox}
\begin{proof}
\rev{
The $1$-Wasserstein distance between two distributions is at least the distance between their means, as $\|\mathbb{E}[{\bf y}-{\bf y}']\|\leq\mathbb{E}\|{\bf y}-{\bf y}'\|$ for any two random variables ${\bf y}$ and ${\bf y}'$.
It thus suffices to prove the bound for the squared distance between the mean of $p_{\hat\omega}$ and that of the transition distribution.

Let ${\bf v}$ and ${\bf v}'$ be the two values of the visible noise, and let the visible noise of every transition equal ${\bf v}'$ with probability $\lambda$ and ${\bf v}$ otherwise, where $\lambda\sim\mathsf{U}[0,1]$.
We prove that the mean of every estimate $p_{\hat\omega}$ then has an expected squared distance of at least $E^2/(6(3i+4)N)$ to the mean of the transition distribution, on average over $\lambda$.
This shows the claim, because the worst case is at least the average.

At a state-action $({\bf x},a)$, let ${\bf y}$ and ${\bf y}'$ be the next states for ${\bf v}$ and ${\bf v}'$, averaged over the hidden noise, so that the transition distribution has the mean ${\bf m}({\bf x},a)={\bf y}+\lambda\,({\bf y}'-{\bf y})$.
Let $\hat{\bf m}({\bf x},a)$ be the mean of $p_{\hat\omega}(\cdot|{\bf x},a)$, and let $\hat\lambda\coloneqq({\bf y}'-{\bf y})^\top(\hat{\bf m}({\bf x},a)-{\bf y})/\|{\bf y}'-{\bf y}\|^2$.
Then
\begin{equation}\label{eq:mean-to-weight}
    (\hat\lambda-\lambda)^2=\frac{\big(({\bf y}'-{\bf y})^\top(\hat{\bf m}({\bf x},a)-{\bf m}({\bf x},a))\big)^2}{\|{\bf y}'-{\bf y}\|^4}\leq\frac{\|\hat{\bf m}({\bf x},a)-{\bf m}({\bf x},a)\|^2}{E^2}
\end{equation}
by the Cauchy--Schwarz inequality and $\|{\bf y}'-{\bf y}\|\geq E$.

We use the above at the new state-action $({\bf x}_i,a_i)$.
As in the proof of \Cref{prop:wm-lower}, the $2N$ transitions and $({\bf x}_i,a_i)$ are determined by the initial states, the action noise, and the environmental noise of $M=2N(i+1)+i$ transitions.
Among these, only the visible noise has a distribution that depends on $\lambda$, as the distribution of the hidden noise given the visible noise is fixed.
Thus, $\hat\lambda$ is an estimate of $\lambda$ from $M$ independent samples that equal ${\bf v}'$ with probability $\lambda$.
Given that $k$ of the samples equal ${\bf v}'$, the posterior distribution of $\lambda$ is $\mathsf{Beta}{(k+1,M-k+1)}$, and its variance $\frac{(k+1)(M-k+1)}{(M+2)^2(M+3)}$
is the smallest expected squared error with which any estimate can recover $\lambda$ given the samples.
Since $k\sim\mathsf{U}\{0,\ldots,M\}$, the average of $(k+1)(M-k+1)$ is $(M+2)(M+3)/6$, so the expected squared error of $\hat\lambda$ is at least $1/(6(M+2))$.
By \Cref{eq:mean-to-weight}, the expected squared distance of $\hat{\bf m}({\bf x}_i,a_i)$ to ${\bf m}({\bf x}_i,a_i)$ is $\geq E^2/(6(M+2))$, which is $\geq E^2/(6(3i+4)N)$ as $M+2\leq(3i+4)N$.
}
\end{proof}

\rev{We now prove \Cref{prop:outweighing}.}

\begin{lavenderbox}
\propoutweighing*
\end{lavenderbox}

\begin{proof}
\rev{$8A^2(1+\log|\Omega|)/N<E^2/(6(3i+4)N)$ if and only if $E^2>48(3i+4)A^2(1+\log|\Omega|)$.}
\end{proof}

\subsection{Proof of Proposition \ref{prop:diffusion}}\label{sec:proof-diffusion}

\propdiffusion*
\begin{proof}
\rev{For an observation $o_i=(\hat{\bf x}_i,\hat a_i,\hat{\bf x}_{i+1})$, inversion returns the latent $\tilde{\bf u}$ with $f_\omega(\hat{\bf x}_i,\hat a_i,\tilde{\bf u})=\hat{\bf x}_{i+1}$, and the diffusion CWM returns $f_\omega({\bf x}_i,a_i,\tilde{\bf u})$.

Let $\hat\tau$ be a trajectory of the MDP and $\tilde{\bf u}_k$ the latent inverted from its $k$-th transition.
Given the past $(\hat{\bf x}_0,\hat a_0,\ldots,\hat{\bf x}_k,\hat a_k)$ and all action noise, $\hat{\bf x}_{k+1}$ has distribution $p(\cdot|\hat{\bf x}_k,\hat a_k)$, which by assumption is the distribution of $f_\omega(\hat{\bf x}_k,\hat a_k,\tilde{\bf u})$ for $\tilde{\bf u}\sim\tilde\nu$.
As $f_\omega(\hat{\bf x}_k,\hat a_k,\cdot)$ is invertible, $\tilde{\bf u}_k$ has distribution $\tilde\nu$ given this past and the action noise.
It is therefore independent of them, and in particular of $\tilde{\bf u}_0,\ldots,\tilde{\bf u}_{k-1}$, which are functions of the past.
Hence, $\hat\tau$ is generated by the equations ${\bf x}_{k+1}=f_\omega({\bf x}_k,a_k,\tilde{\bf u}_k)$ with mutually independent noise $\tilde{\bf u}_k\sim\tilde\nu$ as in \Cref{eq:scm}, which is thus an SCM consistent with the MDP.
Its counterfactual distribution \Cref{eq:counterfactual} is the point mass at $f_\omega({\bf x}_k,a_k,\tilde{\bf u}_k)$, because $\tilde{\bf u}_k$ is determined by $o_k$.
This is the output of the diffusion CWM.

The diffusion WM predicts $f_\omega({\bf x}_i,a_i,\tilde{\bf u})$ with $\tilde{\bf u}\sim\mathsf{N}(0,I)$, whereas the transition distribution is that of $f_\omega({\bf x}_i,a_i,\tilde{\bf u})$ with $\tilde{\bf u}\sim\tilde\nu$.
Applying a map to two distributions does not increase their squared Hellinger distance (\Cref{sec:proof-wm-lower}).
As $f_\omega({\bf x}_i,a_i,\cdot)$ is invertible, this holds for the map and for its inverse, so the squared Hellinger distance between the two distributions equals $\mathsf{H}^2(\mathsf{N}(0,I),\tilde\nu)$.
}
\end{proof}

\subsection{Proof of Proposition \ref{prop:diffusion-stability}}\label{sec:proof-diffusion-stability}

\begin{lavenderbox}
\propdiffusionstability*
\end{lavenderbox}

\begin{proof}
\rev{Let ${\bf u}_i$ be the noise of $f$ behind the factual transition, so that $\hat{\bf x}_{i+1}=f(\hat{\bf x}_i,\hat a_i,{\bf u}_i)$.
By \Cref{eq:noise-lipschitz}, two noise values that lead to the same next state at $(\hat{\bf x}_i,\hat a_i)$ lead to the same next state at $({\bf x}_i,a_i)$, so the counterfactual of $f$ is the point mass at $f({\bf x}_i,a_i,{\bf u}_i)$.
Let $\tilde{\bf u}$ be the latent returned by inversion, and let ${\bf w}\coloneqq T(\tilde{\bf u})$.
We first show that the noise value ${\bf w}$ nearly reproduces the factual transition:
\begin{align*}
    \|f(\hat{\bf x}_i,\hat a_i,{\bf w})-f(\hat{\bf x}_i,\hat a_i,{\bf u}_i)\|
    &\leq\|f(\hat{\bf x}_i,\hat a_i,T(\tilde{\bf u}))-f_\omega(\hat{\bf x}_i,\hat a_i,\tilde{\bf u})\|+\|f_\omega(\hat{\bf x}_i,\hat a_i,\tilde{\bf u})-\hat{\bf x}_{i+1}\|\\
    &\leq\varepsilon+\varepsilon_{\mathsf{inv}}.
\end{align*}
By \Cref{eq:noise-lipschitz}, it thus nearly reproduces the counterfactual, $\|f({\bf x}_i,a_i,{\bf w})-f({\bf x}_i,a_i,{\bf u}_i)\|\leq\Lambda\,(\varepsilon+\varepsilon_{\mathsf{inv}})$.
Finally, the output of the diffusion CWM satisfies $\|f_\omega({\bf x}_i,a_i,\tilde{\bf u})-f({\bf x}_i,a_i,{\bf w})\|\leq\varepsilon$, and the triangle inequality gives the claim.}
\end{proof}

\rev{
The result requires neither that $f_\omega$ is invertible nor that the inversion is exact.
The error $\varepsilon_{\mathsf{inv}}$ is the distance between the factual next state and its reconstruction from the inverted latent under the factual action, which can be measured empirically (see the composition criterion in \Cref{tab:dino:cfquality}).
Along a rollout, such errors can accumulate as for any learned model, but the bias of \ours{} depends on them only through the error $\eta$ of the expected return of a branch (\Cref{prop:learned}).
}


\subsection{Proof of Proposition \ref{prop:rlb-unbiased}}\label{sec:proof-rlb-unbiased}

\rev{We first fix the notation used in the remainder of the appendix.
A \emph{group} consists of the $m$ trajectories of \Cref{eq:rlb} at a step $i$, the factual $\tau^{(1)}_{\geq i}=\hat\tau_{\geq i}$ and counterfactual branches $\tau^{(2)}_{\geq i},\ldots,\tau^{(m)}_{\geq i}$, which all start from $\hat{\bf x}_i$.
As in \Cref{eq:rlb}, $G_i^{(j)}$ is the return-to-go of trajectory $j$ and ${\bf z}_i^{(j)}$ is the score of its first action.
The estimator of the group and its target are
\begin{equation}\label{eq:group}
    \widehat{\bf g}_i\coloneqq\frac1m\sum_{j=1}^m\big(G_i^{(j)}-b_i^{(j)}\big){\bf z}_i^{(j)},
    \qquad
    {\bf g}_i({\bf x})\coloneqq\mathbb{E}_{\tau\sim p_\theta}\big[G_i{\bf z}_i\big|{\bf x}_i={\bf x}\big],
\end{equation}
so that \Cref{eq:rlb} is $\sum_i\gamma^i\widehat{\bf g}_i$, and $\nabla_\theta J(\theta)=\sum_i\gamma^i\,\mathbb{E}[{\bf g}_i({\bf x}_i)]$ by \Cref{eq:reinforce}.
Expanding the baselines $b_i^{(j)}$ gives
\begin{equation}\label{eq:group-centered}
    \widehat{\bf g}_i=\frac1{m-1}\sum_{j=1}^mG_i^{(j)}\big({\bf z}_i^{(j)}-\bar{\bf z}_i\big),
    \qquad
    \bar{\bf z}_i\coloneqq\frac1m\sum_{j=1}^m{\bf z}_i^{(j)},
\end{equation}
and since $\sum_j({\bf z}_i^{(j)}-\bar{\bf z}_i)=0$, the estimator is unchanged when the same quantity is added to all returns of the group.

The score ${\bf z}_i^{(j)}$ depends only on $\hat{\bf x}_i$ and on the first action of trajectory $j$, which by \Cref{eq:scm} is drawn from $\pi_\theta(\cdot|\hat{\bf x}_i)$ using its own action noise.
Hence
\begin{equation}\label{eq:score-zero}
    \mathbb{E}\big[{\bf z}_i^{(j)}\,\big|\,\hat{\bf x}_i\big]=\sum_{a\in A}\nabla_\theta\pi_\theta(a|\hat{\bf x}_i)=0,
\end{equation}
and the same holds given any further variables that are independent of this action noise.}

\rev{We now prove a useful lemma.}

\begin{lavenderbox}
\begin{lemma}\label{lem:posterior}
\rev{Assume that all noise variables in \Cref{eq:scm} are mutually independent.
Let $\hat\tau$ be generated by \Cref{eq:scm} under any policy, with noise $\hat{\bf u}=(\hat{\bf u}_i,\ldots,\hat{\bf u}_{n-1})$ from step $i$ on, and let a branch from $\hat{\bf x}_i$ be drawn from $p_\theta(\cdot|\hat{\bf x}_i,\hat\tau)$ with the exact counterfactuals \Cref{eq:counterfactual}.
Then, jointly with $\hat\tau$, the branch has the same distribution as the trajectory from $\hat{\bf x}_i$ that uses the environmental noise $\hat{\bf u}$ of the factual and fresh action noise.
In particular, given $\hat{\bf x}_i$, the branch is distributed as $p_\theta(\cdot|\hat{\bf x}_i)$.}
\end{lemma}
\end{lavenderbox}

\begin{proof}
\rev{By \Cref{eq:counterfactual}, the branch is generated by ${\bf x}_{k+1}=f({\bf x}_k,a_k,{\bf u}_k')$ for $k\geq i$, where ${\bf u}_k'$ is drawn from $p({\bf u}_k|o_k)$ independently over $k$ given $\hat\tau$, and $a_k$ is drawn with fresh action noise.
By the independence of the noise variables and the form of \Cref{eq:scm}, the environmental noise of the factual satisfies
$p(\hat{\bf u}_i,\ldots,\hat{\bf u}_{n-1}|\hat\tau)=\prod_{k=i}^{n-1}p(\hat{\bf u}_k|o_k)$.
Hence $({\bf u}_i',\ldots,{\bf u}_{n-1}')$ and $\hat{\bf u}$ have the same conditional distribution given $\hat\tau$, and therefore the same joint distribution with $\hat\tau$, which proves the first claim.
The second claim follows because $\hat{\bf u}$ is independent of $\hat{\bf x}_i$ and has its original distribution.}
\end{proof}

\rev{We now prove \Cref{prop:rlb-unbiased}.}

\begin{lavenderbox}
\proprlbunbiased*
\end{lavenderbox}

\begin{proof}
\rev{We show $\mathbb{E}[\widehat{\bf g}_i|\hat{\bf x}_i]={\bf g}_i(\hat{\bf x}_i)$ for every step $i$.
Expanding the baselines in \Cref{eq:group},
\begin{equation}\label{eq:group-expanded}
    \widehat{\bf g}_i=\frac1m\sum_{j=1}^mG_i^{(j)}{\bf z}_i^{(j)}-\frac1{m(m-1)}\sum_{j=1}^m\sum_{\ell\neq j}G_i^{(\ell)}{\bf z}_i^{(j)}.
\end{equation}
In the first sum, $\mathbb{E}[G_i^{(1)}{\bf z}_i^{(1)}|\hat{\bf x}_i]={\bf g}_i(\hat{\bf x}_i)$ by definition, and the same holds for $j\geq2$ since a branch is distributed as $p_\theta(\cdot|\hat{\bf x}_i)$ given $\hat{\bf x}_i$ by \Cref{lem:posterior}.

Every term of the second sum has zero mean.
If $j\geq2$, the first action of branch $j$ is drawn with fresh action noise, which is independent of $\hat\tau$ and of the other branches, and these determine $G_i^{(\ell)}$.
Hence $\mathbb{E}[G_i^{(\ell)}{\bf z}_i^{(j)}|\hat{\bf x}_i]=0$ by \Cref{eq:score-zero}.
If $j=1$ and $\ell\geq2$, \Cref{lem:posterior} lets us replace branch $\ell$ by the trajectory that uses the noise $\hat{\bf u}$ of the factual and fresh action noise.
Its return is then determined by $\hat{\bf x}_i$, $\hat{\bf u}$, and this fresh noise, all of which are independent of the action noise of $\hat a_i$.
Hence $\mathbb{E}[G_i^{(\ell)}{\bf z}_i^{(1)}|\hat{\bf x}_i]=0$ by \Cref{eq:score-zero} again.

Therefore, we conclude that $\mathbb{E}[\widehat{\bf g}_i|\hat{\bf x}_i]={\bf g}_i(\hat{\bf x}_i)$.
The same holds given the whole factual past $(\hat{\bf x}_0,\hat a_0,\ldots,\hat{\bf x}_i)$, of which the noise from step $i$ on is independent.
Summing over the steps of the factual $\hat\tau\sim p_\theta$ gives $\mathbb{E}[\sum_i\gamma^i\widehat{\bf g}_i]=\sum_i\gamma^i\,\mathbb{E}[{\bf g}_i(\hat{\bf x}_i)]=\nabla_\theta J(\theta)$.}
\end{proof}

\subsection{Proof of Corollary \ref{cor:identifiability}}\label{sec:proof-identifiability}

\begin{lavenderbox}
\coridentifiability*
\end{lavenderbox}

\begin{proof}
\rev{
The estimator is a function of the factual $\hat\tau$ and its branches.
The distribution of $\hat\tau$ is $p_\theta$ for every SCM consistent with the MDP, and the branches are drawn given $\hat\tau$ from the counterfactual distribution of the SCM in use.
Their joint distribution is therefore the same as if $\hat\tau$ had been generated by this SCM, to which \Cref{prop:rlb-unbiased} applies.
As $\nabla_\theta J(\theta)$ depends on the MDP only, the estimators for all consistent SCMs have the same expectation (and can differ only in variance).
}
\end{proof}


\subsection{Proof of Proposition \ref{prop:variance}}\label{sec:proof-variance}
\rev{Throughout this section we consider one group, condition on its branching state $\hat{\bf x}_i$ without writing it, and use the notation of \Cref{sec:proof-rlb-unbiased}.
As in \Cref{sec:rlb}, the return-to-go of a trajectory of the group is a function $G(\boldsymbol{\epsilon},{\bf u})$ of its action noise $\boldsymbol{\epsilon}=(\boldsymbol{\epsilon}_i,...,\boldsymbol{\epsilon}_{n-1})$ and environmental noise ${\bf u}=({\bf u}_i,...,{\bf u}_{n-1})$, and we write $G^{(j)}$ for $G_i^{(j)}$.
We assume $G$, ${\bf z}_i$ and $G{\bf z}_i$ have finite second moments.

RLOO gives every trajectory of the group its own environmental noise.
We compare it with a group whose $m$ trajectories share the environmental noise ${\bf u}$ of the factual, and write $\widehat{\bf g}_{\mathsf{shr}}$ for the estimator \Cref{eq:group} of such a group.
If no noise is hidden, this is \ours{}, as the factual transition determines its noise, which the branches use at every step.
\Cref{lem:hidden} below accounts for hidden noise.
In all cases, the action noises of the $m$ trajectories are independent of each other and of the environmental noise.
For REINFORCE, we write $\widehat{\bf g}_b\coloneqq\frac1m\sum_j(G^{(j)}-b)\,{\bf z}_i^{(j)}$, with $m$ independent trajectories from $\hat{\bf x}_i$ as for RLOO and a baseline $b$ that may depend on $\hat{\bf x}_i$ but not on the noise of these trajectories.

Recall $V({\bf u})=\mathbb{E}[G|{\bf u}]$, which we call the hindsight value, and let
\begin{equation*}
    \tilde G\coloneqq G-V({\bf u}),\qquad {\bf g}({\bf u})\coloneqq\mathbb{E}[G{\bf z}_i|{\bf u}],\qquad {\bf g}\coloneqq\mathbb{E}[G{\bf z}_i],\qquad S\coloneqq\mathbb{E}[{\bf z}_i{\bf z}_i^\top],
\end{equation*}
so that ${\bf g}({\bf u})$ is the target given the noise and ${\bf g}=\mathbb{E}[{\bf g}({\bf u})]={\bf g}_i(\hat{\bf x}_i)$ is the target of the group.
A subscript ${\bf u}$, as in $\operatorname{Var}_{\bf u}$ and $\operatorname{Cov}_{\bf u}$, means that the action noise is held fixed.
As in \Cref{prop:variance}, $A\succeq B$ means that $A-B$ is positive semidefinite, that is, ${\bf v}^\top A{\bf v}\geq{\bf v}^\top B{\bf v}$ for every vector ${\bf v}$.}

\rev{We now prove some useful lemmas.}

\begin{lavenderbox}
\begin{lemma}\label{lem:loo}
\rev{Let $(G_j,{\bf z}_j)$ for $j=1,\ldots,m$ be independent and identically distributed, with $\mathbb{E}[{\bf z}_1]=0$ and finite second moments of $G_1$, ${\bf z}_1$ and $G_1{\bf z}_1$, and let
\begin{equation*}
    \widehat{\bf g}=\frac1m\sum_j(G_j-b_j){\bf z}_j,\qquad b_j=\frac1{m-1}\sum_{\ell\neq j}G_\ell.
\end{equation*}
Then $\mathbb{E}[\widehat{\bf g}]={\bf d}\coloneqq\mathbb{E}[G_1{\bf z}_1]$ and
\begin{equation}\label{eq:loo-cov}
    \operatorname{Cov}(\widehat{\bf g})
    =\frac1m\,\mathbb{E}\big[(G_1-\mathbb{E}[G_1])^2{\bf z}_1{\bf z}_1^\top\big]
    +\frac{\operatorname{Var}(G_1)}{m(m-1)}\,\mathbb{E}\big[{\bf z}_1{\bf z}_1^\top\big]
    -\frac{m-2}{m(m-1)}\,{\bf d}{\bf d}^\top.
\end{equation}}
\end{lemma}
\end{lavenderbox}

\begin{proof}
\rev{By \Cref{eq:group-centered}, $\widehat{\bf g}$ is unchanged when a constant is subtracted from all $G_j$, so we may assume $\mathbb{E}[G_1]=0$.
Write $\widehat{\bf g}={\bf H}-{\bf K}$, ${\bf H}=\frac1m\sum_jG_j{\bf z}_j$, and ${\bf K}=\frac1{m(m-1)}\sum_{j\neq\ell}G_\ell{\bf z}_j$.
By independence, and since $G_j$ and ${\bf z}_j$ have zero mean, $\mathbb{E}[{\bf H}]={\bf d}$, $\mathbb{E}[{\bf K}]=0$, and $\operatorname{Cov}({\bf H})=\frac1m(\mathbb{E}[G_1^2{\bf z}_1{\bf z}_1^\top]-{\bf d}{\bf d}^\top)$.
In $\mathbb{E}[{\bf K}{\bf K}^\top]$, the term with indices $(j,\ell)$ and $(j',\ell')$ is $\mathbb{E}[G_\ell G_{\ell'}{\bf z}_j{\bf z}_{j'}^\top]$.
It vanishes unless every index occurs twice, that is, unless $(j',\ell')=(j,\ell)$, which contributes $\mathbb{E}[G_1^2]\,\mathbb{E}[{\bf z}_1{\bf z}_1^\top]$, or $(j',\ell')=(\ell,j)$, which contributes ${\bf d}{\bf d}^\top$.
Hence $\operatorname{Cov}({\bf K})=\frac1{m(m-1)}(\mathbb{E}[G_1^2]\,\mathbb{E}[{\bf z}_1{\bf z}_1^\top]+{\bf d}{\bf d}^\top)$.
Likewise, every term of $\mathbb{E}[{\bf H}{\bf K}^\top]$ contains a factor $G_\ell$ or ${\bf z}_j$ whose index occurs only once, so $\operatorname{Cov}({\bf H},{\bf K})=0$.
Adding $\operatorname{Cov}({\bf H})$ and $\operatorname{Cov}({\bf K})$ gives \Cref{eq:loo-cov}.}
\end{proof}

\begin{lavenderbox}
\begin{lemma}\label{lem:gap}
\rev{The covariances of RLOO and of a group with shared noise satisfy
\begin{equation}\label{eq:gap}
\begin{aligned}
    \operatorname{Cov}(\widehat{\bf g}_{\mathsf{RLOO}})-\operatorname{Cov}(\widehat{\bf g}_{\mathsf{shr}})
    ={}&\frac{\sigma^2_{\mathsf{luck}}}{m-1}\,\mathbb{E}\big[{\bf z}_i{\bf z}_i^\top\big]
    +\frac2m\,\mathbb{E}\Big[\operatorname{Cov}_{\bf u}\big(V({\bf u}),\tilde G\big)\,{\bf z}_i{\bf z}_i^\top\Big]\\
    &-\Big(1-\frac{m-2}{m(m-1)}\Big)\operatorname{Cov}\big({\bf g}({\bf u})\big).
\end{aligned}
\end{equation}}
\end{lemma}
\end{lavenderbox}

\begin{proof}
\rev{For RLOO, the pairs $(G^{(j)},{\bf z}_i^{(j)})$ are independent and identically distributed, with $\mathbb{E}[{\bf z}_i^{(j)}]=0$ by \Cref{eq:score-zero}, so \Cref{lem:loo} applies.
We have $G-\mathbb{E}[G]=(V({\bf u})-\mathbb{E}[V({\bf u})])+\tilde G$, where $V({\bf u})$ is independent of the score ${\bf z}_i$, which depends on the action noise only.
Hence $\mathbb{E}[(G-\mathbb{E}[G])^2{\bf z}_i{\bf z}_i^\top]=\sigma^2_{\mathsf{luck}}S+2\,\mathbb{E}[\operatorname{Cov}_{\bf u}(V({\bf u}),\tilde G)\,{\bf z}_i{\bf z}_i^\top]+\mathbb{E}[\tilde G^2{\bf z}_i{\bf z}_i^\top]$, and $\operatorname{Var}(G)=\sigma^2_{\mathsf{luck}}+\mathbb{E}[\tilde G^2]$ because $\mathbb{E}[\tilde G|{\bf u}]=0$.
As $\frac1m+\frac1{m(m-1)}=\frac1{m-1}$, \Cref{eq:loo-cov} gives
\begin{equation*}
\begin{aligned}
    \operatorname{Cov}(\widehat{\bf g}_{\mathsf{RLOO}})
    ={}&\frac{\sigma^2_{\mathsf{luck}}}{m-1}S
    +\frac2m\,\mathbb{E}\Big[\operatorname{Cov}_{\bf u}\big(V({\bf u}),\tilde G\big)\,{\bf z}_i{\bf z}_i^\top\Big]\\
    &+\frac1m\,\mathbb{E}\big[\tilde G^2{\bf z}_i{\bf z}_i^\top\big]
    +\frac{\mathbb{E}[\tilde G^2]}{m(m-1)}S
    -\frac{m-2}{m(m-1)}\,{\bf g}{\bf g}^\top.
\end{aligned}
\end{equation*}

For shared noise, the pairs are independent and identically distributed given ${\bf u}$, with zero-mean scores and $G-\mathbb{E}[G|{\bf u}]=\tilde G$.
\Cref{lem:loo} applied given ${\bf u}$ yields $\mathbb{E}[\widehat{\bf g}_{\mathsf{shr}}|{\bf u}]={\bf g}({\bf u})$ and
\begin{equation*}
    \operatorname{Cov}(\widehat{\bf g}_{\mathsf{shr}}|{\bf u})
    =\frac1m\,\mathbb{E}\big[\tilde G^2{\bf z}_i{\bf z}_i^\top\big|{\bf u}\big]
    +\frac{\mathbb{E}[\tilde G^2|{\bf u}]}{m(m-1)}S
    -\frac{m-2}{m(m-1)}\,{\bf g}({\bf u}){\bf g}({\bf u})^\top.
\end{equation*}
By the law of total covariance, $\operatorname{Cov}(\widehat{\bf g}_{\mathsf{shr}})=\mathbb{E}[\operatorname{Cov}(\widehat{\bf g}_{\mathsf{shr}}|{\bf u})]+\operatorname{Cov}({\bf g}({\bf u}))$, and $\mathbb{E}[{\bf g}({\bf u}){\bf g}({\bf u})^\top]=\operatorname{Cov}({\bf g}({\bf u}))+{\bf g}{\bf g}^\top$.
Hence
\begin{equation*}
    \operatorname{Cov}(\widehat{\bf g}_{\mathsf{shr}})
    =\frac1m\,\mathbb{E}\big[\tilde G^2{\bf z}_i{\bf z}_i^\top\big]
    +\frac{\mathbb{E}[\tilde G^2]}{m(m-1)}S
    -\frac{m-2}{m(m-1)}\,{\bf g}{\bf g}^\top
    +\Big(1-\frac{m-2}{m(m-1)}\Big)\operatorname{Cov}\big({\bf g}({\bf u})\big).
\end{equation*}
Subtracting the two covariances gives \Cref{eq:gap}.}
\end{proof}

\rev{The three terms of \Cref{eq:gap} have a clear meaning.
The first is the environmental luck that affects RLOO but not \ours{}, as shared noise cancels the luck.
The second is the covariance between luck and the rest of the return, which can be positive or negative.
The third is the price of sharing one noise sample in \ours{}, as ${\bf g}({\bf u})$ varies with that sample, and a group cannot average this variation out.}

\begin{lavenderbox}
\begin{lemma}\label{lem:hidden}
\rev{Let counterfactuals be exact. Then
\begin{equation}\label{eq:hidden-bound}
    \operatorname{Cov}(\widehat{\bf g}_{\mathsf{\ours{}}})
    \preceq\operatorname{Cov}(\widehat{\bf g}_{\mathsf{shr}})+\frac{4\,\sigma^2_{\mathsf{inter}}}{m}\,\mathbb{E}\big[{\bf z}_i{\bf z}_i^\top\big],
\end{equation}
and $\operatorname{Cov}(\widehat{\bf g}_{\mathsf{\ours{}}})=\operatorname{Cov}(\widehat{\bf g}_{\mathsf{shr}})$ if no noise is hidden or if $\sigma_{\mathsf{inter}}=0$.}
\end{lemma}
\end{lavenderbox}

\begin{proof}
\rev{We first show what RLOO, shared noise and \ours{} have in common.

Let ${\bf a}\coloneqq(a_i^{(1)},\ldots,a_i^{(m)})$ be the first actions of the group, which determine its scores.
In all three cases, these are independent draws from $\pi_\theta(\cdot|\hat{\bf x}_i)$, and given ${\bf a}$, every return $G^{(j)}$ is distributed as the return of a trajectory from $\hat{\bf x}_i$ with first action $a_i^{(j)}$.
For a branch of \ours{}, this holds by \Cref{lem:posterior}: jointly with the factual, it is distributed as the trajectory that uses the noise $\hat{\bf u}$ of the factual and fresh action noise, and $\hat{\bf u}$ is independent of the first actions.
By \Cref{eq:group-centered} and the law of total covariance, each of the three estimators satisfies
\begin{equation}\label{eq:pairwise}
    \operatorname{Cov}(\widehat{\bf g})
    =\operatorname{Cov}\big(\mathbb{E}[\widehat{\bf g}|{\bf a}]\big)
    +\frac1{(m-1)^2}\,\mathbb{E}\bigg[\sum_{j,\ell}\operatorname{Cov}\big(G^{(j)},G^{(\ell)}\big|{\bf a}\big)\big({\bf z}_i^{(j)}-\bar{\bf z}_i\big)\big({\bf z}_i^{(\ell)}-\bar{\bf z}_i\big)^{\!\top}\bigg],
\end{equation}
where $\mathbb{E}[\widehat{\bf g}|{\bf a}]$ and the terms with $j=\ell$ are the same for the three.
The estimators thus differ only through the covariances of the returns of two different trajectories.

We next compare these covariances for shared noise and for \ours{}.
Let $Q(a,{\bf u})\coloneqq\mathbb{E}[G(\boldsymbol{\epsilon},{\bf u})|a_i=a]$ be the expected return of a trajectory from $\hat{\bf x}_i$ with first action $a$ and environmental noise ${\bf u}$.
For shared noise, the returns are independent given ${\bf u}$ and ${\bf a}$ with means $Q(a_i^{(j)},{\bf u})$, so that $\operatorname{Cov}(G^{(j)},G^{(\ell)}|{\bf a})=\operatorname{Cov}_{\bf u}(Q(a_i^{(j)},{\bf u}),Q(a_i^{(\ell)},{\bf u}))$ for $j\neq\ell$.
For \ours{}, the same holds for the factual and a branch, again by \Cref{lem:posterior}.
Two branches $j\neq\ell$, in contrast, are independent given $\hat\tau$ and ${\bf a}$, and each uses a noise with the conditional distribution of $\hat{\bf u}$ given $\hat\tau$, so that their returns have the conditional means $\mathbb{E}[Q(a_i^{(j)},\hat{\bf u})|\hat\tau]$ and $\mathbb{E}[Q(a_i^{(\ell)},\hat{\bf u})|\hat\tau]$.
Their covariance given ${\bf a}$ is the covariance of these two conditional means given $\hat a_i$.
Since $\hat{\bf u}$ is independent of $\hat a_i$, the law of total covariance shows that it equals $\operatorname{Cov}_{\bf u}(Q(a_i^{(j)},{\bf u}),Q(a_i^{(\ell)},{\bf u}))-\Gamma_{\hat a_i}(a_i^{(j)},a_i^{(\ell)})$, where
\begin{equation*}
    \Gamma_{\hat a}(a,a')\coloneqq\mathbb{E}\big[\operatorname{Cov}\big(Q(a,\hat{\bf u}),Q(a',\hat{\bf u})\,\big|\,\hat\tau\big)\,\big|\,\hat a_i=\hat a\big]
\end{equation*}
is the part of the covariance that the factual does not reveal.
Subtracting \Cref{eq:pairwise} for shared noise from \Cref{eq:pairwise} for \ours{} gives
\begin{equation}\label{eq:hidden}
    \operatorname{Cov}(\widehat{\bf g}_{\mathsf{\ours{}}})
    =\operatorname{Cov}(\widehat{\bf g}_{\mathsf{shr}})
    -\frac1{(m-1)^2}\,\mathbb{E}\bigg[\sum_{j\neq\ell,\;j,\ell\geq2}\Gamma_{\hat a_i}\big(a_i^{(j)},a_i^{(\ell)}\big)\big({\bf z}_i^{(j)}-\bar{\bf z}_i\big)\big({\bf z}_i^{(\ell)}-\bar{\bf z}_i\big)^{\!\top}\bigg].
\end{equation}

We now show the two cases of equality.
If no noise is hidden, $\hat{\bf u}$ is determined by $\hat\tau$, so that $\Gamma=0$.
If $\sigma_{\mathsf{inter}}=0$, then $G(\boldsymbol{\epsilon},{\bf u})=V({\bf u})+R(\boldsymbol{\epsilon})$ for a function $R$.
Every noise ${\bf u}$ in the support of the posterior given $\hat\tau$ reproduces all transitions of $\hat\tau$, and hence, with the action noise $\hat{\boldsymbol{\epsilon}}$ of the factual, the trajectory $\hat\tau$ itself.
Thus $V({\bf u})=G(\hat{\boldsymbol{\epsilon}},{\bf u})-R(\hat{\boldsymbol{\epsilon}})$ is the same for all such ${\bf u}$, that is, the factual reveals all luck.
As $Q(a,{\bf u})=V({\bf u})+\mathbb{E}[R(\boldsymbol{\epsilon})|a_i=a]$, again $\Gamma=0$.

Finally, we show \Cref{eq:hidden-bound}.
Let ${\bf v}$ be a vector and $w_j\coloneqq{\bf v}^\top({\bf z}_i^{(j)}-\bar{\bf z}_i)$.
In general, $\Gamma_{\hat a}$ is a positive semidefinite kernel, being an average of covariances, so that $\sum_{j,\ell\geq2}\Gamma_{\hat a_i}(a_i^{(j)},a_i^{(\ell)})\,w_jw_\ell\geq0$.
The sum over $j\neq\ell$ is therefore at least $-\sum_{j\geq2}\Gamma_{\hat a_i}(a_i^{(j)},a_i^{(j)})\,w_j^2$.
Here $\Gamma_{\hat a}(a,a)\leq4\sigma^2_{\mathsf{inter}}$.
To see this, write $Q(a,{\bf u})=V({\bf u})+\tilde Q(a,{\bf u})$ with $\tilde Q(a,{\bf u})\coloneqq\mathbb{E}[\tilde G(\boldsymbol{\epsilon},{\bf u})|a_i=a]$.
As shown above, every noise ${\bf u}$ in the support of the posterior given $\hat\tau$ reproduces $\hat\tau$ with the action noise $\hat{\boldsymbol{\epsilon}}$ of the factual, so $V({\bf u})+\tilde G(\hat{\boldsymbol{\epsilon}},{\bf u})$ is the return of $\hat\tau$ for all such ${\bf u}$.
As $\hat{\bf u}$ and $\hat{\boldsymbol{\epsilon}}$ remain independent given $\hat\tau$, the conditional variance of $V(\hat{\bf u})$ given $\hat\tau$ is that of $\tilde G(\hat{\boldsymbol{\epsilon}},\hat{\bf u})$ given $\hat\tau$ and $\hat{\boldsymbol{\epsilon}}$, whose mean given $\hat{\boldsymbol{\epsilon}}$ is at most $\operatorname{Var}_{\bf u}(\tilde G(\hat{\boldsymbol{\epsilon}},{\bf u}))\leq\sigma^2_{\mathsf{inter}}$ by the law of total variance.
Likewise, the mean conditional variance of $\tilde Q(a,\hat{\bf u})$ given $\hat\tau$ is at most $\operatorname{Var}_{\bf u}(\tilde Q(a,{\bf u}))\leq\sigma^2_{\mathsf{inter}}$, as $\tilde Q(a,{\bf u})$ averages $\tilde G(\boldsymbol{\epsilon},{\bf u})$ over the action noise, and the variance of an average is at most the largest variance.
The bound on $\Gamma_{\hat a}(a,a)$ follows by the triangle inequality for the standard deviation, given $\hat\tau$ and then on average over $\hat\tau$.
The scores of the group are independent and identically distributed with zero mean, so $\mathbb{E}[w_j^2]=(1-\frac1m)\,{\bf v}^\top S{\bf v}$.
By \Cref{eq:hidden}, ${\bf v}^\top\operatorname{Cov}(\widehat{\bf g}_{\mathsf{\ours{}}})\,{\bf v}$ thus exceeds ${\bf v}^\top\operatorname{Cov}(\widehat{\bf g}_{\mathsf{shr}})\,{\bf v}$ by at most $\frac{4\sigma^2_{\mathsf{inter}}}{(m-1)^2}\sum_{j\geq2}\mathbb{E}[w_j^2]=\frac{4\sigma^2_{\mathsf{inter}}}{m}\,{\bf v}^\top S{\bf v}$, which is \Cref{eq:hidden-bound} as ${\bf v}$ is arbitrary.}
\end{proof}

\rev{Hidden noise thus reduces only the correlation between two branches, as the factual and each of its branches are always correlated as if they shared the noise (\Cref{lem:posterior}).
It costs at most a multiple of the interaction, as noise values that a factual cannot tell apart give it the same return, so their luck can differ only as much as the effect of its actions does.}

\rev{We now prove \Cref{prop:variance}.}

\begin{lavenderbox}
\propvariance*
\end{lavenderbox}

\begin{proof}
\rev{We first bound the terms of \Cref{eq:gap}.
For every action noise, $|\operatorname{Cov}_{\bf u}(V({\bf u}),\tilde G)|\leq\sigma_{\mathsf{luck}}\sigma_{\mathsf{inter}}$ by the Cauchy--Schwarz inequality, so the second term is $\succeq-\frac2m\,\sigma_{\mathsf{luck}}\sigma_{\mathsf{inter}}\,S$, as ${\bf z}_i{\bf z}_i^\top\succeq0$.
For the third term, ${\bf g}({\bf u})-{\bf g}=\mathbb{E}[(\tilde G-\mathbb{E}[\tilde G|\boldsymbol{\epsilon}]){\bf z}_i|{\bf u}]$, so by the Cauchy--Schwarz inequality $({\bf v}^\top({\bf g}({\bf u})-{\bf g}))^2\leq\mathbb{E}[(\tilde G-\mathbb{E}[\tilde G|\boldsymbol{\epsilon}])^2|{\bf u}]\;{\bf v}^\top S{\bf v}$ for every vector ${\bf v}$, and taking expectations, $\operatorname{Cov}({\bf g}({\bf u}))\preceq\sigma^2_{\mathsf{inter}}S$.
Using $\frac1m\leq\frac1{m-1}$ and $1-\frac{m-2}{m(m-1)}\leq1$, and completing the square,
\begin{equation}\label{eq:gap-bound}
    \operatorname{Cov}(\widehat{\bf g}_{\mathsf{RLOO}})-\operatorname{Cov}(\widehat{\bf g}_{\mathsf{shr}})
    \succeq\frac{(\sigma_{\mathsf{luck}}-\sigma_{\mathsf{inter}})^2-m\,\sigma^2_{\mathsf{inter}}}{m-1}\,S.
\end{equation}
Subtracting \Cref{eq:hidden-bound}, where $\frac4m\leq\frac4{m-1}$, turns $m$ into $m+4$ in the numerator, which is the bound for RLOO.
If $\sigma_{\mathsf{inter}}=0$, then $\tilde G$ does not depend on ${\bf u}$.
The second term of \Cref{eq:gap} is then zero, and so is the third, because ${\bf g}({\bf u})=\mathbb{E}[\tilde G{\bf z}_i]$ is constant.
Since \ours{} then has the covariance of shared noise by \Cref{lem:hidden}, \Cref{eq:gap} is $\operatorname{Cov}(\widehat{\bf g}_{\mathsf{RLOO}})-\operatorname{Cov}(\widehat{\bf g}_{\mathsf{\ours{}}})=\frac{\sigma^2_{\mathsf{luck}}}{m-1}\,S$, which is the bound with equality.

We next compare with REINFORCE with the exact value $b=\mathbb{E}[G]$.
The $m$ terms of $\widehat{\bf g}_b$ are then independent and identically distributed with mean ${\bf g}$, so $\operatorname{Cov}(\widehat{\bf g}_b)=\frac1m(\mathbb{E}[(G-\mathbb{E}[G])^2{\bf z}_i{\bf z}_i^\top]-{\bf g}{\bf g}^\top)$.
Comparing with \Cref{eq:loo-cov} for RLOO gives
\begin{equation}\label{eq:rloo-reinforce}
    \operatorname{Cov}(\widehat{\bf g}_{\mathsf{RLOO}})-\operatorname{Cov}(\widehat{\bf g}_b)=\frac{\operatorname{Var}(G)\,S+{\bf g}{\bf g}^\top}{m(m-1)}\qquad\text{for }b=\mathbb{E}[G],
\end{equation}
which is what RLOO pays for estimating its baseline from the group.
We have ${\bf g}=\mathbb{E}[\tilde G{\bf z}_i]$, because $V({\bf u})$ is independent of ${\bf z}_i$, which has zero mean.
By the Cauchy--Schwarz inequality, ${\bf g}{\bf g}^\top\preceq\mathbb{E}[\tilde G^2]\,S=(\operatorname{Var}(G)-\sigma^2_{\mathsf{luck}})\,S$.
The right side of \Cref{eq:rloo-reinforce} is therefore $\preceq\frac{2\operatorname{Var}(G)-\sigma^2_{\mathsf{luck}}}{m(m-1)}\,S$, and subtracting it from the bound for RLOO gives the bound for REINFORCE.

We finally derive the two conditions.
The bound for RLOO is positive semidefinite if $(\sigma_{\mathsf{luck}}-\sigma_{\mathsf{inter}})^2\geq(m+4)\,\sigma^2_{\mathsf{inter}}$, that is, if $\sigma_{\mathsf{luck}}\geq(1+\sqrt{m+4})\,\sigma_{\mathsf{inter}}$.
In the bound for REINFORCE, $m-1$ times the coefficient of $S$ is $(\sigma_{\mathsf{luck}}-\sigma_{\mathsf{inter}})^2-(m+4)\,\sigma^2_{\mathsf{inter}}+\frac1m\sigma^2_{\mathsf{luck}}-\frac2m\operatorname{Var}(G)$.
For $\sigma_{\mathsf{luck}}=(1+\sqrt{m+4})\,\sigma_{\mathsf{inter}}+t$ with $t\geq0$, the terms without $\operatorname{Var}(G)$ sum to $2\sqrt{m+4}\,\sigma_{\mathsf{inter}}t+t^2+\frac1m\sigma^2_{\mathsf{luck}}\geq\frac{m+1}{m}\,t^2$, so the coefficient is nonnegative if $t\geq\sqrt{2\operatorname{Var}(G)/(m+1)}$.}
\end{proof}

\rev{The condition against REINFORCE asks for more luck than that against RLOO, as REINFORCE with the exact value does not have to estimate its baseline.
By \Cref{eq:rloo-reinforce}, it is never worse than RLOO.}

\subsection{Tightness of the conditions of Proposition \ref{prop:variance}}\label{sec:hurt}

\rev{We show that the conditions of \Cref{prop:variance} cannot be dropped and are tight up to a constant by constructing an example.

Let no noise be hidden, so that \ours{} has the covariance of shared noise by \Cref{lem:hidden}.
If $V({\bf u})$ is constant, there is no luck to cancel, and \Cref{eq:gap} reduces to $-(1-\frac{m-2}{m(m-1)})\operatorname{Cov}({\bf g}({\bf u}))\preceq0$.
Without luck, \ours{} thus never has a smaller covariance than RLOO, and it has a larger one whenever the target given the noise varies with the noise.
The same holds against REINFORCE with the exact value, whose covariance is at most that of RLOO by \Cref{eq:rloo-reinforce}.
An example is $G=a\,u$ for an action $a$ and a noise $u$ that are both uniform on $\{-1,1\}$, where the noise decides which action is better.
This is the known limitation of common random numbers~\citep{glasserman1992some}.

Now we add luck to the example, $G=\lambda w+a\,u$ with a further independent noise $w$ that is uniform on $\{-1,1\}$, so that $\sigma_{\mathsf{luck}}=\lambda$ and $\sigma_{\mathsf{inter}}=1$.
The scores of the two actions sum to zero, so ${\bf z}_i=a\,{\bf v}$ for a vector ${\bf v}$, and the target given the noise is $u\,{\bf v}$, with covariance ${\bf v}{\bf v}^\top=S$.
The second term of \Cref{eq:gap} vanishes, so \Cref{eq:gap} is $\big(\frac{\lambda^2}{m-1}-1+\frac{m-2}{m(m-1)}\big)\,S$.
Hence \ours{} has the smaller covariance if and only if $\sigma_{\mathsf{luck}}\geq\sqrt{m-2+2/m}\;\sigma_{\mathsf{inter}}$.
For REINFORCE with the exact value, $\operatorname{Var}(G)=\lambda^2+1$ and ${\bf g}=0$ in this example, so subtracting \Cref{eq:rloo-reinforce} leaves $\big(\frac{\lambda^2}{m}-1+\frac{m-3}{m(m-1)}\big)\,S$.
Hence \ours{} has the smaller covariance if and only if $\sigma_{\mathsf{luck}}\geq\sqrt{m-1+2/(m-1)}\;\sigma_{\mathsf{inter}}$.
Both thresholds have to grow as $\sqrt m$: the luck that RLOO and REINFORCE carry is averaged over the $m$ trajectories of a group, whereas the dependence of the target on the one noise draw that a group of \ours{} shares is averaged only over groups.
For the choice of $m$, this means that the covariance of a group of \ours{} does not fall below $\operatorname{Cov}({\bf g}({\bf u}))$ however large $m$ is (proof of \Cref{lem:gap}), a floor that only more factual trajectories reduce, so $m$ is worth increasing until the other terms are comparable to this floor.

The term added for REINFORCE is needed as well.
If $\sigma_{\mathsf{inter}}=0$ and ${\bf g}=0$, as when the first action does not affect the return, \Cref{eq:gap,eq:rloo-reinforce} give $\operatorname{Cov}(\widehat{\bf g}_b)-\operatorname{Cov}(\widehat{\bf g}_{\mathsf{\ours{}}})=\frac{m\,\sigma^2_{\mathsf{luck}}-\operatorname{Var}(G)}{m(m-1)}\,S$ for the exact value, which is positive semidefinite if and only if $\sigma_{\mathsf{luck}}\geq\sqrt{\operatorname{Var}(G)/m}$.
This is the condition of \Cref{prop:variance} up to a factor of at most $\sqrt2$.}

\subsection{Proof of Proposition \ref{prop:learned}}\label{sec:proof-learned}

\rev{We keep the setting and notation of \Cref{sec:proof-variance}, that is, we consider one group and condition on its branching state $\hat{\bf x}_i$ without writing it.
We first prove a useful lemma.}

\begin{lavenderbox}
\begin{lemma}\label{lem:luck}
\rev{Suppose that counterfactuals are exact.
Then, for every group and given $\hat{\bf x}_i$, \ours{} is unchanged if the luck that the factual reveals, $\mathbb{E}[V({\bf u})|\hat\tau]$, is subtracted from every return, whereas RLOO, and REINFORCE with $m$ independent trajectories and any state-dependent baseline, have
\begin{equation*}
    \operatorname{Cov}(\widehat{\bf g})\succeq\frac{(\sigma_{\mathsf{luck}}-\sigma_{\mathsf{inter}})_+^2}{m}\,\mathbb{E}[{\bf z}_i{\bf z}_i^{\top}],
\end{equation*}
where $(\cdot)_+$ denotes the positive part.}
\end{lemma}
\end{lavenderbox}

\begin{proof}
\rev{By \Cref{eq:group-centered}, the estimator is unchanged when the same quantity is added to all returns of the group.
This applies to $\mathbb{E}[V({\bf u})|\hat\tau]$, which is a function of the factual.
By \Cref{lem:posterior}, it is the mean hindsight value, given the factual, of the noise that the factual and each of its branches use.
If no noise is hidden, it is the hindsight value $V({\bf u})$ of the noise that the group shares, and \ours{} is \Cref{eq:group-centered} with $\tilde G^{(j)}$ in place of $G^{(j)}$.

For RLOO and REINFORCE, every trajectory has its own noise, so its hindsight value is not common to the group and does not cancel.
Indeed, the $m$ terms of $\widehat{\bf g}_b$ are independent and identically distributed, so $\operatorname{Cov}(\widehat{\bf g}_b)=\frac1m\operatorname{Cov}((G-b){\bf z}_i)$.
Given the action noise of a trajectory, its score ${\bf z}_i$ and the baseline $b$ are fixed, so the conditional covariance of $(G-b){\bf z}_i$ is $\operatorname{Var}_{\bf u}(G)\,{\bf z}_i{\bf z}_i^\top$.
As $G=V({\bf u})+\tilde G$, the standard deviation of $G$ over ${\bf u}$ is at least $\sigma_{\mathsf{luck}}-\sigma_{\mathsf{inter}}$ by the triangle inequality.
A covariance is at least the expected conditional covariance, so $\operatorname{Cov}(\widehat{\bf g}_b)\succeq\frac{(\sigma_{\mathsf{luck}}-\sigma_{\mathsf{inter}})_+^2}{m}\,S$ for every such baseline, and the same holds for RLOO by \Cref{eq:rloo-reinforce}.}
\end{proof}

\rev{We now turn to learned models.
The branches of \ours{} are drawn independently given $\hat\tau$ from the distribution $p_{\theta,\omega}(\cdot|\hat{\bf x}_i,\hat\tau)$ obtained by replacing the exact counterfactual transitions in $p_\theta(\cdot|\hat{\bf x}_i,\hat\tau)$ with $p_\omega({\bf x}_{k+1}|{\bf x}_k,a_k,o_k)$.
We write $\mathbb{E}_\omega$ and $\operatorname{Cov}_\omega$ when branches are generated by $p_\omega$, and $\mathbb{E}$ when they are exact counterfactuals, and let trajectory $2$ be a branch.
The error of the model is
\begin{equation}\label{eq:eta}
\eta\coloneqq\mathbb{E}\Big[\,\Big|\,\mathbb{E}_\omega\big[G^{(2)}\,\big|\,\hat\tau,a_i^{(2)}\big]-\mathbb{E}\big[G^{(2)}\,\big|\,\hat\tau,a_i^{(2)}\big]\Big|\,\Big],
\end{equation}
the mean absolute error of its expected return of a branch, given the factual and the first action of the branch.
We assume $\|{\bf z}_i\|\leq z_{\max}$ for all trajectories, and take the bias and the mean squared error with respect to the target ${\bf g}$ of the group.

RLOO with a WM takes the real trajectory $\hat\tau$ as its root and generates the $m-1$ branches from $\hat{\bf x}_i$ with the model and fresh actions, independently of the root~\citep{xiao2025world}.
REINFORCE uses the real trajectory alone, or $m$ independent ones as in \Cref{prop:variance}, with a baseline that depends on the state only, such as a learned value function.
The proposition assumes $|G^{(j)}-G^{(1)}|\leq A$ for every generated branch $j\geq2$.
We treat all models as fixed, that is, learned from data that are independent of the trajectories of the group.

We now prove \Cref{prop:learned}.}

\begin{lavenderbox}
\proplearned*
\end{lavenderbox}

\begin{proof}
\rev{We first bound the bias.
We take expectations in \Cref{eq:group-expanded}.
The cross terms with the score of a branch vanish as in the proof of \Cref{prop:rlb-unbiased}, which only used that the first action of a branch is drawn with fresh action noise.
The factual contributes $\mathbb{E}[G^{(1)}{\bf z}_i^{(1)}]={\bf g}$, and the $m-1$ branches contribute equal terms, so that
\begin{equation}\label{eq:bias}
    \mathbb{E}_\omega[\widehat{\bf g}_{\mathsf{\ours{}}}]-{\bf g}
    =\frac{m-1}{m}\Big(\mathbb{E}_\omega\big[G^{(2)}{\bf z}_i^{(2)}\big]-{\bf g}\Big)
    -\frac1m\,\mathbb{E}_\omega\big[G^{(2)}{\bf z}_i^{(1)}\big].
\end{equation}
For an exact branch, $\mathbb{E}[G^{(2)}{\bf z}_i^{(2)}]={\bf g}$ and $\mathbb{E}[G^{(2)}{\bf z}_i^{(1)}]=0$ by the proof of \Cref{prop:rlb-unbiased}.
The bias is therefore the difference, between a generated and an exact branch, of the expectation of $G^{(2)}\big(\frac{m-1}{m}{\bf z}_i^{(2)}-\frac1m{\bf z}_i^{(1)}\big)$.
The vector in the bracket is determined by $\hat\tau$ and the first action of the branch, which have the same distribution in both cases, and its norm is at most $z_{\max}$.
Taking the expectation of $G^{(2)}$ given $\hat\tau$ and this action first, the bias is at most $z_{\max}\eta$ in norm.

We next bound the covariance of \ours{}.
Let $D^{(j)}\coloneqq G^{(j)}-G^{(1)}$.
Subtracting $G^{(1)}$ from all returns in \Cref{eq:group-centered} gives
\begin{equation}\label{eq:group-difference}
    \widehat{\bf g}_{\mathsf{\ours{}}}=\frac1{m-1}\sum_{j=2}^mD^{(j)}\big({\bf z}_i^{(j)}-\bar{\bf z}_i\big),
\end{equation}
so that \ours{} depends on the returns only through their differences from the factual.
As $|D^{(j)}|\leq A$, the Cauchy--Schwarz inequality gives $({\bf v}^\top\widehat{\bf g}_{\mathsf{\ours{}}})^2\leq\frac{A^2}{m-1}\sum_{j\geq2}({\bf v}^\top({\bf z}_i^{(j)}-\bar{\bf z}_i))^2$ for every vector ${\bf v}$.
The scores are independent and identically distributed with zero mean, so $\mathbb{E}[({\bf z}_i^{(j)}-\bar{\bf z}_i)({\bf z}_i^{(j)}-\bar{\bf z}_i)^\top]=(1-\frac1m)\,S$, and therefore $\mathbb{E}_\omega[\widehat{\bf g}_{\mathsf{\ours{}}}\widehat{\bf g}_{\mathsf{\ours{}}}^\top]\preceq(1-\frac1m)A^2S$.
A covariance is at most the second moment, so $\operatorname{Cov}_\omega(\widehat{\bf g}_{\mathsf{\ours{}}})\preceq(1-\frac1m)A^2S$.

We next bound the covariance of the other estimators from below.
For RLOO with a WM, we condition on the action noise of the root and on the branches, of which the environmental noise ${\bf u}$ of the root is independent.
The scores in \Cref{eq:group-centered} are then fixed, and only the return $G(\boldsymbol{\epsilon},{\bf u})$ of the root is random, through ${\bf u}$.
Its variance is at least $(\sigma_{\mathsf{luck}}-\sigma_{\mathsf{inter}})_+^2$ as in the proof of \Cref{lem:luck}, so the conditional covariance of the estimator is at least $\frac{(\sigma_{\mathsf{luck}}-\sigma_{\mathsf{inter}})_+^2}{(m-1)^2}\,\big({\bf z}_i^{(1)}-\bar{\bf z}_i\big)\big({\bf z}_i^{(1)}-\bar{\bf z}_i\big)^{\!\top}$.

The scores of the group are independent and identically distributed with zero mean, so the last matrix has expectation $(1-\frac1m)\,S$.
As a covariance is at least the expected conditional covariance, $\operatorname{Cov}(\widehat{\bf g})\succeq\frac{(\sigma_{\mathsf{luck}}-\sigma_{\mathsf{inter}})_+^2}{m(m-1)}\,S$.
For REINFORCE, \Cref{lem:luck} gives the same with $\frac1m$ in place of $\frac1{m(m-1)}$, and its proof with $1$ if only the real trajectory is used.

We finally combine the bounds.
Subtracting the upper bound for \ours{}, where $1-\frac1m\leq1$, from the lower bound gives the claimed inequality.
If $\sigma_{\mathsf{luck}}\geq\sigma_{\mathsf{inter}}+mA$, the lower bound is $\succeq\frac{m}{m-1}A^2S$, so \ours{} has the smaller covariance.
Its mean squared error is the trace of its covariance plus the squared norm of its bias, hence at most $(1-\frac1m)A^2\,\mathbb{E}\|{\bf z}_i\|^2+z_{\max}^2\eta^2$, which is at most $A^2\,\mathbb{E}\|{\bf z}_i\|^2$ if $z_{\max}\eta\leq A\,(\mathbb{E}\|{\bf z}_i\|^2/m)^{1/2}$.
The mean squared error of the other estimators is at least the trace of their covariance, which is then at least $\frac{m}{m-1}A^2\,\mathbb{E}\|{\bf z}_i\|^2$.}
\end{proof}

\subsection{Proof of Proposition \ref{prop:rate}}\label{sec:proof-rate}

\rev{We consider a maximum likelihood CWM $p_{\hat\omega}$ of \Cref{prop:cwm-upper}, learned from a training set $\mathcal{D}$ independent of the trajectories of the policy gradient.
Suppose that, for every group and every step $k$, the probability of any set of inputs $({\bf x}_k,a_k,o_k)$ of the model, where $({\bf x}_k,a_k)$ follows an exact counterfactual branch of $\hat\tau$, is at most $\kappa$ times its probability under the training distribution, which is the distribution of $({\bf x}_i,a_i,o_i)$ in \Cref{eq:error_bound}.
In particular, the training pairs must include pairs as they occur in a group, whose two states coincide at the branching step.
For states drawn independently from a continuous distribution, $\kappa$ is infinite.
For the group at step $i$, $\eta$ is the error \Cref{eq:eta} given $\hat{\bf x}_i$, and $A$ bounds the difference between the returns of a branch and its factual as in \Cref{prop:learned}, for generated and for exact counterfactual branches.
The bias is that of the whole estimator \Cref{eq:rlb}.}

\begin{lavenderbox}
\proprate*
\end{lavenderbox}

\begin{proof}
\rev{We first relate $\eta$ to the error of a generated branch in total variation,
\begin{equation}\label{eq:eta-tv}
    \eta_{\mathsf{TV}}\coloneqq\mathbb{E}_{\hat\tau}\Big[\mathsf{TV}\big(p_{\theta,\omega}(\cdot|\hat{\bf x}_i,\hat\tau),\,p_\theta(\cdot|\hat{\bf x}_i,\hat\tau)\big)\Big],
\end{equation}
where $\mathsf{TV}(P,Q)=\sup_E|P(E)-Q(E)|$, for which the expectations of any function bounded by $B$ differ by at most $2B\,\mathsf{TV}(P,Q)$.
Given $\hat\tau$, the return of the factual is fixed, so the error in \Cref{eq:eta} is that of the expected value of $G^{(2)}-G^{(1)}$, which is bounded by $A$.
Given $\hat\tau$ and the first action of the branch, a generated and an exact branch differ in total variation by an amount whose average over this action is the distance inside \Cref{eq:eta-tv}.
Hence $\eta\leq2A\,\eta_{\mathsf{TV}}$.

We next show that, for every model $p_\omega$,
\begin{equation}\label{eq:rollout}
    \eta_{\mathsf{TV}}\leq\sum_{k=i}^{n-1}\mathbb{E}\Big[\mathsf{TV}\big(p_\omega(\cdot|{\bf x}_k,a_k,o_k),\,p(\cdot|{\bf x}_k,a_k,o_k)\big)\Big],
\end{equation}
where $({\bf x}_k,a_k)$ follows an exact counterfactual branch of $\hat\tau$.
Fix $\hat\tau$, and for $k=i,\ldots,n$, let $P_k$ be the distribution of the branch that uses the exact counterfactual transitions before step $k$ and $p_\omega$ from step $k$ on, so that $P_i$ is the generated and $P_n$ the exact branch.
The distributions $P_k$ and $P_{k+1}$ differ only in the transition at step $k$, taken from a state and action distributed as on the exact branch, and are obtained from the distribution of the trajectory up to ${\bf x}_{k+1}$ by the same later transitions.
Their total variation distance is therefore at most the expected total variation distance of this transition, and \Cref{eq:rollout} follows from the triangle inequality and the expectation over $\hat\tau$.

We next bound the terms of \Cref{eq:rollout} for $p_\omega=p_{\hat\omega}$, averaged over $\hat{\bf x}_i$.
By coverage assumption, the $\mathsf{H}^2$ distance between the model and the exact counterfactual, averaged over the inputs of step $k$, is at most $\kappa$ times its average over the training distribution, whose expectation over $\mathcal{D}$ is $\leq (1+\log|\Omega|)/N$ by \Cref{eq:error_bound}.
The total variation distance of two distributions is at most $\sqrt2$ times their Hellinger distance, so by Jensen's inequality the $k$-th term of \Cref{eq:rollout} has expectation $\leq \sqrt{2\kappa(1+\log|\Omega|)/N}$ over $\hat{\bf x}_i$ and $\mathcal{D}$.

We now prove the claims.
By \Cref{eq:rollout}, the expectation of $\eta_{\mathsf{TV}}$ over $\hat{\bf x}_i$ and $\mathcal{D}$ is at most $n-i$ times this bound, and the bound on $\eta$ follows as $\eta\leq2A\,\eta_{\mathsf{TV}}$.
As \Cref{eq:rlb} is $\sum_i\gamma^i\,\widehat{\bf g}_i$ and $\nabla_\theta J(\theta)=\sum_i\gamma^i\,\mathbb{E}[{\bf g}_i(\hat{\bf x}_i)]$, its bias is the sum over $i$ of $\gamma^i$ times the mean over $\hat{\bf x}_i$ of the bias of the group at step $i$, which is $\leq z_{\max}\eta$ in norm by \Cref{prop:learned}.
The bound on the bias follows as $\gamma\leq1$ and $\sum_{i=0}^{n-1}(n-i)=n(n+1)/2$.
}
\end{proof}

\subsection{Sparse branching}\label{sec:branching}
\rev{Branching at every step costs $m-1$ model rollouts per step.
In practice, it is useful to branch sparsely, where we branch only at the first step and use the same $m-1$ branches at every later step:
\begin{equation}\label{eq:rlb-sparse}
    \nabla_\theta J(\theta)\approx\frac1m\sum_{j=1}^m\sum_{i=0}^{n-1}\gamma^i\big(G_i^{(j)}-b_i^{(j)}\big){\bf z}_i^{(j)},
    \qquad \tau^{(1)}=\hat\tau,
    \quad \tau^{(2)},\ldots,\tau^{(m)}\overset{\text{iid}}{\sim}p_\theta(\cdot|\hat{\bf x}_0,\hat\tau),
\end{equation}
with $b_i^{(j)}$ as in \Cref{eq:rlb}, where ${\bf z}_i^{(j)}$ is the score of trajectory $j$ at its own state ${\bf x}_i^{(j)}$.
The trajectories of the group at a step $i\geq1$ thus no longer share a state.
RLOO with sparse branching uses independent branches instead~\citep{kool2019buy}.}

\begin{lavenderbox}
\begin{corollary}\label{cor:branching}
\rev{Under the assumptions of \Cref{prop:rlb-unbiased}, the estimator \Cref{eq:rlb-sparse} is unbiased, also with the exact counterfactuals of any other SCM of the form \Cref{eq:scm} that is consistent with the MDP.}
\end{corollary}
\end{lavenderbox}

\begin{proof}
\rev{We expand the baselines as in \Cref{eq:group-expanded}, at every step $i$.
By \Cref{lem:posterior}, every branch is distributed as $p_\theta(\cdot|\hat{\bf x}_0)$ given $\hat{\bf x}_0$, so $\mathbb{E}[\sum_i\gamma^iG_i^{(j)}{\bf z}_i^{(j)}]=\nabla_\theta J(\theta)$ for every $j$ by \Cref{eq:reinforce}.
The cross terms $\mathbb{E}[G_i^{(\ell)}{\bf z}_i^{(j)}]$ with $\ell\neq j$ vanish as in the proof of \Cref{prop:rlb-unbiased}.
If $j\geq2$, the action of branch $j$ at step $i$ is drawn with fresh action noise, which is independent of its state ${\bf x}_i^{(j)}$ and of the other trajectories.
If $j=1$, \Cref{lem:posterior} lets us replace branch $\ell$ by the trajectory that uses the noise of the factual and fresh action noise, all of which are independent of the action noise of $\hat a_i$, as is $\hat{\bf x}_i$.
The second claim follows as in the proof of \Cref{cor:identifiability}.}
\end{proof}

\begin{lavenderbox}
\begin{proposition}\label{prop:branching}
\rev{Let the branches in \Cref{eq:rlb-sparse} be generated by a CWM, let scores be bounded by $z_{\max}$, and let the returns-to-go of a branch, generated or exact, and of its factual differ by $\leq A$ at every step.
Then the bias of \Cref{eq:rlb-sparse} is $\leq 2A\,z_{\max}\,n\,\eta_{\mathsf{TV}}$ in norm, where $\eta_{\mathsf{TV}}$ is the error \Cref{eq:eta-tv} of a branch from $\hat{\bf x}_0$, averaged over $\hat{\bf x}_0$.
For a maximum likelihood CWM as in \Cref{prop:rate}, with coverage along the branches from $\hat{\bf x}_0$, the bias is $\leq 2A\,z_{\max}\,n^2\sqrt{2\kappa(1+\log|\Omega|)/N}$ on average over the train set.}
\end{proposition}
\end{lavenderbox}

\begin{proof}
\rev{We take expectations as in the proof of \Cref{cor:branching}.
The cross terms with the score of a branch vanish for every model, as they only use that its actions are drawn with fresh action noise.
As in \Cref{eq:bias}, the bias is therefore the difference, between a generated and an exact branch, of the expectation of $\sum_i\gamma^iG_i^{(2)}{\bf c}_i$ with ${\bf c}_i\coloneqq\frac{m-1}{m}{\bf z}_i^{(2)}-\frac1m{\bf z}_i^{(1)}$.
Unlike in \Cref{prop:learned}, ${\bf c}_i$ depends on the state of the branch at step $i$, whose distribution depends on the model, so we bound the difference in total variation.
Given $\hat\tau$, the score ${\bf z}_i^{(2)}$ has zero mean for both branches, so $G_i^{(2)}$ can be replaced by $G_i^{(2)}-G_i^{(1)}$, and $(G_i^{(2)}-G_i^{(1)})\,{\bf c}_i$ is bounded by $A\,z_{\max}$ in norm.
Every step thus contributes $\leq 2A\,z_{\max}\,\eta_{\mathsf{TV}}$, and the first claim follows as $\gamma\leq1$.
The second claim follows from \Cref{eq:rollout} for $i=0$ and the bound on its terms in the proof of \Cref{prop:rate}.}
\end{proof}

\rev{Compared to \Cref{prop:rate}, the bound grows by a factor of at most two, and coverage is needed only along the branches from the first state.}

\rev{We finally discuss the variance.
Let $\widehat{\bf g}_i$ be the term of step $i$ in \Cref{eq:rlb-sparse} without the factor $\gamma^i$.
For any function $v$ of the state, with average $\bar v$ over the states of the group at step $i$, \Cref{eq:group-centered} gives
\begin{equation}\label{eq:drift}
    \widehat{\bf g}_i=\frac1{m-1}\sum_{j=1}^m\big(G_i^{(j)}-v({\bf x}_i^{(j)})\big)\big({\bf z}_i^{(j)}-\bar{\bf z}_i\big)
    +\frac1{m-1}\sum_{j=1}^m\big(v({\bf x}_i^{(j)})-\bar v\big)\,{\bf z}_i^{(j)}.
\end{equation}
The last sum is a drift term, as the baseline of a trajectory is built from returns at the states of the other trajectories.
It vanishes when the group shares a state, as when branching at every step, and it otherwise has zero mean, so it contributes to the variance but not to the bias.
It is present for RLOO with sparse branching as well, whose trajectories also differ through their noise.

The comparison of \Cref{prop:learned} still holds for the term of every step, with two changes.
First, $\widehat{\bf g}_i$ satisfies \Cref{eq:group-difference} with the returns-to-go and scores of step $i$, and these scores are uncorrelated with zero mean, so the proof of \Cref{prop:learned} gives $\mathbb{E}_\omega[\widehat{\bf g}_i\widehat{\bf g}_i^\top]\preceq A^2\bar S_i$, where $\bar S_i$ is the average of $\mathbb{E}[{\bf z}_i{\bf z}_i^\top]$ over the $m$ trajectories and $A$ includes the drift.
Second, for RLOO with any WM and sparse branching, and for REINFORCE with any state-dependent baseline, conditioning also on the noise of the real trajectory before step $i$ gives $\operatorname{Cov}(\widehat{\bf g}_i)\succeq\frac1{m^2}\,\mathbb{E}[(\sigma_{\mathsf{luck}}-\sigma_{\mathsf{inter}})_+^2\,{\bf z}_i{\bf z}_i^\top]$, where ${\bf z}_i$ is the score of the real trajectory and $\sigma_{\mathsf{luck}}$ and $\sigma_{\mathsf{inter}}$ are those of its state $\hat{\bf x}_i$.
}

\section{Extended related work}
\label{sec:related}
In this section, we provide an extended discussion of related work across domains.

\paragraph{Policy gradients with grouped baselines.}
\ours{} constructs learning signals by comparing multiple samples associated with a common input.
Its parent method, RLOO~\citep{kool2019buy,ahmadian2024back} uses the mean return of other independently sampled trajectories as a leave-one-out baseline.
Relatedly, GRPO~\citep{shao2024deepseekmath} normalizes rewards within groups of responses to the same prompt for clipped policy updates.
Yet, sharing a starting state or prompt does not itself couple environmental randomness.
We adapt RLOO baseline but generate counterfactual branches conditioned on a factual.
Given the close relationship between RLOO and GRPO~\citep{lan2026from}, we conjecture that our approach could be extended to GRPO, which we leave as future work.

\paragraph{Policy gradients with shared randomness.}
Our work is closely related with prior work that leverages common random numbers to reduce the variance of policy gradient estimators.
TRPO's ``vine'' procedure~\citep{schulman2015trust} compares actions through rollouts using common random numbers and requires simulator resets to sampled states.
Input-dependent baselines~\citep{mao2018variance} condition value predictions on observed future exogenous inputs.
Hindsight learning~\citep{sinclair2023hindsight} uses logged exogenous inputs and known dynamics to evaluate actions through hindsight planning.
\rev{Exogenous-state methods remove exogenous state variables and rewards when the reward decomposes additively~\citep{dietterich2018discovering}, or learn with a sample complexity that is nearly independent of an unknown exogenous part of the state that does not affect rewards~\citep{efroni2022sample}.}
rMC~\citep{bazerghi2025relative} estimates policy gradients for finite MDPs from return differences between a root trajectory and coupled true counterfactuals for all alternative actions run until they merge with the root.
We employ learned counterfactual dynamics, allowing policy learning without online simulator resets or direct online observation of environmental noise.

\paragraph{Exogenous inputs and states.} \rev{
Among the policy gradient methods above, those that use exogenous inputs or states are the closest to ours.
They are given the noise, as inputs~\citep{mao2018variance,sinclair2023hindsight} or as a part of the state~\citep{dietterich2018discovering,efroni2022sample}, whereas \ours{} only needs the factual observations, from which a CWM predicts the counterfactual transitions.
The noise thus need not be a separate input or a part of the state, nor behave additively.
If the noise is given as an input sequence, the baseline of \ours{} is a Monte Carlo estimate of the input-dependent baseline of \citet{mao2018variance}.
If the reward decomposes additively as in \citet{dietterich2018discovering}, \ours{} removes the exogenous return without discovering the decomposition.
In \citet{efroni2022sample}, the exogenous part of the state does not affect rewards and is to be ignored, whereas the noise that we consider affects returns.
Hindsight learning~\citep{sinclair2023hindsight} assumes known dynamics and lets a planner see the future inputs, which biases its values towards foresight, whereas the policy of \ours{} does not see the future noise, so its gradient is unbiased as long as CWM is exact.}

\paragraph{Hindsight and counterfactual credit assignment.}
Prior work has used hindsight and counterfactual predictions to distinguish action effects from luck.
HCA~\citep{harutyunyan2019hindsight} assigns credit using action probabilities conditioned on future observations.
CCA~\citep{mesnard2020counterfactual} learns value baselines from hindsight features, encouraging independence from current action.
QCA~\citep{mesnard2023quantile} compares return distributions at the quantile level and predicts it from future observations.
COMA~\citep{foerster2018counterfactual} uses a centralized critic to marginalize one agent's action while fixing the others'\rev{, and its baseline follows difference rewards~\citep{wolpert2002optimal}.
In our multi-agent bandit, where the other agents are the environmental noise of an agent, \ours{} is a sampled version of this baseline, and the learned CWM computes it from each agent's own observation without a centralized critic.}
We build baselines from generated counterfactual trajectories, without a separate critic that predicts returns, coupling transition outcomes through factual observations instead of e.g. return quantiles.
We note that an older preprint of CCA~\citep{mesnard2020counterfactualv1} has theorized a ``vine'' extension motivated by that of TRPO in Appendix~E.2, and separately theorized a model extension in Appendix~E.3 (see also Corollary~1 in the latest version), yet the precise combination of leave-one-out baseline with counterfactual models as in our work has not been considered, and their implemented methods are all model-free and use learned value baselines.

\paragraph{Counterfactual inference for policy search and evaluation.}
Similar to our CWMs, prior work has used observed experience to improve simulations of alternative decisions.
\citet{abbeel2006using} correct approximate dynamics using real trajectories for local policy improvement in nearly deterministic settings.
CF-GPS~\citep{buesing2018woulda} infers scenarios from logged episodes, simulates alternative policies, and improves policies through return-weighted regression.
It motivates factual conditioning when scenario distributions are harder to model than causal mechanisms, thereby inspiring our approach.
\citet{oberst2019counterfactual} reuse inferred Gumbel noise under alternative policies for counterfactual off-policy evaluation.
Distinguished from these approaches, we employ a grouped policy gradient estimator using learned counterfactual transitions, where both the factual and counterfactual contribute to each update.
\rev{In addition, CF-GPS is given the mechanisms and infers only the noise of a logged episode, whereas a CWM learns the counterfactual transition itself.}

\paragraph{Identifiability of counterfactuals.} \rev{
The transitions of an MDP do not identify its counterfactuals, as SCMs with the same transitions can disagree on them~\citep{oberst2019counterfactual}.
Prior work is therefore given the noise or the mechanisms~\citep{mao2018variance,sinclair2023hindsight,buesing2018woulda}, or restricts the mechanisms such that the transitions determine the counterfactuals, e.g., to Gumbel-max mechanisms~\citep{oberst2019counterfactual}.
\ours{} needs neither, as it uses counterfactuals only as a means to estimate the policy gradient, for which every SCM consistent with the MDP gives an unbiased estimate and the choice among them only affects the variance (\Cref{cor:identifiability}).}

\paragraph{Learned and generative counterfactual inference.}
Our approach is related to, and can benefit from, methods for inferring and learning counterfactual dynamics.
CoPhy~\citep{baradel2019cophy} predicts physical trajectories under altered initial conditions from factual experience, using paired factual and counterfactual training data similarly to our CWMs.
Filtered-CoPhy~\citep{janny2022filtered} extends this to video prediction without object position or physical property supervision, using paired videos.
We study a similar prediction problem in action-conditioned prediction setting, and further study modeling difficulty depending on environmental properties.
Diff-SCM~\citep{sanchez2022diffusion} and DCM~\citep{chao2023modeling} use diffusion-based latent inference to generate counterfactuals in class-conditional image generation and medical application, respectively.
CWMDT~\citep{shen2025counterfactual} conditions video diffusion on scene descriptions modified by a language model.
Distinctly, we apply diffusion counterfactual inference to an action-conditioned WM trained on interaction data.

\paragraph{Advantage functions.}
\rev{The weight $G^{(j)}-b^{(j)}$ that \ours{} gives to a trajectory estimates $G-V({\bf u})$, the return of the trajectory relative to the expected return of the policy under the same environmental noise.
It differs from the usual advantage, which is relative to the expected return over all noise, by the luck $V({\bf u})-\mathbb{E}[V({\bf u})]$, which a state-dependent baseline cannot remove (\Cref{lem:luck}).
This hindsight advantage is targeted by input-dependent baselines~\citep{mao2018variance} and future-conditional value functions~\citep{mesnard2020counterfactual} with a learned critic, whereas \ours{} estimates it in a critic-free manner.
If part of the noise is hidden, the baseline estimates $\mathbb{E}[V({\bf u})|\hat\tau]$ instead, which is a posterior value function~\citep{nota2021posterior} with the noise as the hidden variable.}

\section{Experiment details}
\label{sec:implementation}

\subsection{Contextual bandit}
\label{app:bandit}

\paragraph{Environment and agents.}
We adapt the setting of \citet{mesnard2020counterfactual}, using deterministic reward as outlined in \Cref{sec:bandit}.
The policy is a linear layer followed by softmax to give 21 action probabilities, and it receives the context ${\bf c}\in\{-10, ..., 10\}$ which is represented by a 21-dimensional one-hot vector.
For REINFORCE with baseline, we use a separate linear function of the context, fitted by squared loss.
For CCA, we follow \citet{mesnard2020counterfactual} to reproduce the method and apply it to our setting.
All methods use Adam with learning rate $4\times10^{-4}$ and entropy regularization of 0.004. Each training uses $10^7$ environmental interactions, with batch size 256 per update.

\paragraph{Offline model learning.}
For WM and CWM training, we use REINFORCE with baseline as the reference policy, with its action probabilities equally mixed with a uniform policy.
We create paired data using uniformly sampled contexts and two independent actions, and the same environmental noise.
We collect 4,096 training, 1,024 validation, and 2,048 test pairs, keeping both members of every pair in the same split.
This costs 14,336 environmental interactions per dataset.

Within next state ${\bf x} = ({\bf c}, a + {\bf u})$, both models are trained to predict $a + {\bf u}$ since ${\bf c}$ is already provided as input.
The models use the same paired dataset, where the ordinary WM treats both pair members as separate examples conditioned on context and action.
The counterfactual model trains on both directions, conditioned on one and predicting the other, and the other way around as well.

Each model has two hidden layers of eight SiLU units, Gaussian mean and scale outputs, and a learned linear connection from inputs to the mean.
This makes these models slightly smaller than the policy.
For both models, we clip the predicted Gaussian standard deviation to have a floor of 0.05, so that maximum likelihood does not explode for a constant prediction target.

Each model is trained with 5,000 Adam updates with 256 examples sampled with replacement per update.
We use cosine learning rate scheduler from $3\times10^{-3}$ to $3\times10^{-5}$.
After training, we use the final checkpoint and keep it frozen throughout policy learning.

\paragraph{Variance of RLOO and \ours{}.} \rev{We evaluate \Cref{prop:variance} and \Cref{lem:gap} for this problem, which has one step and no hidden noise.
For a context ${\bf c}$ and an action $a$ with mean $\bar a\coloneqq\mathbb{E}[a]$ under $\pi_\theta(\cdot|{\bf c})$, the return is $G=\sigma^2-({\bf c}-a-{\bf u})^2$.
Its hindsight value and the rest of the return are
\begin{equation*}
    V({\bf u})=\sigma^2-\mathbb{E}[({\bf c}-a)^2]+2({\bf c}-\bar a){\bf u}-{\bf u}^2,
    \qquad
    \tilde G=\mathbb{E}[({\bf c}-a)^2]-({\bf c}-a)^2+2(\bar a-a){\bf u},
\end{equation*}
so that $\sigma^2_{\mathsf{luck}}=2\sigma^4+4({\bf c}-\bar a)^2\sigma^2$ and $\sigma^2_{\mathsf{inter}}=4\sigma^2\max_a(a-\bar a)^2$.
Hence $\sigma^2_{\mathsf{luck}}$ grows as $\sigma^4$ and $\sigma^2_{\mathsf{inter}}$ only as $\sigma^2$, and the condition of \Cref{prop:variance} holds for large noise.
For any noise level, with $\operatorname{Cov}_{\bf u}(V({\bf u}),\tilde G)=4({\bf c}-\bar a)(\bar a-a)\sigma^2$, ${\bf g}({\bf u})-\mathbb{E}[{\bf g}({\bf u})]=-2{\bf u}\,\nabla_\theta\bar a$ and the score ${\bf z}_0$ of the action, \Cref{eq:gap} gives
\begin{align*}
    \operatorname{Cov}(\widehat{\bf g}_{\mathsf{RLOO}})-\operatorname{Cov}(\widehat{\bf g}_{\mathsf{\ours{}}})
    ={}&\frac{2\sigma^4+4({\bf c}-\bar a)^2\sigma^2}{m-1}\,\mathbb{E}\big[{\bf z}_0{\bf z}_0^\top\big]
    +\frac{8({\bf c}-\bar a)\sigma^2}{m}\,\mathbb{E}\big[(\bar a-a)\,{\bf z}_0{\bf z}_0^\top\big]\\
    &-4\Big(1-\frac{m-2}{m(m-1)}\Big)\sigma^2\,\nabla_\theta\bar a\,\nabla_\theta\bar a^\top.
\end{align*}
The first term grows as $\sigma^4$ and the others as $\sigma^2$.
Hence the covariance of RLOO grows as $\sigma^4$ and that of \ours{} only as $\sigma^2$, as measured in \Cref{tab:app:bias:bandit}, and the advantage of \ours{} widens with the noise level, as in \Cref{fig:bandit} (left).
The last term is the price of sharing one noise draw in a group (\Cref{sec:hurt}), which is why a moderate group size is best in \Cref{tab:m-sweep}.}

\subsection{Multi-agent bandit}
\label{app:multiagent}

\paragraph{Environment and agents.}
Based on the contextual bandit problem, we adapt the multi-agent setting of \citet{mesnard2020counterfactual}.
Contexts are independent across agents, and actions are sampled independently given the contexts and current policies.
Each agent observes only its own context before acting.
Each agent has its own policy and baseline with the architecture and optimization settings of
\Cref{app:bandit}, and parameters are not shared between agents.

\paragraph{Offline model learning.}
Similarly as in contextual bandit, we use the agents trained with REINFORCE with baseline as the reference policy, with each agent's policy mixed equally with uniform actions.
We generate data in groups, where each group contains a factual joint episode across the agents and one alternative for each agent with shared environmental randomness (all other agents).
Herein, each alternative resamples only that agent's action while retaining all contexts and other agents' factual actions.
This gives one pair per agent using $K+1$ joint environment interactions in total for all agents.
We collect 4,096 training, 1,024 validation, and 2,048 test groups per seed, keeping every observation from a group in the same split.

Each agent has its own ordinary and CWM which predicts the scalar terminal cost ${\bf x}^i=-r$. The ordinary model takes $({\bf c}^i,a^i)$, and the counterfactual model additionally takes the factual observation $o^i$ defined in \Cref{sec:multiagent}. Neither observes other agents' contexts, actions, or individual costs. Both use two hidden layers of 16 SiLU units and a learned linear connection to the mean, again yielding models slightly smaller than the policy.

The models use Gaussian densities restricted to $[0,400K]$, the possible joint-cost interval. These are continuous approximations to integer costs and the underlying Gaussian scale has a floor of 0.05 to prevent explosion.
As in contextual bandits, ordinary WMs train on both pair elements and counterfactual models are trained on both directions.
Each independent model receives 5,000 updates with 256 examples per update and the same Adam schedule as in \Cref{app:bandit}.

\subsection{The dinosaur game}\label{sec:dino}

\paragraph{Environment and agents.}
Chrome Dino~\citep{chromedino2014} is a side-scrolling platform game.
The dinosaur remains at a fixed horizontal location while obstacles move from right to left.
We use three actions $A = \{\texttt{stand}, \texttt{jump}, \texttt{duck}\}$.
A policy observes the four most recent decision frames in grayscale at
$128\times64$ and selects an action, repeating the action across four frames.

Since jumping is an almost universal solution in the distributed version of the game, we modify the game so that the three actions have distinct consequences.
Under our modification, cacti and low birds require a jump, middle birds require a duck, and high birds must be ignored by standing.
We additionally increase the frequency of birds and allow them to appear throughout an episode.
Game speed starts at 20 pixels per raw frame, increases by one every 20 raw frames, and is capped at 43.

The reward is $+1$ if an obstacle is passed, $-1$ upon collision, and zero
otherwise.
Training rollouts continue after a collision and have a fixed length of 20 decisions, or equivalently 80 raw frames.
During testing, evaluation terminates at the first collision or after 375 decisions.

A branching root is a state from which multiple policy rollouts are generated for RLOO and \ours{}.
We branch when the nearest obstacle satisfies
$8<\mathrm{TTC}\leq12$ raw frames, where
\[
\mathrm{TTC} =
\frac{\text{horizontal distance to the nearest obstacle}}
     {\text{game speed}}.
\]
All methods use the same convolutional policy.
Its three convolutional layers have 32, 64, and 64 channels, kernel sizes 8, 4, and 3, and strides 4, 2, and 1. We use orthogonal initialization, Adam with learning rate $10^{-3}$, gradient clipping at norm 20, discount $\gamma=0.99$, and entropy coefficient 0.04.
The learned-baseline control uses a separate value network with the same convolutional architecture and a scalar output, optimized with learning rate $3\times10^{-2}$.

For all methods, each update contains 32 trajectories.
\ours{} and RLOO arrange these as eight groups of $m=4$ trajectories.
Every method is trained for 2,000 updates and \rev{fifteen} seeds, giving 64,000 trajectories and 1.28 million rollout decisions per seed.

\paragraph{Offline model learning.}
The diffusion WM, which is also used for counterfactual world modeling, is trained on 1,000 recorded episodes of 150 decisions each, totaling 150,000 observed decisions.
Each frame is an RGB image with 320$\times$240 resolution.
Training examples contain eight preceding frames, the single causal action, and the next frame.
A pretrained VAE maps frames to latent space.
We jointly train an action embedding and a 101M-parameter conditional U-Net diffusion model using the velocity prediction objective.
Training uses AdamW with learning rate $2\times10^{-4}$, gradient clipping at 1, batch size \rev{32}, \texttt{bfloat16} precision, and EMA with warmup and decay 0.9999.
We apply classifier-free guidance~\citep{ho2022classifier} by dropping the action with probability 0.2 during training.
All experiments use the EMA checkpoint after 9,000 steps.

We generate each frame with 10 DDIM~\citep{song2020denoising} noise inversion steps and with guidance scale 1.0.
The generated frame is appended to the context, so a policy rollout is generated autoregressively for 20 decisions.
A frozen visual classifier maps each pair of consecutive generated frames to collision, no-event, or obstacle-clear reward.
For policy input, generated frames are converted to grayscale and resized to $128\times64$.

\rev{
For \ours{}~+~CWM, at every step, DDIM inversion uses the corresponding factual
eight-frame context, factual action, and factual next frame to abduct factual noise.
This noise is shared across the $m-1$ imagined group members, while each member conditions on its sampled action and recursively generated context.
RLOO~+~WM instead uses independent fresh Gaussian noise.
}

\paragraph{Validation and testing.}
For policy evaluation, we save a checkpoint every 100 policy updates.
For each method, every checkpoint is evaluated on 30 validation episodes for each training seed.
We select one checkpoint index using its mean validation performance across the \rev{fifteen} seeds.
The selected checkpoint is then evaluated once on 100 test episodes per seed.
Validation and test episodes use fixed and different simulator seeds that are shared by all methods.
All policies predict action greedily in the real simulator.
An episode ends at the first collision or after 375 decisions.
We report the fraction of encountered obstacles that cause a collision and the mean number of decisions survived before the first collision.
Survival is capped at 375 decisions.
The main table reports the mean and sample standard deviation across
the \rev{fifteen} training seeds.

To evaluate model imaginations, we follow the criteria of \citet{melistas2024benchmarking} for assessing counterfactuals.
For each root we retain the factual action and uniformly select one of the other two actions as the intervention.
CWM and WM therefore receive exactly the same contexts and intervention actions.
CWM uses the DDIM-inverted factual noise and WM uses fresh Gaussian noise.
Composition is the mean pixel $\ell_1$ error, on the 0--255 scale, between the recorded next frame and a generation made under the factual action.
Effectiveness is the macro-F1 score attained by a trained action predictor, i.e. an inverse dynamics model, applied to the generated counterfactual.
Realism is FID against the recorded next frames.
Minimality is counterfactual latent divergence (CLD), computed with an independently trained unconditional VAE.

We additionally use a Dino-specific minimality measure for newly entering obstacles.
Existing obstacles are visible in the context frames, whereas a newly entering obstacle is not and must be generated by the model.
Using only recorded frames, we select transitions where a new obstacle appears in the fixed right-edge region marked by the red dashed boxes.
The obstacle-preservation score is the mean pixel $\ell_1$ error between the generated counterfactual and the recorded next frame within this region, measured on the 0--255 scale.

\section{Supplementary results}
\label{sec:additional}

\rev{We present supplementary results we could not include in the main text due to space limits.}

\subsection{Analysis of bias and variance}
\label{app:bias}

\rev{We empirically analyze the bias and variance of the policy gradient estimators, in particular RLOO and \ours{} under the use of learned (counterfactual) world models.}

\begin{table}[t!]
\centering
\small
\caption{\rev{Gradient variance and bias of contextual bandit at a uniform policy. - denotes unbiased.}}
\label{tab:app:bias:bandit}
\begin{tabular}{lrrrccc}
\toprule
& \multicolumn{3}{c}{Variance} & \multicolumn{3}{c}{Relative bias} \\
\cmidrule(lr){2-4} \cmidrule(lr){5-7}
Method & $\sigma=0.1$ & $\sigma=30$ & $\sigma=60$ & $\sigma=0.1$ & $\sigma=30$ & $\sigma=60$ \\
\midrule
REINFORCE            & 95.8 & 21,122 & 301,493 & - & - & - \\
REINFORCE + baseline & 95.4 & 20,995 & 301,304 & - & - & - \\
CCA                  & 36.8 & 8,465 & 121,238 & 0.48 & 2.20 & 8.70 \\
RLOO                 & 70.6 & 18,794 & 268,276 & - & - & - \\
RLOO + WM            & 70.5 & 18,539 & 265,281 & 0.004 & 0.52 & 1.79 \\
\ours{}                  & 70.6 & \textbf{1,551} & \textbf{5,989} & - & - & - \\
\ours{} + CWM            & 70.6 & \textbf{1,551} & \textbf{5,987} & \textbf{0.002} & \textbf{0.001} & \textbf{0.003} \\
\bottomrule
\end{tabular}
\end{table}

\begin{table}[t!]
\centering
\small
\setlength{\tabcolsep}{4pt}
\caption{\rev{Gradient variance and bias of Dino game. Brackets are 95\% confidence interval.}}
\label{tab:app:bias:dino}
\begin{tabular}{rccccccc}
\toprule
& \multicolumn{2}{c}{Variance} & \multicolumn{2}{c}{Variance (with learned models)} & \multicolumn{2}{c}{Bias$^2$ (with learned models)} \\
\cmidrule(lr){2-3} \cmidrule(lr){4-5} \cmidrule(lr){6-7}
Updates & \ours{} & RLOO & \ours{}~+~CWM & RLOO~+~WM & \ours{}~+~CWM & RLOO~+~WM & \\
\midrule
500 & \textbf{43.8} & 59.6 & 61.2 & \textbf{60.3} & 1.05 [$+$0.34, $+$2.12] & \textbf{0.00} [$-$0.18, $+$0.35] & \\
1,000 & \textbf{23.8} & 59.1 & \textbf{54.2} & 89.2 & \textbf{0.02} [$-$0.10, $+$0.24] & 0.25 [$-$0.21, $+$1.19] & \\
1,500 & \textbf{27.0} & 63.3 & \textbf{50.5} & 79.7 & \textbf{0.05} [$-$0.08, $+$0.31] & 0.37 [$-$0.04, $+$1.00] & \\
\bottomrule
\end{tabular}
\end{table}

\paragraph{Contextual bandit.}
\rev{
We compare all methods in the contextual bandit of \Cref{sec:bandit} in terms of variance and size of the bias relative to that of the gradient. We fix the same uniform policy and use the same budget of 256 transitions per batch across all methods, using five seeds. Because the true policy gradient can be computed exactly in this environment, the bias can be measured directly.

The results are in \Cref{tab:app:bias:bandit}.
With almost zero environmental noise ($\sigma=0.1$), \ours{} and RLOO have the same variance, because there is little noise to share. As the noise grows, \ours{} has the lowest variance of all methods, 12--14$\times$ lower than RLOO and REINFORCE at $\sigma=30$, and 45--50$\times$ lower at $\sigma=60$. \ours{}~+~CWM keeps this reduction, while RLOO~+~WM does not, presumably due to the difficulty of learning the WM (\Cref{sec:bandit}). CCA and RLOO~+~WM are biased, and their bias grows with the noise, whereas \ours{}~+~CWM is empirically nearly unbiased (relative bias $\leq$0.003).

We further measure the bias of \ours{}~+~CWM during CWM training, at the final \ours{}~+~CWM policy rather than the uniform policy of \Cref{tab:app:bias:bandit}. At $\sigma=30$, the relative bias is 1.0 at initialization, briefly rises to 1.43 for an intermediate model, and falls to 0.0025 for the fully trained CWM.
The result shows that, as CWM becomes more accurate through training, it reduces the bias of policy gradient estimation, consistent with our \Cref{prop:learned,prop:rate}.
}

\paragraph{Dino.}
\rev{
In Chrome Dino (\Cref{sec:exp-dino}), the true policy gradient is unknown, so we estimate the bias of \ours{}~+~CWM and RLOO~+~WM by respectively comparing against the simulator \ours{} and RLOO, both of which are unbiased, using 640 pairs and subtracting the contribution of sampling noise. For comparisons, we take three policies learned with 500, 1,000, and 1,500 updates of REINFORCE with a learned baseline, and use one checkpoint of each learned world model.
}

\rev{The results are in \Cref{tab:app:bias:dino}.
Learned CWM barely changes the direction of the gradient, as its estimated gradient points in nearly the same direction as the unbiased gradient of \ours{} with simulator (cosine similarity 0.993--0.996). A bias is detectable only at update 500, where the squared bias (1.05) is still below 2\% of the squared error of a single batch. At updates 1,000 and 1,500, the confidence intervals of the squared bias include zero, and even their upper limits (0.24 and 0.31) are below 1\% of the squared error of a single batch.

We find that the learned CWM mainly increases variance. \ours{}~+~CWM has around 1.4--2.3$\times$ the variance of simulator \ours{}, which has the lowest variance of the four estimators at every update. It has variance comparable to simulator RLOO at update 500 and lower at updates 1,000 and 1,500, while using only one real trajectory per group instead of four. It also has 37--39\% lower variance than RLOO~+~WM at updates 1,000 and 1,500.}

\subsection{Analysis of the group size}
\label{app:m-sweep}

\begin{table}[!t]
\centering
\small
\caption{\rev{Gradient variance in the contextual bandit for different group sizes $m$, averaged over five seeds. Bold marks the lowest \ours{} variance within sampling error.
}}
\label{tab:m-sweep}
\begin{tabular}{llrrrrrr}
\toprule
& & \multicolumn{6}{c}{$m$} \\
\cmidrule(l){3-8}
$\sigma$ & Method & 2 & 4 & 8 & 16 & 32 & 64 \\
\midrule
30 & \ours{}  & 2,155 & \textbf{1,544} & \textbf{1,539} & 1,866 & 2,649 & 4,242 \\
   & RLOO & 28,102 & 18,761 & 16,113 & 15,051 & 14,534 & 14,394 \\
\midrule
60 & \ours{}  & 8,318 & \textbf{5,964} & \textbf{5,946} & 7,214 & 10,247 & 16,426 \\
   & RLOO & 400,882 & 267,586 & 229,721 & 214,363 & 206,629 & 203,740 \\
\bottomrule
\end{tabular}
\end{table}

\begin{table}[t!]
\centering
\small
\caption{\rev{Expected reward of \ours{} and RLOO after 10M interactions in the contextual bandit, across group sizes $m$. Mean and 95\% normal confidence interval over six seeds. The optimum is zero.}}
\label{tab:m-sweep-learning}
\begin{tabular}{llcccc}
\toprule
$\sigma$ & Method & $m=2$ & $m=4$ & $m=8$ & $m=16$ \\
\midrule
30 & \ours{}  & $-$16.26 $\pm$ 0.24 & \textbf{$-$15.01 $\pm$ 0.30} & $-$16.58 $\pm$ 0.31 & $-$18.87 $\pm$ 0.11 \\
   & RLOO & $-$41.83 $\pm$ 0.19 & $-$38.94 $\pm$ 0.14 & $-$37.83 $\pm$ 0.09 & $-$37.30 $\pm$ 0.07 \\
\midrule
60 & \ours{}  & $-$28.73 $\pm$ 0.25 & \textbf{$-$25.49 $\pm$ 0.18} & $-$25.83 $\pm$ 0.19 & $-$28.00 $\pm$ 0.24 \\
   & RLOO & $-$59.16 $\pm$ 0.48 & $-$56.92 $\pm$ 0.44 & $-$56.06 $\pm$ 0.11 & $-$55.58 $\pm$ 0.19 \\
\bottomrule
\end{tabular}
\end{table}

\rev{We measure how group size $m$ affects gradient variance in the contextual bandit of \Cref{sec:bandit}, at uniform policy and with exact counterfactuals. We fix the budget at 256 trajectories per batch, so a batch contains $256/m$ groups, and estimate the variance from 4,000 batches for each of five seeds.

\Cref{tab:m-sweep} shows that the variance of \ours{} is lowest at $m=4$ to $8$ and
increases for larger groups, whereas that of RLOO keeps decreasing with $m$. This
matches the analysis of \Cref{sec:hurt}. A larger group reduces the terms that
decay with $m$, but leaves fewer groups to average the variation of the target
with the noise draw that each group shares. \ours{} has a lower variance than RLOO
at every tested $m$, by a factor of 3.4 to 13.0 at $\sigma=30$ and 12.4 to 48.2
at $\sigma=60$.

We also train \ours{} with exact counterfactuals and RLOO for each group size. \Cref{tab:m-sweep-learning} reports the exact
expected reward at the end of training. The group size $m=4$ used in our
experiments gives \ours{} the highest final reward at both noise levels. RLOO instead improves steadily with
$m$, in line with its variance, but even its best group size remains far behind
\ours{} at every group size. Together, these results support a moderate group size
for \ours{}.}

\subsection{Multi-agent bandit}
\label{app:mult_base}

\rev{We include supplementary results for multi-agent bandit (\Cref{sec:multiagent}).}

\begin{figure*}[t!]
\centering
\includegraphics[width=\textwidth]{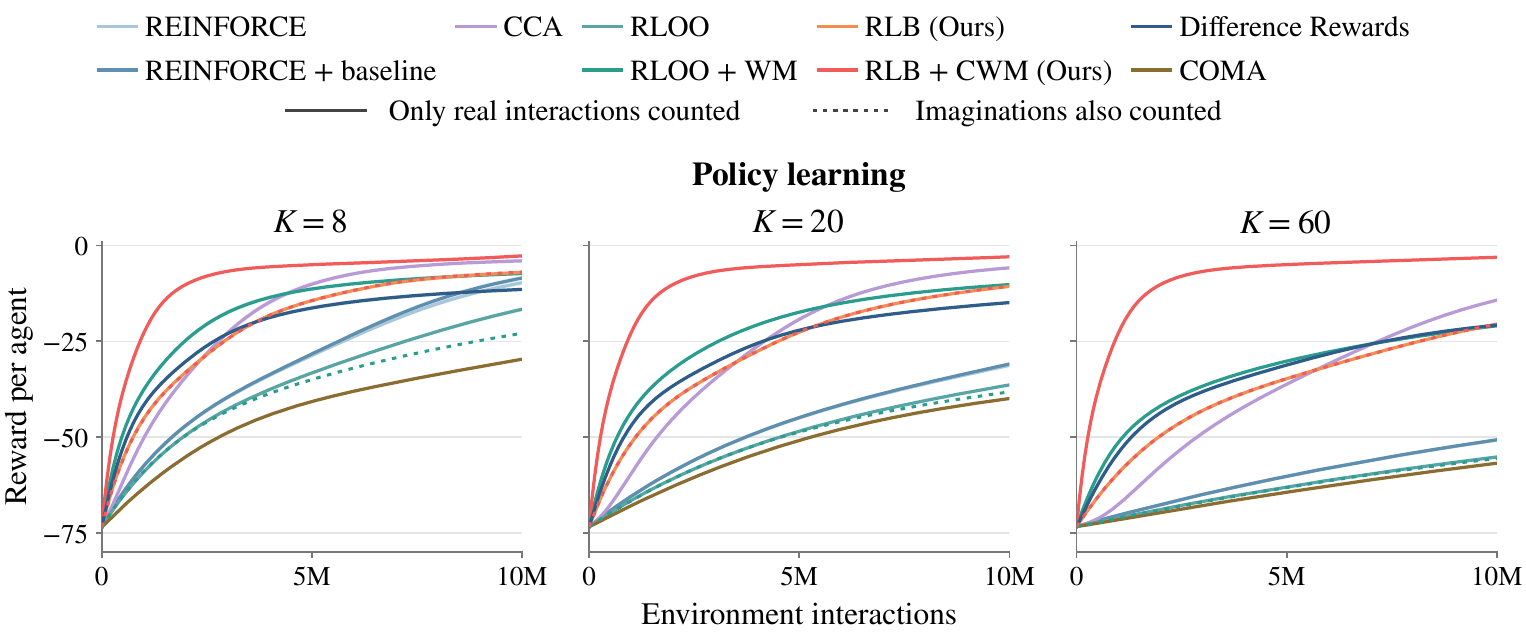}
\caption{\rev{Policy learning results for the multi-agent bandit problem, with additional baselines difference rewards~\citep{wolpert2002optimal} and COMA~\citep{foerster2018counterfactual}.}}
\label{fig:multiagent-additional-baselines}
\end{figure*}

\paragraph{Additional baselines.}
\rev{
We further compare with difference rewards~\citep{wolpert2002optimal} and counterfactual multi-agent (COMA) policy gradients~\citep{foerster2018counterfactual}, both of which use multi-agent counterfactual baselines (\Cref{sec:related}). To account for the simulator use by the simulator-based estimators, we count every factual and counterfactual reward evaluation as an environment interaction. As a result, with $K$ agents in the environment, each episode costs $1+K$ interactions for difference rewards and $1+20K$ for COMA, whereas each \ours{} with group size four costs $1+3K$.

The results are in \Cref{fig:multiagent-additional-baselines}.
At 10M environment interactions, \ours{} outperforms difference rewards for $K=8$ and $K=20$ and performs competitively for $K=60$. It outperforms COMA across every $K$. When only real environment interactions for policy training are counted, \ours{}~+~CWM achieves the highest return among the methods shown across all three settings. When generated samples are also counted, \ours{}~+~CWM closely matches the simulator-based \ours{}.

Exogenous-reward removal~\citep{dietterich2018discovering} is also related.
Its discovery procedure uses state transitions to identify exogenous state
components and estimate their reward contribution. Our one-step multi-agent
bandit does not provide this transition structure, and each agent observes
only its own context before acting. Applying the method to the additional
post-action feedback would require adapting its discovery criterion.
We therefore do not include it as a direct reproduction.}

\subsection{The dinosaur game}
\label{app:dino-actor-critic}

\rev{We include supplementary results for Chrome Dino (\Cref{sec:exp-dino}).}

\begin{figure*}[t!]
\centering
\includegraphics[width=\textwidth]{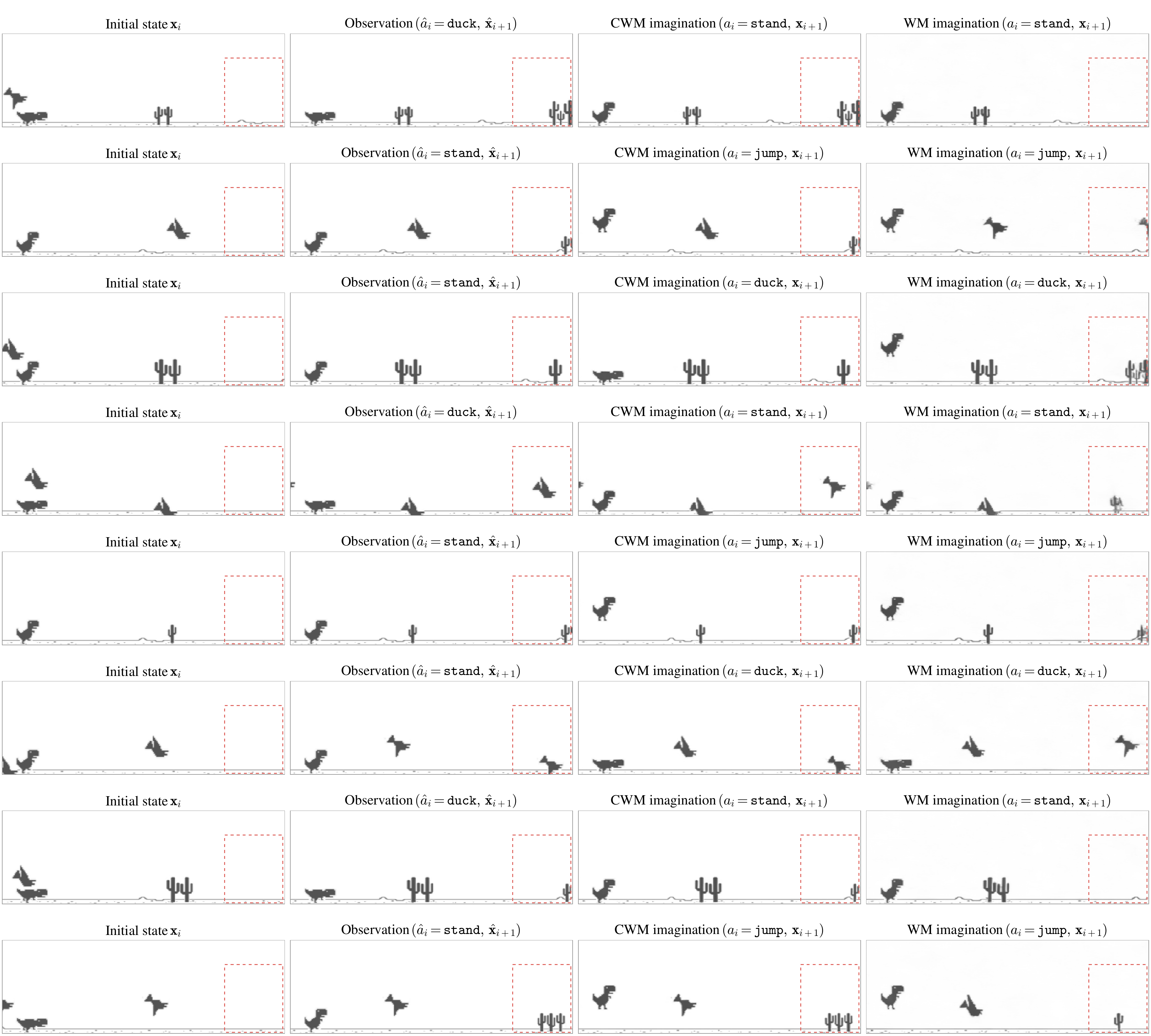}
\caption{Additional examples.
CWM and WM use the same intervention action.}
\label{fig:app:dino:examples1}
\end{figure*}

\paragraph{Qualitative examples.}
\Cref{fig:app:dino:examples1} shows eight additional qualitative examples of CWM and WM generations using different factual and intervention actions.

\paragraph{Additional baselines.}
\rev{We add two actor-critic baselines, advantage actor-critic (A2C)~\citep{mnih2016asynchronous} and proximal policy optimization (PPO)~\citep{schulman2017proximal}, using the same policy architecture as the other methods in \Cref{sec:exp-dino}. We train each policy for 2,000 steps over fifteen seeds and select a checkpoint using a separate validation set. We evaluate each policy greedily on 100 episodes shared across methods. As a result, we find that A2C obtains a collision rate (\%) of 7.07 $\pm$ 4.92, whereas PPO obtains 3.27 $\pm$ 0.86. For comparison, \ours{} obtains 3.29 $\pm$ 0.36 (\Cref{tab:dino:policy}). PPO and \ours{} thus compete in mean collision rate, while \ours{} exhibits improved stability across seeds.}

\paragraph{Impact of diffusion inversion hyperparameters.}

\begin{table*}[t!]
\centering
\caption{\rev{Impact of diffusion inversion steps and guidance scale.}}
\label{tab:ddim-ablation}
\resizebox{\textwidth}{!}{%
\begin{tabular}{cccccc}
\toprule
DDIM steps
    & Guidance
    & Composition (\%) $\downarrow$
    & Failure (\%) $\downarrow$
    & Effectiveness (\%) $\uparrow$
    & Relative runtime $\downarrow$ \\
\midrule
5
    & 1.0
    & 0.371
    & 34.4
    & 99.5
    & 0.55$\times$ \\
\textbf{10}
    & \textbf{1.0}
    & \textbf{0.197}
    & \textbf{9.9}
    & \textbf{99.8}
    & \textbf{1.00$\times$} \\
20
    & 1.0
    & 0.121
    & 3.9
    & 99.8
    & 1.86$\times$ \\
\midrule
10
    & 1.5
    & 0.202
    & 11.4
    & 100.0
    & 1.01$\times$ \\
10
    & 2.0
    & 0.210
    & 12.2
    & 100.0
    & 1.01$\times$ \\
10
    & 3.0
    & 0.226
    & 16.1
    & 100.0
    & 1.06$\times$ \\
\bottomrule
\end{tabular}%
}
\end{table*}

\rev{We evaluate the sensitivity of the diffusion counterfactual inference to the number of DDIM inversion steps and the classifier-free guidance~(CFG) scale for action conditioning~\citep{sanchez2022diffusion}. We perform evaluation using a fixed set of 100 held-out starting states. From every state, we generate one factual rollout and three counterfactual rollouts over eight policy decisions.

We measure three metrics. Composition and effectiveness are defined as in
\Cref{tab:dino:cfquality} and \Cref{sec:implementation}, with composition given as a percentage of the [0, 255] pixel range. Unlike in \Cref{tab:dino:cfquality}, we evaluate rollouts rather than single transitions, because the CWM generates its branches autoregressively during policy training, so the inversion settings matter through how errors accumulate along a rollout. Composition is therefore averaged over the generated frames of the factual rollout, each generated from the previous generated frames, and its values are higher than for the single transitions of \Cref{tab:dino:cfquality}.
We also report failure, the percentage of generated frames with at least one missing or invented obstacle. We add it because a small obstacle can disappear while changing the average pixel error only slightly.

As shown in \Cref{tab:ddim-ablation}, increasing from five to ten DDIM steps reduces the composition error from 0.371\% to 0.197\% and failure from 34.4\% to 9.9\%. Twenty steps further reduce these errors but require 1.86$\times$ the runtime of ten steps. We find that increasing the guidance scale beyond 1.0 provides no meaningful improvement in effectiveness, which is already nearly perfect, and increases both errors. Given the computational cost of running the CWM repeatedly during policy training, we use ten DDIM steps and a guidance scale of 1.0 as the best quality--runtime trade-off.

}

\paragraph{Factual action information in counterfactuals.}
\rev{
While exact diffusion inversion would recover latent noise independent of the action underlying the factual (\Cref{sec:counterfactual-models}), and therefore yield unbiased policy gradient estimate in \ours{} (\Cref{prop:rlb-unbiased}), diffusion inversion is not exact in practice, which means the information of the action can leak into the baseline and introduce bias (\Cref{prop:learned}). To test if this happens in practice, we train a small nonlinear action classifier to predict the factual action from the inverted noise, and also the generated counterfactual.
By training an action classifier on the inverted noise using 600 training data, 300 validation data, and 3,000 unseen test data, we find that the factual action is recoverable with 98\% test accuracy, whereas the chance level is 33\%. Thus, the inverted latent does retain substantial information about the factual action.
}

\rev{
While this may appear as inducing substantial bias in \ours{}, we remark that what matters for the bias is whether this information reaches the counterfactual branches, as \Cref{prop:learned,prop:rate} predict that the bias is bounded by the error of counterfactuals and their returns. Therefore, we additionally train the same classifier to predict the factual action from generated counterfactuals. For the same data, we draw a new action uniformly from three choices and generate the counterfactual from the inverted noise with this action. We then give the classifier the counterfactual, the last context frame, and the new action. Its test accuracy is 53.4\% $\pm$ 0.8\% over five seeds, no higher than the 54.8\% $\pm$ 0.6\% of a classifier that sees only the context frame and the new action\footnote{\rev{Both are above chance because the agent's action depends on the scene.}}.
Thus, the classifier detects no information of factual action in the generated counterfactual beyond the context and new action, consistent with the small overall bias of \ours{}~+~CWM measured in \Cref{app:bias}.
}

\begin{table}[t!]
\centering
\small
\caption{\rev{Wall-clock time cost of Dino model and policy training on RTX 3090 GPUs.}}
\label{tab:app:dino-wallclock}
\begin{tabular}{lccc}
\toprule
Method or stage & Workload & GPUs & Time cost \\
\midrule
Diffusion WM training
    & 9,000 steps
    & 8
    & 2 h 15 min \\
\midrule
REINFORCE
    & 100 updates
    & 1
    & 7.48 min \\
REINFORCE + baseline
    & 100 updates
    & 1
    & 7.59 min \\
RLOO
    & 100 updates
    & 1
    & 7.61 min \\
\ours{}
    & 100 updates
    & 1
    & 7.33 min \\
RLOO + WM
    & 100 updates
    & 1
    & 58.91 min \\
\ours{} + CWM
    & 100 updates
    & 1
    & 74.95 min \\
\bottomrule
\end{tabular}
\end{table}

\rev{\paragraph{Training cost.}
We measure end-to-end wall-clock time cost on NVIDIA RTX 3090 GPUs. Policy training times are measured by running each method for 100 updates on one GPU. They include environment simulation, policy optimization, and online WM or CWM inference. In terms of diffusion model evaluations, each policy update of RLOO~+~WM and \ours{}~+~CWM contains eight groups, each with one factual trajectory and three 20-step generated branches. Both methods therefore generate 480 frames per update, which requires 4,800 U-Net evaluations. \ours{}~+~CWM additionally uses ten DDIM diffusion inversion steps per frame, and inverts the 160 factual transitions, requiring 6,400 U-Net evaluations per update in total. The diffusion WM is trained once offline for 9,000 optimization steps using eight GPUs, which takes a modest time of 2 hours and 15 minutes. As shown in \Cref{tab:app:dino-wallclock}, the simulator-based methods require approximately 7--8 minutes per 100 policy updates, whereas RLOO~+~WM and \ours{}~+~CWM require 58.9 and 75.0 minutes, respectively. As such, the empirical cost overhead of CWM compared to WM due to DDIM inversion is around 27\%. We leave faster diffusion based on few-step distillation~\citep{wan2025wan} as future work.}

\subsection{Linear key-to-door \& task interleaving}

\begin{figure*}[!t]
\centering
\includegraphics[width=\textwidth]{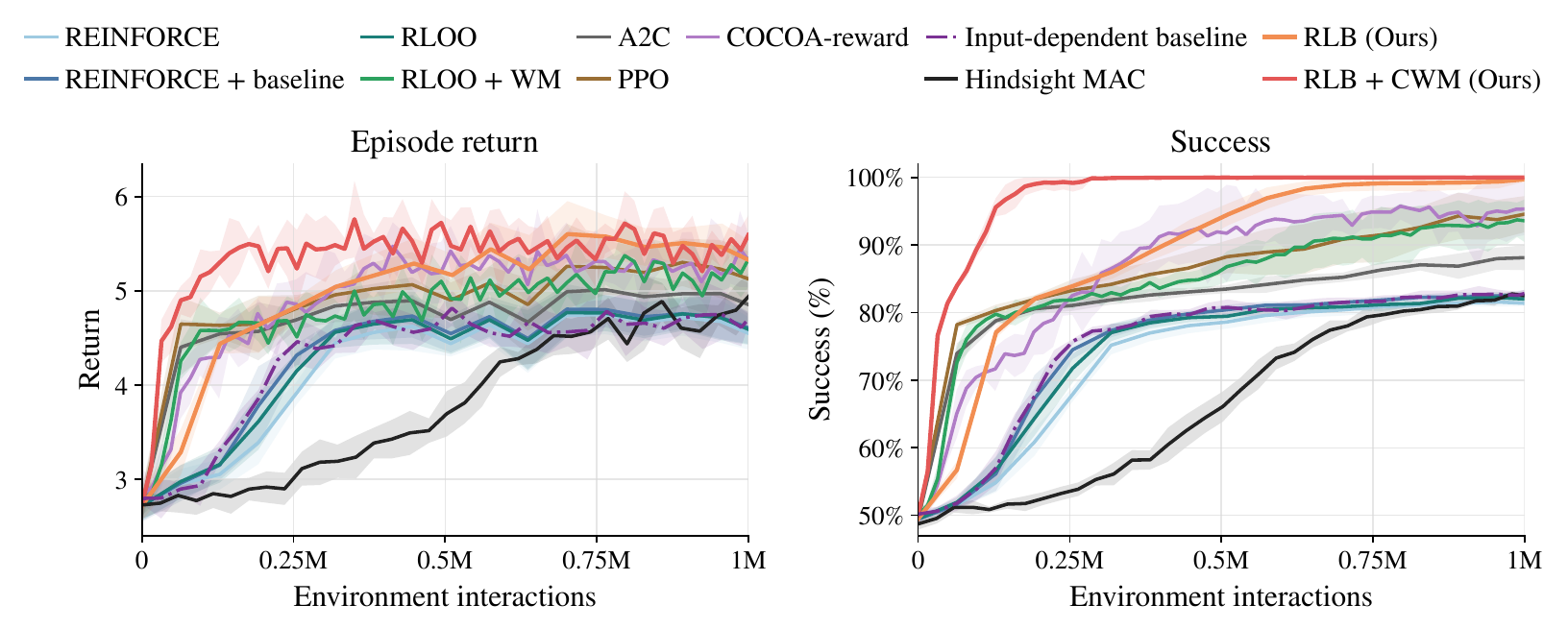}
\caption{\rev{Results for the task interleaving environment across five randomized trials.}}
\label{fig:app:task-interleaving}
\end{figure*}
\begin{figure*}[!t]
\centering
\includegraphics[width=\textwidth]{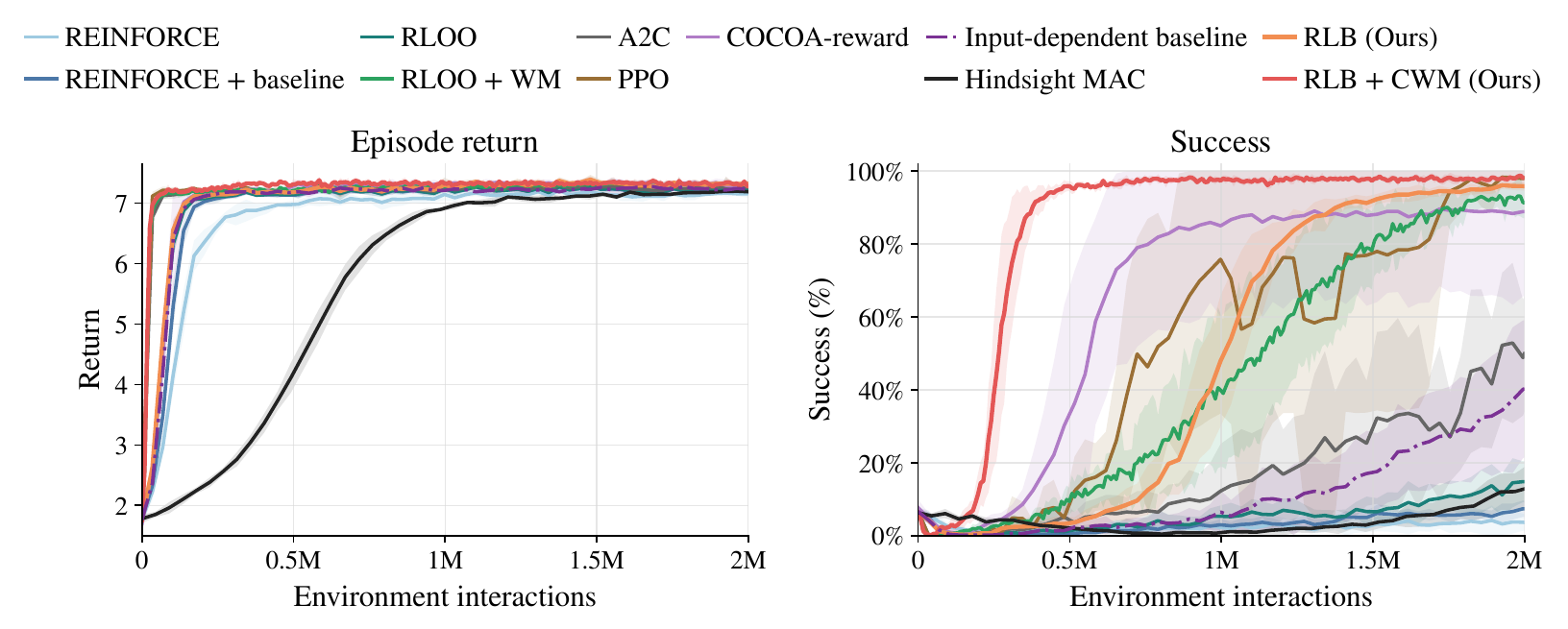}
\caption{\rev{Results for the linear key-to-door environment across five randomized trials.
}}
\label{fig:app:key-to-door}
\end{figure*}

\rev{To expand our experiments to more complex sequential environments, we leverage the linear key-to-door and task interleaving environments introduced by \citet{mesnard2020counterfactual}, using the configurations of \citet{meulemans2023would}. These environments require credit assignment across multiple intervening decisions and stochastic rewards. In addition to the main methods, we include actor-critic methods A2C and PPO, as well as COCOA-reward~\citep{meulemans2023would}, which learns how much each action contributes to each later reward, input-dependent baseline~\citep{mao2018variance} which computes a baseline conditioned on future exogenous inputs, and Hindsight Mean Actor Critic (MAC)~\citep{sinclair2023hindsight}, which computes state-action values conditioned on future exogenous inputs using exact dynamic programming and access to the environment dynamics.

\paragraph{Environments.}
In linear key-to-door, the agent traverses a sequence containing a key, stochastic distractor rewards, a door, and a terminal treasure. Collecting the treasure requires picking up the key and subsequently opening the door, while unrelated rewards are observed between these two decisions.
In task interleaving, the agent makes choices in several contexts whose rewards are delayed and interleaved. In each context, it chooses between a rewarding and a non-rewarding object, but observes the outcome only when that context is revisited. Meanwhile, it may make choices and receive rewards in other contexts. The agent must therefore assign each delayed reward to the decision that produced it. We use the configuration of \citet{meulemans2023would} with five contexts, two objects per context, at most three simultaneously open contexts, a maximum episode length of 100, and termination probability of 0.05.
In both environments, the noise is not additive.
In key-to-door, a distractor reward has a random value and is received only if the agent picks it up, and in task interleaving, the random placement of the objects decides which choice is rewarded.

\paragraph{Setup.}
We run five seeds for each method and report learning as a function of environment interactions. Each policy is evaluated over 512 episodes using stochastic action selection, which is 22,016 transitions in Key-to-Door, where every episode lasts 43 steps, and about 10,200 transitions in Task Interleaving, where episodes last 19.9 steps on average.
In task interleaving, REINFORCE, REINFORCE with a learned baseline, RLOO, A2C, PPO, simulator-based \ours{}, the input-dependent baseline, and Hindsight MAC use 32 factual trajectories per update. COCOA-reward, RLOO~+~WM, and \ours{}~+~CWM use eight factual trajectories per update and therefore perform approximately four times as many policy updates. We compare the methods through approximately 1M interactions. In key-to-door, simulator-based methods use eight factual trajectories per update, whereas RLOO~+~WM and \ours{}~+~CWM use two, each expanded into a group of $m=4$, and therefore perform about four times as many updates. We compare them through approximately 2M interactions.
The task interleaving WM and CWM are trained offline on 39,672 training transitions collected using a separately trained reference policy. The key-to-door paired dataset contains 77,056 transitions, comprising 44,032 training transitions and 33,024 held-out validation and test transitions.
Hindsight MAC additionally queries a white-box planner when computing its policy updates, using approximately 12 additional transitions per factual transition in key-to-door and 8.24 in task interleaving.

\paragraph{Results.}
In task interleaving, as shown in \Cref{fig:app:task-interleaving}, \ours{}~+~CWM obtains 99.99 $\pm$ 0.01\% success, and simulator-based \ours{} obtains 99.70 $\pm$ 0.36\% at approximately 1M interactions. Both outperform the remaining methods. RLOO~+~WM reaches 93.58 $\pm$ 3.14\%, compared with 82.01 $\pm$ 0.51\% for RLOO, showing that policy learning with (counterfactual) world models improve environment interaction efficiency by replacing the branches in the group with imaginations.

In key-to-door, as shown in \Cref{fig:app:key-to-door}, \ours{}~+~CWM obtains the highest mean success 98.16 $\pm$ 0.70\%, followed closely by PPO at 98.10 $\pm$ 1.91\%. Simulator-based \ours{} reaches 95.83 $\pm$ 1.20\%, while COCOA-reward exhibits considerably greater seed-to-seed variability.

For the remaining methods, the input-dependent baseline obtains 40.3 $\pm$ 18.8\% in key-to-door and 82.8 $\pm$ 0.6\% in task interleaving.
Hindsight MAC obtains 12.8 $\pm$ 6.0\% and 82.5 $\pm$ 0.4\%, respectively, where we find that the requirement for an online white-box environment access for planning makes the method inefficient in terms of environment interaction efficiency.

Overall, these results show that \ours{} is effective in sequential credit assignment problems with complexity beyond bandits, and that a learned CWM can reduce the environment interactions required for learning long-term dependencies during policy learning, extending our findings in \Cref{sec:experiments}.}

\subsection{Sensitivity to the choice of SCMs}
\label{coupling}

\begin{figure*}[!t]
\centering
\includegraphics[width=0.7\textwidth]{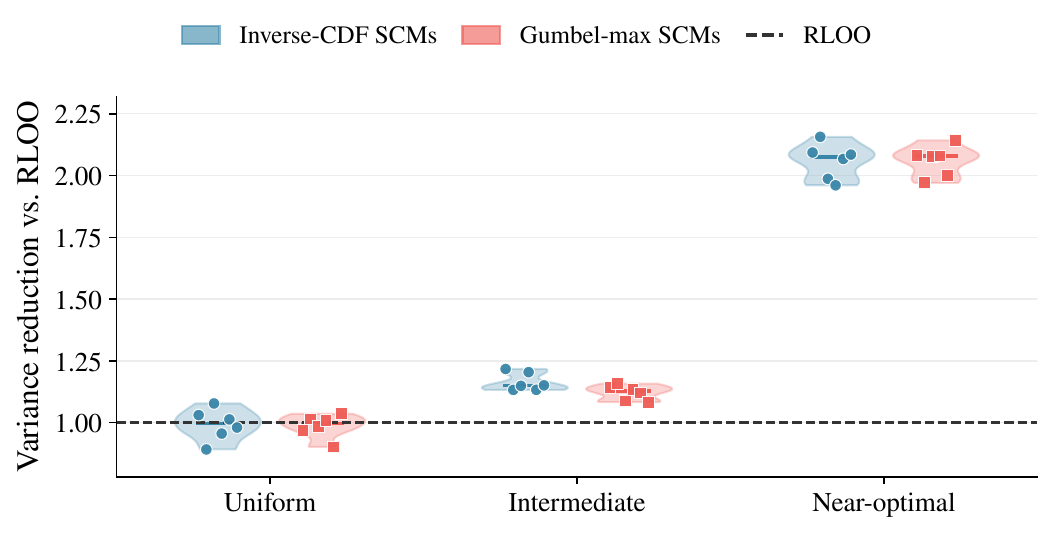}
\caption{\rev{Sensitivity to the choice of SCMs in RiverSwim.
}}
\label{fig:app:coupling}
\end{figure*}

\rev{Different SCMs can induce the same MDP while coupling counterfactual outcomes differently. By \Cref{cor:identifiability}, the choice does not bias \ours{}, but it can change its variance. We test the extent of this in the RiverSwim benchmark~\citep{strehl2008analysis}, which has many compatible SCMs.

RiverSwim has six states in a row, and the agent starts in the leftmost one. It can swim left or right. Swimming left moves it left with probability 0.9 and keeps it in place with probability 0.1. Swimming right moves it right with probability 0.35, keeps it in place with probability 0.6, and moves it left with probability 0.05. Swimming left in the leftmost state gives a reward of 0.005, and swimming right in the rightmost state gives a reward of 1. All other rewards are zero. The agent must therefore give up the small nearby reward and swim against the current to reach the large one. Episodes last 40 steps, and the return is the sum of rewards discounted by $\gamma=0.99$ per step.

We generate each transition from shared noise with two types of SCM. In both, a transition has three possible outcomes (move left, stay, move right), whose probabilities depend on the action. An inverse-CDF SCM draws a uniform number $u\in[0,1]$, lists the three outcomes in a fixed order, splits $[0,1]$ into consecutive intervals whose lengths are their probabilities, and returns the outcome whose interval contains $u$. A Gumbel-max SCM~\citep{oberst2019counterfactual} draws one Gumbel variable per outcome and returns the outcome with the largest sum of log probability and Gumbel variable.

All of these SCMs give the same transition probabilities, but they couple the outcomes of different actions differently. For example, with the order (left, stay, right), a small $u$ gives ``left'' under both actions, whereas with the order (right, stay, left), a small $u$ gives ``right'' under swimming right but ``stay'' under swimming left. We test all six orderings of the three outcomes. For inverse-CDF sampling, reversing an ordering gives the same coupling, so the six orderings give three distinct couplings. For Gumbel-max sampling, the Gumbel variables are exchangeable, so all six give the same coupling. We compare all of them with RLOO, which uses independent noise.

At a uniform, an intermediate, and a near-optimal policy, we estimate gradients using 400 batches of 64 groups with $m=4$, and compare them with the exact gradient. We also train five policy seeds for each ordering.

\Cref{fig:app:coupling} shows the results. Every estimated mean gradient has cosine similarity above 0.998 with the exact gradient, and the final return is similar (16.0445 for RLOO vs. about 16.0512 for \ours{}), allowing us to compare their variance in a controlled manner. Compared with RLOO, the shared-noise SCMs give no consistent variance reduction at the uniform policy (ratio 0.89--1.08), a small one at the intermediate policy (1.08--1.22$\times$), and about a twofold one near the optimum (1.96--2.16$\times$). Importantly, \ours{} remains unbiased under all tested SCMs, and the differences in variance reduction are small, indicating robustness to the choice of SCM.}

\subsection{A regime where \ours{} is predicted to not help}
\label{app:interaction-bandit}

\begin{table}[t]
\centering
\small
\caption{\rev{Gradient variance of \ours{} relative to RLOO on interaction bandit. Values $<1$ favor \ours{}.}}
\label{tab:interaction-bandit}
\begin{tabular}{lccccccc}
\toprule
$\lambda=\sigma_{\mathsf{luck}}$ & 0 & 0.5 & 1 & 1.5 & 1.58 & 2 & 3 \\
\midrule
\ours{} / RLOO & 3.45 & 2.77 & 1.75 & 1.08 & 1.00 & 0.70 & 0.35 \\
\bottomrule
\end{tabular}
\end{table}

\rev{\Cref{sec:hurt} shows that \ours{} can have a larger gradient variance than RLOO when the noise mostly interacts with the action rather than acting as shared luck. We test this prediction in a minimal one-step bandit with actions $a\in\{-1,1\}$ and return $G=\lambda w+a(u+\mu)$, where $u$ and $w$ are independent and uniform on $\{-1,1\}$. The noise $u$ decides which action is better, so $\sigma_{\mathsf{inter}}=1$. The noise $w$ is luck shared by both actions, so $\sigma_{\mathsf{luck}}=\lambda$. The offset $\mu>0$ makes $a=1$ better on average. The policy is a sigmoid of one logit. We use groups of $m=4$ and exact counterfactuals, which here share both $u$ and $w$ with the factual. For $m=4$, \Cref{sec:hurt} predicts that \ours{} has the larger variance whenever $\sigma_{\mathsf{luck}}<\sqrt{m-2+2/m}\,\sigma_{\mathsf{inter}}\approx1.58$.

We first estimate the gradient variance at the uniform policy with $\mu=0.2$, using $2\times10^{6}$ groups per estimator. \Cref{tab:interaction-bandit} shows that, without luck, \ours{} has 3.45$\times$ the variance of RLOO. The ratio decreases as luck grows, equals 1.00 at $\lambda=1.58$, and falls to 0.35 at $\lambda=3$. The crossover thus matches the theory.

We then train both estimators from the uniform policy with $\mu=0.05$, one group per update, and 2,000 SGD updates with learning rate 0.05 over 200 seeds. Intervals are 95\% normal confidence intervals over seeds. Without luck, \ours{} reaches a probability of 0.835 $\pm$ 0.017 of choosing the better action, compared with 0.855 $\pm$ 0.007 for RLOO. \ours{} remains unbiased, but it learns slightly more slowly and varies more across seeds. With $\lambda=3$, the order reverses. RLOO drops to 0.790 $\pm$ 0.031, whereas \ours{} is unchanged at 0.835 $\pm$ 0.017, as the shared luck cancels within each group. These results confirm our findings across the main text and in \Cref{sec:hurt} that \ours{} helps in settings where environmental complexity and luck dominate action in a consistent manner.}

\end{document}